%% file: main.tex
\documentclass[letterpaper]{article} 
\usepackage[preprint]{aaai2027}
\nocopyright
\usepackage[hyphens]{url}  
\usepackage{graphicx} 
\usepackage{natbib}  
\usepackage{caption} 
\usepackage{algorithm}
\usepackage{algorithmic}
\usepackage{booktabs}
\usepackage{amsmath}
\usepackage{amssymb}
\usepackage{array}
\usepackage{fvextra}
\usepackage{framed}

\newcommand{\gt}{m}
\newcommand{\pr}{\hat{m}}

\title{Decoupling Perception from Description: Computation-Grounded Representation Alignment between Multivariate Time Series and Language}
\input{authors}

\begin{document}
\maketitle

\begin{abstract}
Training multimodal models to align time series with language runs into a
self-supervision trap. The usual recipe asks an LLM to read a series and write a
description, so label quality is capped by the perceptual skill the model is
supposed to learn. The data can never teach more than the labeler already knows.
A second gap makes this worse: most datasets use a single variable, but the
patterns that matter (cross-channel correlation, lead-lag structure,
co-occurring anomalies) appear only with several variables, right where the
labeling LLM's limits are most exposed. These two problems create a trilemma:
existing methods are reliable, realistic, or scalable, but none achieves all
three. We resolve this by decoupling perception from description. Deterministic
code computes a set of statistics from real, open-source multivariate series;
the LLM verbalizes those precomputed facts. Perception, which LLMs do poorly, is
handled by computation, while the LLM handles expression. This produces CGTime,
our 4B-parameter computation-grounded time-series--language model. CGTime
outperforms far larger general-purpose models on multivariate understanding
tasks: it
attains the best multivariate fact score on our held-out benchmark (0.283
vs.\ 0.173 for GPT-4o-mini and 0.203 for GPT-5.4-nano), a gap that survives
Holm-corrected paired significance tests against every baseline. It also states
verifiable numerical facts in generated captions more accurately and covers a
broader range of statistical properties.
\end{abstract}

\section{Introduction}

Large language models have limitations when interpreting time series directly,
including errors in identifying trends, peaks, and relations across channels
\citep{merrill2024language}. Classical time-series tools and foundation models
primarily return numerical predictions rather than natural-language explanations
\citep{jin2024timellm,ansari2024chronos,woo2024moirai,kong2025timemqa}. Meanwhile,
LLMs can serve as reasoning components in autonomous agents that act on temporal
signals \citep{yao2023react,schick2023toolformer}. Recent work addresses this gap
by building multimodal models that join time series with language, allowing
users to query temporal data in plain language
\citep{xie2024chatts,kong2025timemqa,wang2025itformer,zhang2024llmtssurvey}. But
training these models requires paired data, and a common construction procedure
can introduce a supervision bottleneck. When an LLM reads a series and writes
its description, errors in its interpretation can enter the resulting labels.

Time-MQA illustrates this bottleneck: it derives around 200K question--answer
pairs by prompting an LLM over real series, so perceptual errors from the
labeling model may enter the training data \citep{kong2025timemqa}.
PATRA and ITFormer improve the model side through a pattern-aware module and a
lightweight encoder, respectively, but neither changes the source of
supervision \citep{lu2026patra,wang2025itformer}. ChatTS instead synthesizes
series with prescribed properties, which makes the labels controllable but uses
synthetic rather than measured series. TimeOmni-1 relies on manual annotation,
which is costly to scale \citep{xie2024chatts,guan2025timeomni}. This
reliability--realism--scalability trilemma becomes more pronounced for
multivariate data. Direct labeling is difficult for both human annotators and
LLMs: visual comparisons of heterogeneous multivariate series are sensitive to
the chosen normalization and display \citep{aigner2011bertin}, while LLM
accuracy drops on value extraction and computation even for a single numerical
sequence \citep{arai2025evaluating}. With $K$ variables, a labeler must also
consider $O(K^2)$ candidate channel pairs. Existing multivariate coverage
remains partial: ChatTS uses synthetic series, TimeOmni-1 includes only a small
real evaluation set, ITFormer focuses on a single engine domain, and PATRA
focuses on within-channel patterns.

Decoupling perception from description addresses both scales. Deterministic
computation allows the pipeline to construct supervision for multivariate
series without asking a labeler to inspect every channel pair. For univariate
series, it provides exact, verifiable targets for statistics that direct
numerical reading may estimate imprecisely. We exclude causal links, which
require assumptions beyond observational data.

\begin{figure*}[t]
    \centering
    \includegraphics[width=\linewidth]{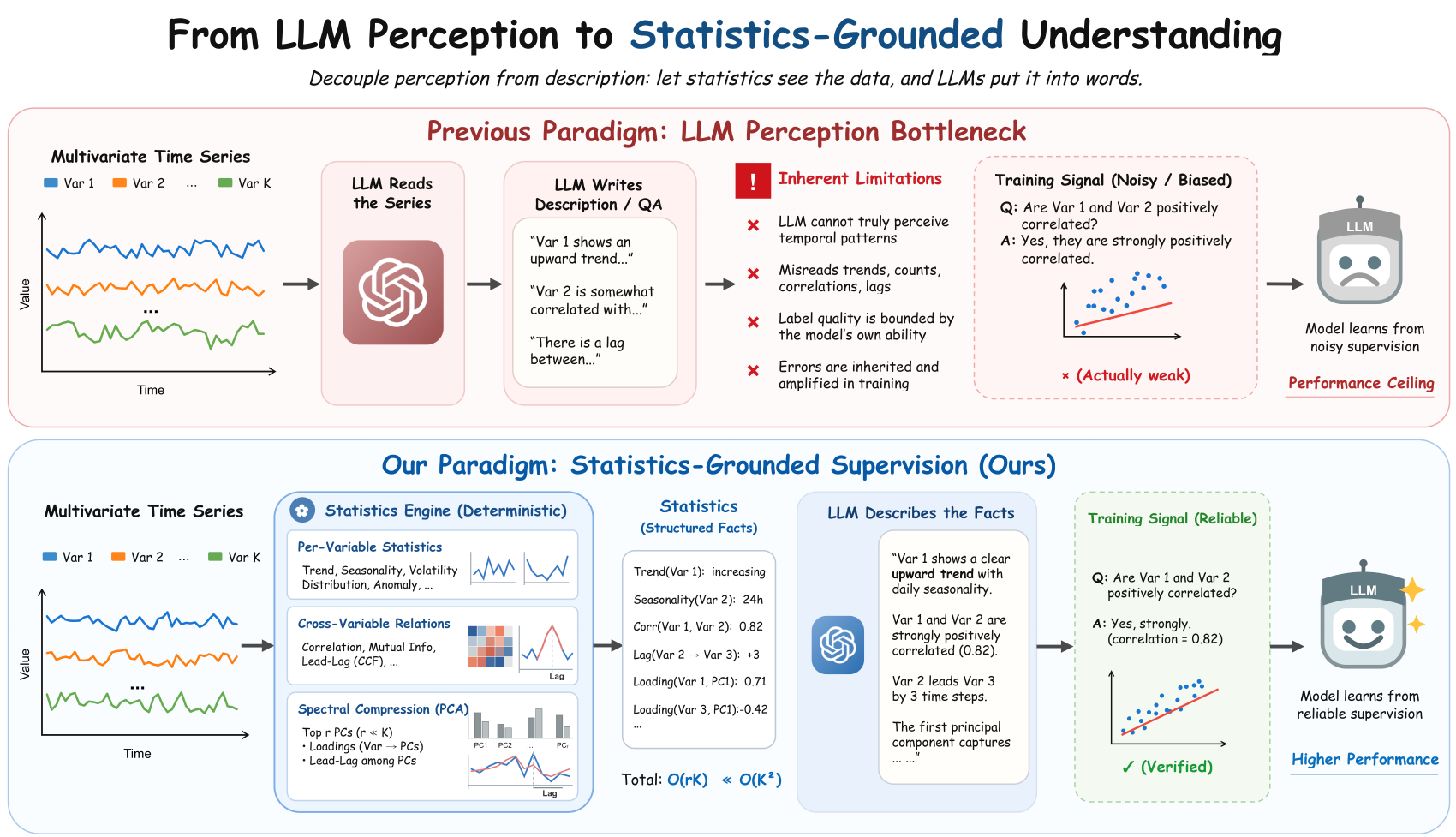}
    \caption{Overview of statistics-grounded supervision. Pipelines based on
    LLM-generated labels ask the same model to interpret a time series and
    write its label, so perceptual errors can enter the supervision (top).
    CGTime instead computes statistics with deterministic code, asks an LLM to
    verbalize these facts, and reuses the facts to verify the training signal
    (bottom).}
    \label{fig:framework}
\end{figure*}

We train CGTime, a computation-grounded time-series--language model
designed to address the trilemma by decoupling perception from description.
Deterministic code computes a fixed set of statistics from real, open-source
multivariate series, and the LLM verbalizes those precomputed facts. This
division targets all three objectives: reliability through computed rather than
model-estimated statistics, realism through measured rather than synthesized
series, and scalability through automated computation and verbalization.
Representative
sources include ETT transformer telemetry \citep{zhou2021informer}, the
cross-domain Monash Forecasting Archive \citep{godahewa2021monash}, ERA5-based
weather records \citep{hersbach2020era5}, FRED-MD macroeconomic indicators
\citep{mccracken2016fredmd}, and measured air-quality and energy systems
\citep{devito2008airquality,candanedo2017appliances,salam2018tetouan}.
Appendix~\ref{app:dataset_sources} lists the major dataset sources. The same
statistics anchor the captions and provide a verifiable reward during
reinforcement learning \citep{guo2025deepseekr1}. The supervised and RL stages
therefore use computed facts as a shared reference.

We build a pipeline around this principle (Figure~\ref{fig:framework}). For each multivariate
series, we compute a fixed set of statistics and generate captions that verbalize
them. To avoid the $O(K^2)$ blowup from describing every channel pair, we apply
PCA to extract $r \ll K$ principal components \citep{jolliffe2002pca} and
describe how each channel correlates with the retained components, together with lead-lag relations between selected channels and their most correlated components.
This reduces the description to $O(rK)$ terms while retaining the
cross-channel structure represented by the selected components. Training uses
a difficulty curriculum, from coarse summaries to exact values, with numeric
rewards in the RL stage
\citep{shao2024deepseekmath}. Section~\ref{sec:method} details each step.

\begin{figure*}[t]
    \centering
    \includegraphics[width=\linewidth]{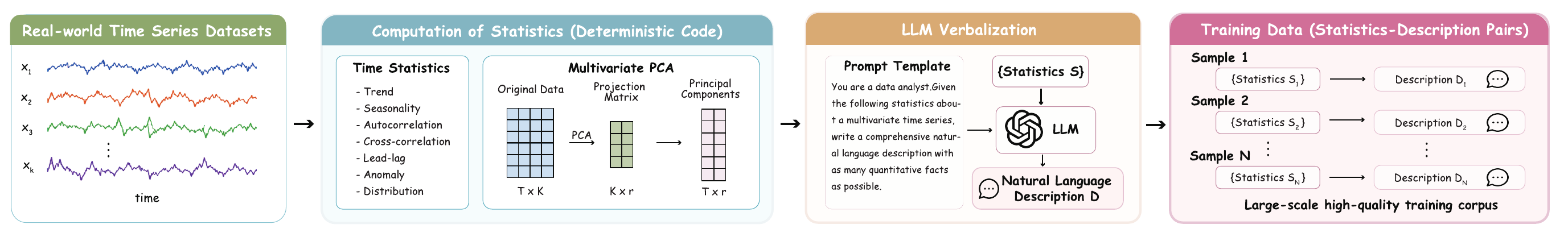}
    \caption{Computation-grounded time-series data generation. Deterministic
    code extracts time-domain statistics and a PCA-based summary of
    cross-channel structure from each real multivariate series. An LLM is
    prompted to verbalize the supplied statistics, yielding
    statistics--description pairs grounded in the original series for training
    CGTime.}
    \label{fig:data_gen}
\end{figure*}

We make the following contributions.
\begin{itemize}
\item \textbf{A design principle for time-series--language alignment: decouple
perception from description.} We identify a supervision bottleneck that arises
when one model both interprets temporal patterns and expresses them in language.
Separating these roles is designed to combine reliability from deterministically
computed facts, realism from measured multivariate series, and scalability from
automated computation and verbalization. We instantiate this principle with a
spectral step that uses PCA to reduce description complexity from $O(K^2)$ to
$O(rK)$ while retaining information about cross-channel structure.
\item \textbf{Statistics as a dual-use training signal.} The same computed
statistics are ground truth for supervised fine-tuning and a verifiable reward for
reinforcement learning, so both stages optimize against a consistent,
computation-grounded signal. The curriculum moves from coarse summaries to
precise values before the RL stage applies numerical rewards. Our 4B-parameter
model outperforms much larger general-purpose models on the multivariate
benchmark (mean fact score 0.283 vs.\ 0.173 for GPT-4o-mini and 0.203 for
GPT-5.4-nano, significant under Holm-corrected paired tests against every
baseline).
\item \textbf{A reproducible pipeline and benchmark.} We will release the
complete pipeline, with detailed documentation of data curation, deterministic
statistic computation, supervision generation, training, and evaluation. The
release will include the source code in a Git repository, the CGTime checkpoint,
and the benchmark data and evaluation protocol.
\end{itemize}

\section{Method}
\label{sec:method}
We formalize the decoupling principle as a computation-anchored alignment
problem. \S\ref{sec:problem} defines the metric-function framework.
\S\ref{sec:metrics}--\ref{sec:labels} construct computation-grounded facts and
multi-level text supervisions. \S\ref{sec:arch}--\ref{sec:sft} describe the
model architecture and three-stage SFT. \S\ref{sec:reward}--\ref{sec:grpo}
detail the verifiable reward and GRPO optimization. The evaluation protocol
anchors evaluation to the same computed facts.

\subsection{Problem Formulation}
\label{sec:problem}

Let $\{X_i\}_{i=1}^{N}$ denote a dataset of $N$ multivariate time-series
samples, where $i$ indexes an individual sample. Sample $i$ is represented as
$X_i \in \mathbb{R}^{K_i \times T_i}$, with $K_i$ channels and $T_i$ time
steps, and is equipped with time and channel masks $A_i^{(t)}$ and $A_i^{(c)}$
to handle variable lengths and channel counts. The model input consists of
$X_i$ and a natural-language prompt $p_i$; the target $y_i$ can be either a
numerical answer about a single metric or a free-text description. We learn
$\pi_\theta(y_i \mid X_i, p_i)$, whose generated numerical facts must agree
with a deterministic computation pipeline.

Formally, let $\mathcal{F} = \{f_a\}_{a=1}^{A}$ denote a fixed library of $A$
deterministic metric functions. In our implementation, $A=169$; the value type,
error definition, and tolerance for each answerable statistical property are
specified in Appendix~\ref{appendix_scoring_functions}
(Table~\ref{tab:metric_full}). The index $a$ identifies a particular statistical
property (e.g., a marginal, cross-channel, or system-level property), and $f_a$
computes that property from a masked time-series sample. For sample $i$, define
\begin{multline}
\mathcal{M}_i = \{(a,m_{i,a}) \mid a\in\{1,\ldots,A\},\\
m_{i,a}=f_a(X_i,A_i^{(t)},A_i^{(c)}),\
f_a \text{ is defined for sample } i\}.
\label{eq:computed_facts}
\end{multline}
Thus, $\mathcal{M}_i$ is the sample-specific set of valid metric--value pairs:
each element records a metric index $a$ and its deterministically computed value
$m_{i,a}$. Its cardinality may vary across samples because not every metric is
defined for every input. Metrics requiring more channels, a longer duration, or
a different data type are automatically excluded from QA construction, reward
computation, and evaluation. The model is never asked to estimate an undefined
statistic nor penalized for missing one.

This instantiates the decoupling principle: $\mathcal{F}$ handles perception
(reproducible numerical facts), while the language model handles expression.
Training and evaluation are both anchored to $\mathcal{M}_i$, so correctness is
verified against computed facts rather than subjective preferences.

\subsection{Computation-Grounded Metric Set}
\label{sec:metrics}

The metric family $\mathcal{F}$ covers univariate, bivariate, and multivariate
time-series structure. Every metric is deterministically computed from real, open-source series; 
no model-based inference or human judgment enters the ground-truth values.

We illustrate the design with multivariate PCA. After mask-aware normalization,
we compute the sample covariance matrix and order its principal components by
decreasing eigenvalue. Each component's explained-variance ratio is its
eigenvalue divided by the sum of all eigenvalues, and we retain the smallest
prefix whose cumulative ratio reaches a fixed threshold $\tau$. Answerable
summaries include the PC1 explained-variance ratio, the number of retained
components, and per-channel sync/decoupling ratios, without requiring the model
to enumerate the full spectral decomposition.

Windowed statistics characterize correlation structure and lead-lag relations.
Rolling stability is measured by the average Euclidean change between the
upper-triangular correlation vectors of adjacent windows. For each highly
synchronized channel, we identify the retained principal component with which it has
the largest absolute correlation and compute their cross-correlation over
bounded lags; system-level aggregates summarize these channel-to-factor lags.
Risk indicators use the Mahalanobis distance, which measures each multivariate
time point's covariance-normalized distance from the sample mean, together with
the fraction of time points that exceed predefined thresholds.

\begin{figure*}[t]
    \centering
    \includegraphics[width=\linewidth]{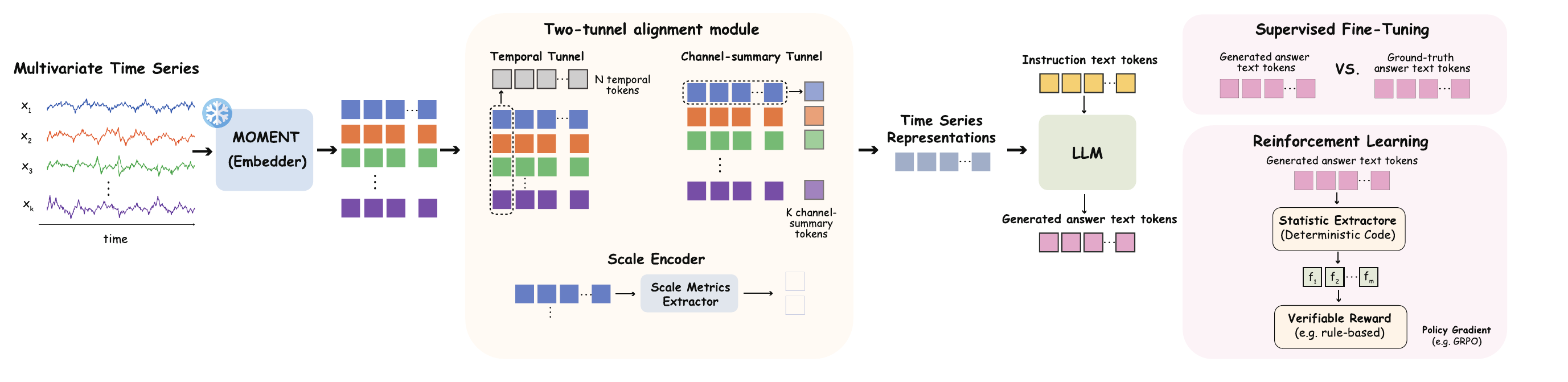}
    \caption{CGTime architecture and training. A frozen MOMENT encoder maps
    each channel to patch embeddings. The two-tunnel alignment module
    compresses them into temporal and channel-summary tokens, while the scale
    encoder adds per-channel mean and standard-deviation tokens. These
    representations condition the LLM together with the instruction tokens;
    SFT uses reference answers, and RL evaluates generated answers with
    deterministic statistic extraction and a verifiable reward.}
    \label{fig:model_arch}
\end{figure*}

\subsection{Data Generation}
\label{sec:labels}

We construct two kinds of text supervision for each sample.

\paragraph{Free-text descriptions.} Each caption example pairs a sample with a
prompt and target description. The examples are indexed by a granularity level
$\ell \in \{1,2,3,4\}$ and a prompt family $g$; for multivariate samples, $g$
identifies one of four analysis perspectives:
overall, pattern, stability, or risk. Level~1/2 are short basic captions (trend direction, ranges, peaks, basic
statistics), with facts from rule-based computation and wording polished by an
LLM given the computed facts; Level~3/4 are longer, analysis-oriented complex
captions where computed facts from $\mathcal{M}_i$ are injected into the
generation prompt, covering multivariate structure, stability, risk, and
spectral information. During training, the model sees only $X_i$ and $p_i$;
metric values supervise solely through the target text, reward, and evaluation.

\paragraph{Single-indicator QA.} Each QA example pairs a sample and a
metric-specific prompt with one answerable fact from $\mathcal{M}_i$. We include
only metrics suitable for numerical QA, excluding categorical and Boolean
properties so that all answers share a numerical reward kernel. Each prompt may
describe the metric's definition and range but never reveals its computed value.
The task is therefore controlled retrieval of a single fact from the raw series.
For SFT we additionally generate
chain-of-thought rationales for the QA pairs: given the series, the question,
and the computed answer, an LLM writes an intermediate reasoning trace ending
in the ground-truth value, and samples whose final value disagrees with the
computed answer are dropped. RL uses the raw QA pairs without CoT labels.

\paragraph{Sampling weights.} Caption levels use the sampling ratio
$w_1:w_2:w_3:w_4=1:1:3:2$,
giving longer Level~3/4 samples a larger training share while Level~1/2 supply
stable short-text alignment. For multivariate descriptions, prompt families
(overall, pattern, stability, risk) are weighted
$(0.4, 0.25, 0.2, 0.15)$,
ensuring the same series is described from multiple semantic angles.

\subsection{Model Architecture}
\label{sec:arch}

CGTime consists of a frozen time-series encoder, a trainable projector, a
scale encoder, and an autoregressive language model.

\paragraph{Time-series encoder.} We use the pretrained MOMENT embedding model
\citep{goswami2024moment} and keep it frozen during all training stages. For an
input $X_i$, the encoder produces one embedding of dimension $d_{ts}$ for each
valid channel and temporal patch. The embedding therefore has shape
$K_i \times N_i \times d_{ts}$, where $N_i$ is the number of patches.

\paragraph{Projector.} The $K_i\times N_i$ MOMENT patch
embeddings are flattened and processed by a global Transformer, allowing
channel--patch interaction before compression. After reshaping, a second
Transformer operates across channels independently at each patch, followed by
a two-layer projector into the LLM hidden space. Two learned-query attention
reductions then produce $N_i$ temporal tokens by pooling channels within each
patch and $K_i$ channel-summary tokens by pooling patches within each channel.
For the channel-axis reduction, the reported configuration uses a
masked mean when $K_i<2$.

\paragraph{Scale encoder.} For each channel $k$, we compute its mean
$\mu_{i,k}$ and standard deviation $\sigma_{i,k}$ over valid time steps. The
mean passes through a signed log transform and the standard deviation through
a log transform; a small MLP maps them to two channel-specific scale tokens
$s_{i,k}^{\mu},s_{i,k}^{\sigma}\in\mathbb{R}^{d_{lm}}$.

\paragraph{LLM integration.} The prompt template contains a
\texttt{<ts> </ts>} placeholder pair. At the forward pass, this interval is
replaced by the $N_i$ temporal tokens, the $K_i$ channel-summary tokens, and
the two scale tokens associated with each valid channel, together with learned
boundary and type embeddings. Let $B_i$ denote the ordinary text-token
embedding sequence; the complete input is obtained by splicing this
time-series block, denoted $I_i$, into $B_i$.

The language model autoregressively predicts the target conditioned on the
prompt and this spliced time-series block. Cross-entropy is computed only at
target-text positions; time-series, scale, boundary, and prompt tokens are
assigned an ignore index.

\subsection{Supervised Fine-Tuning}
\label{sec:sft}

Supervised training proceeds in three stages of increasing difficulty. All
stages use standard next-token cross-entropy, computed only over target-text
positions.

\paragraph{Stage~1: Basic alignment.} Short basic captions train the projector
$\phi$, scale encoder $\psi$, and boundary embeddings $e$ with the LLM frozen.
This establishes the mapping from time-series tokens to language representations
without altering generation capability.

\paragraph{Stage~2: Multi-granularity caption SFT.} Both basic and complex captions are
used with per-level weights $w_\ell$ (\S\ref{sec:labels}). The LLM is unfrozen,
and all parameters $(\theta, \phi, \psi, e)$ update jointly. Each level's
cross-entropy contribution is scaled by its sampling weight, with additional
prompt-family weighting for
multivariate samples. This stage scales the model from short factual descriptions
to long-form analysis incorporating PCA, correlation-structure, lead-lag,
spectral, and risk metrics.

\paragraph{Joint task warmup.} Before RL, the model is fine-tuned
on a mixture of Metric-QA, Caption, and TSQA examples. For QA, the model is
trained to produce structured outputs with \texttt{<think>}, \texttt{<answer>}, and
\texttt{\textbackslash boxed\{\}} tokens, familiarizing it with the
structured-output convention and reducing reward sparsity from format errors
during early RL steps.

\subsection{Verifiable Reward}
\label{sec:reward}

RL draws prompts from a joint pool that combines internal Metric-QA and caption
prompts with prompts from the external TSQA benchmark. The three subtasks are
sampled with equal weight. For each prompt the policy samples $G$ candidates.
For the internal tasks, rewards are computed from the same metric set
$\mathcal{M}_i$ used during SFT; for TSQA tasks, free-text prompts use ROUGE-L as the factual reward,
whereas choice prompts use valid-choice and exact-answer as rewards; each uses a task-appropriate format component.

\paragraph{QA reward.}
Outputs follow a structured format with \texttt{<think>}, \texttt{<answer>}, and
\texttt{\textbackslash boxed\{\}} tokens. A format reward $r_{\text{fmt}}(y) \in [-0.6, 0.6]$
gates the factual reward: when $r_{\text{fmt}}(y) < \eta$, the factual reward is
zeroed. The predicted value $\hat{m}_{i,a}$ is extracted strictly from
\texttt{\textbackslash boxed\{\}} within \texttt{<answer>}. For continuous
metrics, a Gaussian kernel scores the prediction:
\begin{equation}
\kappa_a(\hat{m}_{i,a}, m_{i,a})
= \exp\!\left(-\frac{e_a(\hat{m}_{i,a}, m_{i,a})^2}{2\sigma_a^2}\right),
\end{equation}
where $e_a$ is absolute or relative error. Integer metrics use a step function
(exact match = 1, half-tolerance = 0.5). The factual reward for a prompt targeting
multiple metrics is the metric-weighted mean of their kernel scores. The final
QA reward adds this factual component, scaled by $\lambda_{\text{qa}}$, to the
format reward.

\paragraph{Caption reward.}
Free-text captions are not format-gated. A rule-based extractor identifies
metric--value pairs $\widehat{\mathcal{M}}(y) = \{(a, \hat{m}_{i,a})\}$ from the
generated text. Precision averages the Gaussian kernel scores of all scorable
extracted claims. Recall counts extractions whose kernel score reaches threshold
$h$ and divides by the smaller of the number of answerable metrics and a cap
$D_{\max}$; the result is capped at one because captions are expected to mention
salient facts rather than enumerate every metric. Their standard $F_\beta$
combination forms the factual caption reward. The final caption reward scales
this component by $\lambda_{\text{cap}}$ and optionally adds a density term.

Metric-QA and caption tasks share the same supervision logic: the reward measures
whether extractable numerical facts in generated text agree with $\mathcal{M}_i$.
Full kernel definitions, per-metric error definitions and tolerances, and the
metric-agnostic relative-error buckets used for evaluation are provided in
Appendix~\ref{appendix_scoring_functions}.

\subsection{GRPO Optimization}
\label{sec:grpo}

We optimize via GRPO following \citet{shao2024deepseekmath,guo2025deepseekr1}.
For each prompt, the old policy samples $G$ candidates. Their rewards are
standardized within the group, eliminating the need for a learned value model,
and the resulting advantages are clipped to $[-3,3]$. We then optimize the
standard token-level GRPO surrogate: the current-to-old policy probability ratio
is clipped to $[1-\epsilon_c,1+\epsilon_c]$, and the objective uses the more
conservative of the clipped and unclipped advantage-weighted terms. To limit
drift from the frozen SFT reference policy, we add a K3 KL penalty with weight
$\beta_{kl}$; for log-probability difference $\delta$ between the reference and
current policies, this penalty is $\exp(\delta)-\delta-1$.
For Metric-QA, the reward bandwidth is set per split (a $\sigma$-multiplier of $0.5$ for
univariate and bivariate samples and $1.0$ for multivariate), balancing the
reward signal across splits without reweighting the loss.
Caption instead uses a fixed multiplier of $0.5$, while TSQA does not use the
Gaussian factual reward.

\input{experiments/6_experiments}

\clearpage
\bibliography{aaai2027}

\clearpage
\appendix
\section*{Supplementary Material}
\input{experiments/appendix}

\end{document}

%% file: authors.tex
\author{
    Xinran Feng\textsuperscript{\rm 1}\equalcontrib,
    Yi Xie\textsuperscript{\rm 2,\rm 3}\equalcontrib,
    Chao Zhang\textsuperscript{\rm 1},
    Ruikun Li\textsuperscript{\rm 1},\\
    Wanyun Ling\textsuperscript{\rm 2,\rm 3},
    Ziyue Li\textsuperscript{\rm 2,\rm 3},
    Chenxi Liu\textsuperscript{\rm 4}
}

\affiliations{
    \textsuperscript{\rm 1}Thrust of Financial Technology, The Hong Kong University of Science and Technology (Guangzhou)\\
    \textsuperscript{\rm 2}Technical University of Munich\\
    \textsuperscript{\rm 3}Heilbronn Data Science Center, Munich Data Science Institute\\
    \textsuperscript{\rm 4}Hong Kong Institute of Science \& Innovation, Chinese Academy of Sciences\\
    xfeng863@connect.hkust-gz.edu.cn, \{chaoz, ruikunli\}@hkust-gz.edu.cn,
    \{aaron.xie, wanyun.ling, ziyue.li\}@tum.de\\
    chenxi.liu@cair-cas.org.hk
}

%% file: experiments/6_experiments.tex
\section{Experiments}

We evaluate our model and baselines on internal and external benchmarks and conduct ablations to answer the following research questions:

\begin{itemize}
    \item \textbf{RQ 1:} Can LLMs and time-series--language models directly read and understand statistical properties of (multivariate) time series without external tools? (\S\ref{sec:evaluation_results})
    \item \textbf{RQ 2:} Do our statistics-grounded data and training recipe instill such capabilities into our model? (\S\ref{sec:evaluation_results})
    \item \textbf{RQ 3:} How does each main component of our method contribute to the overall gains? (Appendix~\ref{ablation_studies})
\end{itemize}

\subsection{Experimental Setup}

\subsubsection{Benchmarks}

Metric-QA and Captioning use the same frozen held-out set of 2,000 time-series
observations, equally divided into univariate and multivariate splits (1,000
each), but construct different task-specific requests from each 
observation.

For external evaluation, we additionally use 3,264 cleaned TSQA questions
covering open-ended QA, multiple-choice questions, and true-false tasks.
Training-data construction, filtering, and exact split statistics are detailed
in Appendix~\ref{appendix_data_details}.

\subsubsection{Baselines}

We evaluate the following models on both internal benchmarks:
TimeOmni-1-7B \citep{guan2025timeomni}, Time-MQA-Mistral-7B \citep{kong2025timemqa}, 
Time-MQA-Qwen-2.5-7B \citep{kong2025timemqa}, ChatTS-14B \citep{xie2024chatts}, 
and GPT-OSS-20B \citep{gptoss2025}. Because Qwen3-4B-Instruct-2507
\citep{yang2025qwen3} is our language backbone, we also include
Qwen2.5-7B-Instruct \citep{yang2024qwen2}, Qwen3-14B
\citep{yang2025qwen3}, and Qwen3-4B-Instruct-2507
\citep{yang2025qwen3}. Additional proprietary baselines are GPT-4o-mini
\citep{gpt4omini2024} and GPT-5.4-nano \citep{gpt54nano2026}; on TSQA, we
additionally evaluate GPT-4o \citep{gpt4o2024}.

Complete prompts, model-specific input procedures, decoding limits, and
subsampling statistics are provided in
Appendices~\ref{appendix_prompts} and~\ref{appendix_inference_configuration}.

\subsubsection{Significance Test}
Because all models answer the same questions, we compare their per-item
Gaussian scores using paired normal-approximation mean tests. We apply Holm
correction to the prespecified multivariate comparisons between our model and
each baseline; other comparisons use unadjusted 95\% confidence intervals.
Appendix~\ref{appendix_significance_test} details the test and
multiple-comparison procedure.

\subsection{Evaluation Results}
\label{sec:evaluation_results}

\input{experiments/tables/captioning_results}
\subsubsection{Captioning}

\input{experiments/tables/qa_main_combined}
We give all models the same captioning prompts and use a rule-based extractor
to identify numerical claims in their outputs. Conditional precision averages
Gaussian claim scores over captions with at least one recognized claim; recall
also includes captions with no recognized claim and normalizes by the finite
prompted properties for each record. Table~\ref{tab:captioning} reports the
internal results.
Our model achieves the highest conditional precision (0.399).  The next-highest
score, 0.348 from Time-MQA-Qwen-2.5-7B, is based on recognized claims in only
525 of 2,000 captions, whereas our model has recognized claims in 1,999.

\begin{figure}[t]
    \centering
    \includegraphics[width=\linewidth]{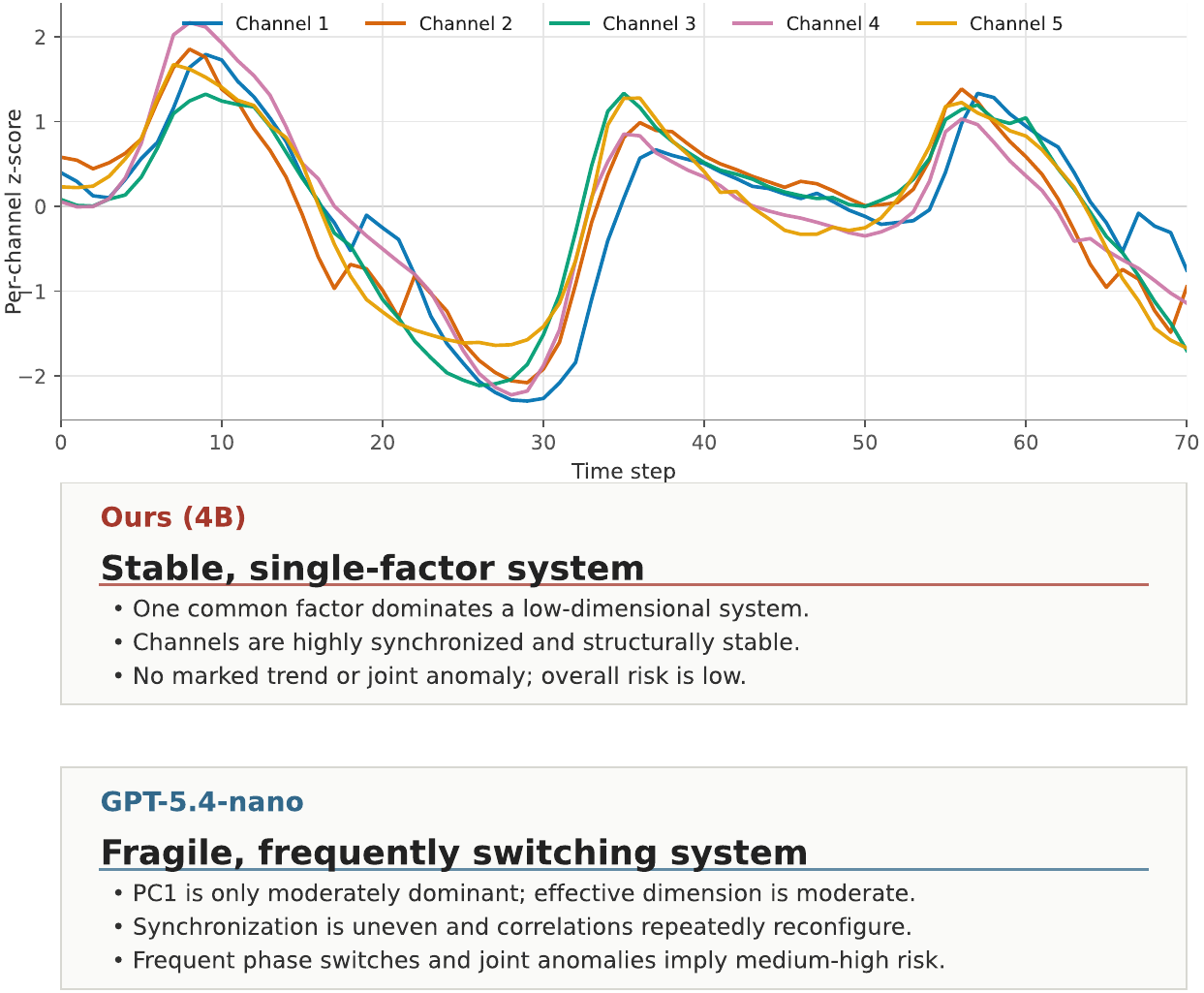}
    \caption{Contrasting diagnoses of the same five-channel series. CGTime correctly identifies a stable,
low-dimensional system. GPT-5.4-nano diagnoses frequent structural switching and elevated anomaly risk. 
}
    \label{fig:caption_case_study}
\end{figure}

A caption containing few accurate numerical claims can attain high precision despite limited comprehensiveness, so we also report recall.
Our model reaches the highest recall (0.186 versus $0.018\sim0.141$ for the external models),
a substantial improvement indicating more comprehensive captions with more
numerical claims.

Although task familiarity may partly contribute, reporting the most precise
computed numbers requires understanding the underlying statistical properties
and accurately predicting their values rather than mere task familiarity. Our
model therefore identifies relevant properties while generating comprehensive
and relatively accurate numerical content, both crucial to time-series captioning.

Figure~\ref{fig:caption_case_study} contrasts the models' opposing
conclusions on a representative five-channel case: 
where our model correctly identifies a stable,
low-dimensional system governed by a common factor, GPT-5.4-nano
diagnoses frequent structural switching and elevated anomaly risk. This
illustrates the practical implications of their performance differences;
Appendix~\ref{case_study} provides more comparisons with complete captions.

\subsubsection{Metric-QA}
Table~\ref{tab:qa_main} reports overall and multivariate Metric-QA results
with paired Gaussian comparisons. Our 4B model scores 0.288 overall, exceeding
the strongest external baseline, GPT-5.4-nano at 0.244.  The unadjusted paired
95\% confidence intervals favor our model against all ten external baselines;
against GPT-5.4-nano, the difference is $+0.044$ with a 95\% confidence
interval of $[+0.022,+0.065]$.  On the
multivariate subset, its 0.283 exceeds every baseline at face value, including
the proprietary LLMs; this advantage is statistically significant, and all
ten paired differences survive Holm--Bonferroni correction. We additionally
report the hard-threshold Rel25 metric; Appendix~\ref{full_split_analysis}
provides the complete univariate and multivariate analysis.

Table~\ref{tab:qa_main} answers RQ1: despite receiving the instructions needed
to calculate the statistical properties, off-the-shelf models, including
proprietary models, perform poorly on the multivariate subset, and most score
lower there than overall.
GPT-5.4-nano, for example, drops from 0.244 to 0.203,
whereas our model largely maintains its performance (0.288 overall versus
0.283 multivariate). The poor performances of external baselines are
consistent with capability limitations under the
evaluation protocol; Appendix~\ref{appendix_statistic_family} examines the
contrast in more detail through a statistic-family analysis. These results also
address RQ2: our trained 4B model outperforms
larger external models, including proprietary LLMs, under the same protocol.
Qualitative inspection further shows that the model has learned to reconstruct
complex statistics in its rollouts. Detailed case studies are provided in
Appendix~\ref{case_study}.

\input{experiments/tables/tsqa_results}
\subsubsection{TSQA}
Table~\ref{tab:tsqa_results} provides external-benchmark evidence for RQ2. We
split TSQA 90/10 for training and evaluation, mix the training portion with
our internal data during Joint-SFT warmup and Joint-GRPO, and average
per-question accuracy and ROUGE-L within each subtask. Our model achieves the
highest ROUGE-L under our pipeline and is numerically close to PATRA's
strongest reported results 
\footnote{Not a head-to-head comparison since PATRA has not been fully open-sourced.}
. These results partially address RQ2 through
external-benchmark performance. Note that although GPT-4o performs best on this benchmark, 
it still scores 0.533 when the time series is removed in a TS-blind
ablation, indicating substantial exploitable
signal in the question text and answer options alone. Our model outperforms all other baselines.

\subsubsection{More Results}

Appendix~\ref{ablation_studies} provides the RQ3 ablations;
Appendix~\ref{case_study} gives the caption case studies; and
Appendix~\ref{appendix_limitations} discusses limitations.
Appendices~\ref{appendix_significance_test}--\ref{appendix_statistic_family}
provide significance-test details, complete two-split performance analysis,
and statistic-family bottleneck analysis.

\section{Conclusion}

We present a statistics-grounded approach to supervising multivariate
time-series--language models. Unlike traditional methods that ask LLMs to
generate training signals directly from textualized raw time series, we use
deterministic calculations to extract verifiable statistical properties from
real multivariate series and ask LLMs to describe the series using these
properties and use these description as training data. CGTime outperforms all external baselines, including proprietary
LLMs, on the Metric-QA test set, generates more comprehensive and accurate free-form captions in caption task, and performs
competitively on TSQA. These findings support
statistics-grounded supervision as a feasible way to instill numerical
reasoning capabilities into time-series--language models.

%% file: experiments/tables/captioning_results.tex
\begin{table}[t]
\centering
\footnotesize
\setlength{\tabcolsep}{5pt}
\begin{tabular}{lcc}
\toprule
Model  & Precision & Recall \\
\midrule
\multicolumn{3}{l}{\emph{Proprietary LLMs}}\\
GPT-5.4-nano        & 0.270 & 0.097 \\
GPT-4o-mini         & 0.306 & 0.063 \\
\midrule
\multicolumn{3}{l}{\emph{Open-source general-purpose LLMs}}\\
GPT-OSS-20B        & 0.179 & 0.141 \\
Qwen2.5-7B-Instruct & 0.259 & 0.100 \\
Qwen3-14B           & 0.247 & 0.110 \\
Qwen3-4B-Instruct-2507 & 0.199 & 0.122 \\
\midrule
\multicolumn{3}{l}{\emph{Time-series specialists}}\\
TimeOmni-1-7B       & 0.163 & 0.060 \\
ChatTS-14B          & 0.235 & 0.094 \\
Time-MQA-Mistral-7B & 0.112 & 0.018 \\
Time-MQA-Qwen-2.5-7B & 0.348 & 0.024 \\
\midrule
CGTime (4B) & \textbf{0.399} & \textbf{0.186} \\
\bottomrule
\end{tabular}
\caption{Captioning on the 2,000-caption data set, with 1,000
univariate and 1,000 multivariate records.
Each claim is scored with the same type-aware Gaussian scoring function as Metric-QA with $\sigma$-multiplier=0.5.
}
\label{tab:captioning}
\end{table}

%% file: experiments/tables/qa_main_combined.tex
\begin{table*}[t]
\centering
\small
\setlength{\tabcolsep}{2pt}
\begin{tabular}{@{}>{\raggedright\arraybackslash}p{80pt} cc cc cc cc@{}}
\toprule
 & \multicolumn{6}{c}{Gaussian} & \multicolumn{2}{c}{Rel25} \\
\cmidrule(lr){2-7}\cmidrule(lr){8-9}
 & \multicolumn{2}{c}{Score}
 & \multicolumn{2}{c}{Paired difference (Ours 4B $-$ baseline)}
 & \multicolumn{2}{c}{Relative gain (\%)}
 & \multicolumn{2}{c}{Score} \\
\cmidrule(lr){2-3}\cmidrule(lr){4-5}\cmidrule(lr){6-7}\cmidrule(lr){8-9}
Model
& Overall
& Multivariate
& Overall $\Delta$ [95\% CI]
& Multivariate $\Delta$ ($z$)$^{\star}$
& Overall
& Multivariate
& Overall
& Multivariate \\
\midrule
\multicolumn{9}{l}{\emph{Proprietary LLMs}}\\
GPT-5.4-nano
  & 0.244 & 0.203
  & $+0.044\;[+0.022,+0.065]$
  & $\mathbf{+0.080\;(+6.71)}$
  & $+17.9$ & $+39.5$
  & 0.312 & 0.211 \\
GPT-4o-mini
  & 0.219 & 0.173
  & $+0.069\;[+0.046,+0.092]$
  & $\mathbf{+0.111\;(+7.64)}$
  & $+31.5$ & $+64.1$
  & 0.309 & 0.223 \\
\midrule
\multicolumn{9}{l}{\emph{Open-source general-purpose LLMs}}\\
GPT-OSS-20B
  & 0.223 & 0.198
  & $+0.064\;[+0.042,+0.087]$
  & $\mathbf{+0.085\;(+6.12)}$
  & $+28.9$ & $+42.8$
  & 0.278 & 0.221 \\
Qwen2.5-7B-Instruct
  & 0.193 & 0.153
  & $+0.094\;[+0.072,+0.117]$
  & $\mathbf{+0.130\;(+9.33)}$
  & $+48.8$ & $+84.5$
  & 0.267 & 0.183 \\
Qwen3-14B
  & 0.208 & 0.162
  & $+0.080\;[+0.057,+0.102]$
  & $\mathbf{+0.121\;(+9.15)}$
  & $+38.3$ & $+74.8$
  & 0.300 & 0.217 \\
Qwen3-4B-Instruct-2507
  & 0.169 & 0.136
  & $+0.119\;[+0.097,+0.140]$
  & $\mathbf{+0.147\;(+11.11)}$
  & $+70.2$ & $+108.1$
  & 0.212 & 0.159 \\
\midrule
\multicolumn{9}{l}{\emph{Time-series specialists}}\\
TimeOmni-1-7B
  & 0.172 & 0.144
  & $+0.115\;[+0.093,+0.137]$
  & $\mathbf{+0.139\;(+9.84)}$
  & $+66.9$ & $+97.0$
  & 0.235 & 0.175 \\
ChatTS-14B
  & 0.210 & 0.134
  & $+0.078\;[+0.056,+0.099]$
  & $\mathbf{+0.149\;(+10.84)}$
  & $+37.0$ & $+111.2$
  & 0.262 & 0.165 \\
Time-MQA-Mistral-7B
  & 0.050 & 0.046
  & $+0.238\;[+0.219,+0.257]$
  & $\mathbf{+0.237\;(+17.37)}$
  & $+474.6$ & $+513.7$
  & 0.082 & 0.067 \\
Time-MQA-Qwen-2.5-7B
  & 0.120 & 0.063
  & $+0.168\;[+0.146,+0.189]$
  & $\mathbf{+0.220\;(+16.02)}$
  & $+139.4$ & $+347.2$
  & 0.164 & 0.084 \\
\midrule
CGTime (4B)
  & \textbf{0.288} & \textbf{0.283} & --- & --- & --- & ---
  & \textbf{0.355} & \textbf{0.322} \\
\bottomrule
\end{tabular}
\caption{Metric-QA results on the fixed 2,000-request test set, comprising
1,000 univariate and 1,000 multivariate requests.
Gaussian uses the bandwidth multiplier $m{=}0.5$; Rel25 accepts
relative error at most 25\%. 
Relative gain is
$100(s_{\mathrm{Ours\,4B}}-s_{\mathrm{baseline}})/s_{\mathrm{baseline}}$; 
a negative value would indicate that CGTime (4B) scores lower.
$^{\star}$The ten multivariate Gaussian contrasts form the primary family,
and all survive Holm--Bonferroni correction at family-wise
$\alpha{=}0.05$.
}
\label{tab:qa_main}
\end{table*}

%% file: experiments/tables/tsqa_results.tex
\begin{table}[t]
\centering
\footnotesize
\setlength{\tabcolsep}{5pt}
\begin{tabular}{lcc}
\toprule
Model & Acc. & ROUGE-L \\
\midrule
\multicolumn{3}{l}{\emph{}}\\
CGTime (4B) & 0.650 & \textbf{0.278} \\
GPT-4o                 & \textbf{0.738} & 0.148 \\
\quad$\hookrightarrow$ \emph{TS-blind (ablation)} & 0.533 & 0.091 \\
ChatTS-14B             & 0.616 & 0.179 \\
Qwen2.5-7B-Instruct    & 0.630 & 0.166 \\
Time-MQA-Mistral-7B    & 0.334 & 0.147 \\
Time-MQA-Qwen-2.5-7B   & 0.503 & 0.147 \\
\midrule
\multicolumn{3}{l}{\emph{cited (different split and pipeline)}}\\
PATRA-7B               & 0.545 & 0.264 \\
\bottomrule
\end{tabular}
\caption{Results on the TSQA (Time-MQA) test subset. Acc.\ is the
equal-weighted macro-average of multiple-choice and true/false accuracy, while
ROUGE-L scores open-ended responses.}
\label{tab:tsqa_results}
\end{table}

%% file: experiments/appendix.tex
\DefineVerbatimEnvironment{prompt}{Verbatim}{frame=single,fontsize=\small,breaklines=true,breakanywhere=true}
\newenvironment{resultbox}{%
  \def\FrameCommand{\fboxsep=3pt\fboxrule=0.4pt\fbox}%
  \def\FrameHeightAdjust{0pt}%
  \MakeFramed{\advance\hsize-\width\FrameRestore}%
}{\endMakeFramed}

\noindent This supplementary material complements the main paper in four
parts. It first provides the broader context, dataset construction details,
source provenance, and implementation; it then records the prompts, inference
configuration, scoring rules, and extraction procedures used in evaluation.
The remaining sections present split-level and statistic-family analyses,
ablations, and qualitative examples, followed by limitations and
reproducibility information.

\input{experiments/appendix/related_work.tex}
\input{experiments/appendix/experiment_setup.tex}
\input{experiments/appendix/dataset_sources.tex}

\input{experiments/appendix/implementation.tex}

\input{experiments/appendix/prompt.tex}
\input{experiments/appendix/inference.tex}

\input{experiments/appendix/evaluation.tex}

\input{experiments/appendix/scoring.tex}
\input{experiments/appendix/answer_extraction.tex}

\input{experiments/appendix/significance.tex}

\input{experiments/appendix/full_split_analysis.tex}
\input{experiments/appendix/bottleneck.tex}

\input{experiments/appendix/ablations.tex}
\input{experiments/appendix/case_study.tex}

\input{experiments/appendix/limitations.tex}

\input{experiments/appendix/reproduce.tex}
\input{experiments/appendix/full_statistical_tests.tex}
\input{experiments/appendix/reproduction_background.tex}

%% file: experiments/appendix/related_work.tex
\section{Related work}

\paragraph{Time-series foundation models.} Time-series foundation models are
pretrained across domains to support general forecasting. Time-LLM reprograms
a frozen LLM for forecasting
\citep{jin2024timellm}, while Chronos tokenizes values into a fixed vocabulary
and models them as a language \citep{ansari2024chronos}. Moirai
\citep{woo2024moirai}, TimesFM \citep{das2024timesfm}, and MOMENT
\citep{goswami2024moment} pursue universal zero-shot forecasting. These models
transfer across domains, but produce forecasts or labels rather than
natural-language explanations and do not provide a language interface
\citep{zhang2024llmtssurvey}. We study models that describe a series in
language instead of only forecasting its continuation.

\paragraph{Multimodal time-series--language models.} Models that pair time
series with text allow users to query temporal data in natural language
\citep{zhang2024llmtssurvey}. This pairing is useful because LLMs alone still
misread trends and cross-channel relations \citep{merrill2024language}.
Existing supervision comes from several sources. Time-MQA prompts an LLM over
real series to produce about 37K open-ended reasoning questions 
in 200K question--answer pairs \citep{kong2025timemqa}. 
PATRA uses pattern-aware alignment with a balanced
reward \citep{lu2026patra}, while ITFormer fuses a lightweight encoder with a
frozen LLM \citep{wang2025itformer}. Both retain LLM-derived labels, whose
quality depends on what the labeling model can infer. ChatTS instead generates
series from prescribed attributes, so its captions are correct by construction
but its series are templated rather than real \citep{xie2024chatts}.
TimeOmni-1 uses human annotation for a small reasoning suite, with about 2.3K
curated examples among 23K samples; this produces reliable labels but is
difficult to scale \citep{guan2025timeomni}. Variable coverage is also limited:
ITFormer's QA data come from a single aero-engine domain, ChatTS uses synthetic
multivariate series, and PATRA aligns only within-channel trend and seasonality
\citep{wang2025itformer,xie2024chatts,lu2026patra}. Our labels are computed
deterministically from real multivariate series, providing reliable, realistic,
and scalable supervision while explicitly including cross-channel structure.

\paragraph{Time-series description and benchmarks.} Other work targets
open-ended descriptions of temporal patterns rather than structured QA.
ChatTime trains a multimodal foundation model with bimodal text and series
input/output \citep{wang2024chattime}; TSLM targets caption generation with
synthetic caption--series pairs and cross-modal retrieval denoising
\citep{trabelsi2025tslm}. ChatTS also supports description generation, but on
synthetic multivariate series \citep{xie2024chatts}. BEDTime tests recognition,
discrimination, and open generation of univariate visual descriptions across
text, numeric, and plot modalities \citep{sen2025bedtime}; TimeSeriesExam
procedurally generates
multiple-choice questions over core understanding categories
\citep{cai2024timeseriesexam}; and TSRBench spans perception, reasoning,
prediction, and decision-making across 15 tasks and two input modalities
\citep{yu2026tsrbench}. These benchmarks cover univariate description and a
broad set of reasoning tasks, but their labels are generally human-written,
synthetic, or template-based, with limited cross-channel structure. Our
benchmark derives captions and QA targets from computed multivariate statistics
over real series. The resulting evaluation measures recoverable numerical
facts rather than subjective wording alone.

\paragraph{Reinforcement learning from verifiable rewards.} Reinforcement
learning from verifiable rewards replaces learned preference models with
automatically checkable rewards when ground truth is available
\citep{zhang2025rlsurvey}. Early work trains outcome verifiers for math word
problems \citep{cobbe2021trainingverifiers} and compares process supervision
with outcome supervision
\citep{uesato2022processoutcome,lightman2023letsverify};
STaR bootstraps reasoning chains from verifiable final answers
\citep{zelikman2022star}. DeepSeek-R1 scales this approach through large-scale
RL on checkable tasks \citep{guo2025deepseekr1}, and DeepSeekMath introduces the
GRPO objective used in our work \citep{shao2024deepseekmath}. RLVR for
mathematics and programming can rely on symbolic checkers or code execution.
Time-series alignment usually depends on model-generated or human-judged
labels, which do not provide the same objective signal. Our computed statistics
fill this gap: we use the statistics in $\mathcal{M}_i$ as deterministic
rewards for both QA and captioning. RL therefore optimizes the same facts used
during supervised training.

%% file: experiments/appendix/experiment_setup.tex
\section{Training and Evaluation Data}
\label{appendix_data_details}

This section defines the observation unit, the training and evaluation pools,
and the filtering steps that connect the internal tasks and TSQA to the
reported experiments.

The statistics-grounded dataset is described in
\S\ref{sec:metrics}--\ref{sec:labels}. It contains approximately 1,000,000
observations in each of the univariate, bivariate, and multivariate splits,
or approximately 3,000,000 in total. An observation comprises one time
series together with its raw values, computed statistical properties, and
captions generated as described in \S\ref{sec:labels}. We reserve a candidate
pool of 10,000 observations from each split. Because the bivariate split is
not an evaluation target, the test sets contain only univariate and
multivariate observations.

We assign each series a deterministic signature and use it to audit overlap
between the training and candidate pools. After this audit, the univariate and
multivariate candidate pools contain 8,432 and 9,999 observations,
respectively. Both internal evaluation request sets are subsequently
constructed only from these audited candidate pools and therefore contain no
time-series signature present in training.

Training uses a subset of the full dataset because of resource constraints.
The Metric-QA and captioning training subsets are constructed independently
from the shared underlying observation collection and are not paired
one-to-one by observation. We construct the Metric-QA subset using per-metric
quota sampling, yielding approximately 6,000 univariate pairs and 11,000 pairs
from each of the bivariate and multivariate splits, with all 169
statistical-property types represented. CoT targets are then generated from
these pairs of time series and ground-truth statistics as described in
\S\ref{sec:labels}. Removing problematic CoT samples leaves 27,536 Metric-QA
training examples. The captioning training set contains 27,501 examples, which
keeps the two tasks approximately balanced.

For Metric-QA evaluation, we use a fixed test set of 2,000 observations, with
1,000 univariate and 1,000 multivariate observations sampled from their
corresponding held-out pools. Each observation is used for only one statistical
property, which avoids testing a model repeatedly on the same time
series. Captioning evaluation uses a separate request set over the same 2,000
time series, with 1,000 requests per variable-count split. Each request
describes 16 candidate properties without revealing their numerical
values. Metric-QA requests instantiate only properties defined for the
observation. Captioning panels are fixed by split and prompt family and can
name an undefined candidate; such candidates are removed before scoring and
do not penalize the model.

TSQA is the only external benchmark included in training. We retain its
open-ended, multiple-choice, and true-or-false subsets because the remaining
subtasks are incompatible with our training objective. These subsets initially
contain 37,628 questions. We remove 1,870 questions without a valid time series
or prompt and split the remainder 90/10 into 32,182 training questions and
3,576 evaluation questions, including 1,937 open-ended evaluation questions.
We then remove 683 questions (626 training and 57 evaluation) that contain
multiple \texttt{<ts>} tags, leaving 31,556 training and 3,519 evaluation
questions. A deterministic time-series-signature audit identifies overlap with
the training set in 255 evaluation questions involving 247 time series. After
removing these questions, the final cleaned evaluation set contains
3,264 questions: 848 multiple-choice, 632 true-or-false, and 1,784 open-ended
questions.

SFT and RL use the same TSQA training split and statistics-grounded captioning
subset. For Metric-QA, SFT uses the CoT examples, whereas RL uses resampled QA
prompts without CoT labels.

%% file: experiments/appendix/dataset_sources.tex
\section{Source Datasets}
\label{app:dataset_sources}

Our construction pool draws from two public collections: the Monash Time
Series Forecasting Archive \citep{godahewa2021monash} and Time-300B
\citep{shi2024timemoe}. We selected 41 files spanning energy, weather,
transportation, environmental, health, economic, and web-demand domains.
Table~\ref{tab:dataset_sources} lists their local identifiers after selection
and preprocessing. Some series were resampled, reindexed, or reformatted; the
table groups these versions with their underlying series rather than treating
them as separately acquired datasets.

\begin{table*}[t]
\centering
{\small
\setlength{\tabcolsep}{3.5pt}
\renewcommand{\arraystretch}{1.06}
\begin{tabular}{@{}
  >{\raggedright\arraybackslash}p{0.37\textwidth}
  >{\raggedright\arraybackslash}p{0.57\textwidth}
  @{}}
\toprule
\textbf{Local dataset identifiers} & \textbf{Real-world content} \\
\midrule
\texttt{ETTh1}, \texttt{ETTh2}, \texttt{ETTm1}, \texttt{ETTm2}
& Transformer loads and oil temperature at hourly and 15-minute resolutions
  \\
\texttt{electricity}, \texttt{electricity\_hourly\_*},
\texttt{electricity\_weekly\_*}
& Electricity consumption for 321 clients in hourly and weekly benchmark
  representations \\
\texttt{electricity\_demand}
& Half-hourly electricity demand for five Australian regions
  \\
\texttt{traffic}, \texttt{traffic\_hourly\_*},
\texttt{traffic\_weekly\_*}
& Freeway occupancy from 862 San Francisco Bay Area sensors, represented at
  hourly and weekly frequencies \\
\texttt{covid\_deaths\_*}, \texttt{hospital\_*}, \texttt{nn5\_*},
\texttt{kaggle\_web\_traffic\_*}
& COVID-19 deaths, hospital demand, banking withdrawals, and Wikipedia page
  views \\
\texttt{m4\_daily}
& One daily series retained after rescaling and reindexing \\
\texttt{solar\_*}, \texttt{wind\_*}, \texttt{saugeenday\_*},
\texttt{river\_flow}, \texttt{sunspot\_*}, \texttt{us\_births\_*}
& Solar and wind generation, river flow, sunspots, and daily U.S.\ births;
  the processed collection includes resampled versions and repeated value
  sequences \\
\texttt{WTH}, \texttt{weather}, \texttt{exchange\_rate},
\texttt{national\_illness}
& Hourly and 10-minute meteorological measurements, eight daily exchange-rate
  series, and seven weekly influenza indicators \\
\texttt{oikolab\_weather\_dataset}
& Hourly multivariate historical weather fields \\
\texttt{fred\_md\_dataset}
& Monthly U.S.\ macroeconomic indicators \\
\texttt{NASDAQCOM}, \texttt{SP500}
& Daily observations of the NASDAQ Composite and S\&P 500 indices \\
\texttt{airquality}
& Hourly gas concentrations and multisensor responses collected in an Italian
  city \\
\texttt{energy}
& Appliance consumption, indoor temperature and humidity, and nearby weather
  measurements from a low-energy house \\
\texttt{tcpc}
& Weather and power consumption for three distribution zones in Tetouan,
  Morocco \\
\texttt{metro}
& Hourly westbound I-94 traffic volume between Minneapolis and St.\ Paul
  \\
\bottomrule
\end{tabular}
}
\caption{Real-world time-series files selected from the Monash archive and
Time-300B. Each group may include resampled or reformatted versions of the same
underlying series.}
\label{tab:dataset_sources}
\end{table*}

\texttt{river\_flow} and \texttt{saugeenday} encode the same value sequence in
different formats; the two \texttt{sunspot} variants do likewise. The
one-minute solar and wind versions are downsampled from their four-second
counterparts. We treat these relationships as transformations rather than
independent sources.

The source families span the domains used by our statistics-grounded
construction. The procedure converts their heterogeneous raw formats into the
common observation representation defined in
Appendix~\ref{appendix_data_details}. A source's presence in the pool does not
mean that it appears in every generated subset. Task-specific filters on
length, channel count, and sampling determine which series are eligible.

%% file: experiments/appendix/implementation.tex
\section{Implementation Details}

This section specifies the shared time-series--language architecture and the
Base-SFT, Joint-SFT, and Joint-GRPO lineage that produces the reported
model. Training proceeds through Base-SFT, a Joint-SFT warmup that
includes TSQA, and Joint-GRPO. Every result labelled ``CGTime (4B)'' uses the
same model selected after Joint-GRPO. The main paper's Stage~1 and Stage~2 are
stage types repeated within Base-SFT across the univariate, bivariate, and
multivariate domains, whereas Replay repeats only Stage~2. The main paper's
joint task warmup is the Joint-SFT phase below. Computed statistics supervise
the internal Metric-QA and Captioning components, whereas TSQA retains its
benchmark-provided labels and references.

\subsection{Training configuration}
\label{appendix_implementation_details}
\noindent
The descriptions below distinguish configured training horizons, executed
steps, and checkpoints selected for evaluation.

\subsection*{Architecture and compute topology}
The three phases share the following architecture and compute configuration:
\begin{itemize}
  \item \textbf{Time-series encoder}: \textbf{MOMENT-1-base}, kept frozen
  without optimizer state.
  \item \textbf{Global channel--patch context}: the $C\times N$ MOMENT patch
  tokens are flattened and processed by a 1-layer, 8-head Transformer with
  dropout 0.1 before projection.
  \item \textbf{Per-patch channel context}: the alignment module applies a
  separate 1-layer, 8-head Transformer with dropout 0.1 across the $C$
  channels at each patch. This step is bypassed when $C=1$ and is followed by
  a two-layer GELU MLP into the LLM hidden space.
  \item \textbf{Two-axis compression}: two 8-head learned-query attention
  pools reduce channels per patch and patches per channel, yielding $N+C$
  tokens. Channel reduction uses a masked mean when $C<2$.
  \item \textbf{Language model}: \textbf{Qwen3-4B-Instruct-2507}.
  \item \textbf{Special tokens}: four tokens,
  \texttt{<ts></ts><scale></scale>}; the tokenizer vocabulary and embeddings
  are extended accordingly.
  \item \textbf{Learned boundary/type embeddings}: trainable start/end
  embeddings for the time-series, scale, temporal, and channel blocks,
  together with temporal and channel type embeddings.
  \item \textbf{Precision and compute topology}: bf16 compute; Base-SFT uses
  two-GPU per-layer FSDP, Joint-SFT uses one GPU with FSDP disabled, and
  Joint-GRPO uses one GPU with FSDP and DDP disabled.
\end{itemize}

\subsection*{(1) Base-SFT}
\begin{itemize}
  \item \textbf{Initialization and stage order}: we first train a Stage-1
  checkpoint on basic captions as a time-series--language alignment warmup, then run the
  full SFT curriculum from univariate Stage 1 through univariate Stage 2,
  bivariate Stage 1, bivariate Stage 2, multivariate Stage 1, multivariate
  Stage 2, and replay Stage 2, in that order.
  \item \textbf{Data}: for each univariate, bivariate, and multivariate domain,
  each Stage 1 epoch comprises approximately 100K training examples covering
  basic properties, whereas each Stage 2 epoch comprises approximately 50K
  examples spanning basic and complex properties. The two basic strata use
  sampling weights 1 and 1, whereas the
  two complex strata use weights 3 and 2. These weights determine which caption
  stratum supplies the target for each example in each epoch. Across Stage 2,
  the overall, pattern, stability, and risk prompt families are sampled with
  weights 0.4, 0.25, 0.2, and 0.15. These variable-count stages are followed by a
  replay Stage 2.
  \item \textbf{Trainable parameters}: in Stage 1, MOMENT and the LLM are
  frozen while the complete alignment module described above is trained. In
  Stage 2, the LLM is
  unfrozen and optimized jointly with the alignment module; MOMENT remains
  frozen.
  \item \textbf{Training duration and stage initialization}: Stage 1 is trained
  for three epochs and Stage 2 for two epochs in each variable-count setting. Replay contains
  only Stage 2 and is trained for two epochs. Stage transitions use
  validation-selected best checkpoints rather than final checkpoints.
  Univariate Stage 2 combines the warmup checkpoint with the best univariate
  Stage-1 checkpoint. Each later domain begins Stage 1 from the preceding
  domain's Stage-2 best, and its Stage 2 combines that checkpoint with its own
  Stage-1 best. Replay Stage 2 starts from the multivariate Stage-2 best.
  \item \textbf{Learning rate}: LLM $10^{-5}$ / alignment module $10^{-4}$.
  \item \textbf{Batch (Stage 1 / Stage 2, per GPU)}: univariate 64/48,
  bivariate 40/30, multivariate 30/24, and replay Stage 2 24.
  \item \textbf{Maximum sequence length}: Stage 1 uses 512 tokens. Stage 2
  uses 768 tokens for univariate and bivariate training, and 1024 for
  multivariate and replay training.
  \item Gradient checkpointing was enabled, with gradient accumulation set to 1.
\end{itemize}

\subsection*{(2) Joint-SFT warmup}
\begin{itemize}
  \item \textbf{Initialization}: the checkpoint selected after replay Stage 2.
  \item \textbf{Data}: a joint mixture of Metric-QA, Captioning, and TSQA
  examples. The pool contains 86{,}593 examples before filtering. Removing twelve
  unusable examples before the split gives train, validation, and test partitions
  of 83{,}118, 1{,}732, and 1{,}731.
  \item \textbf{Learning rate}: LLM $5\times10^{-6}$ / alignment module $5\times10^{-5}$.
  \item \textbf{Batch $=$ 8}, \textbf{gradient accumulation $=$ 1},
  \textbf{epochs $=$ 2}, and \textbf{maximum sequence length $=$ 1792}.
  Gradient checkpointing was enabled. The 1792-token limit accommodates the
  longest examples in the joint pool, including long Metric-QA CoTs, rather
  than being specific to TSQA.
  \item Training used one GPU with FSDP disabled and completed at step
  20{,}780, which was retained for the next phase.
\end{itemize}

\subsection*{(3) Joint-GRPO}
\begin{itemize}
  \item \textbf{Initialization}: the policy and frozen reference are both
  initialized from the selected Joint-SFT checkpoint.
  \item \textbf{Prompt pool}: before filtering, the pool contains
  \textbf{84{,}862} prompts: 25{,}805 Metric-QA, 27{,}501 Captioning, 17{,}346
  TSQA open-ended, and 14{,}210 TSQA choice prompts. Filtering removes one
  Metric-QA and one Captioning prompt for each of nine series with a non-finite
  value, leaving \textbf{84{,}844} prompts. The merged pool is shuffled and
  sampled uniformly, yielding broadly equal weights for Metric-QA,
  Captioning, and TSQA.
  \item \textbf{Optimizer}: AdamW with LLM learning rate $10^{-6}$ and
  alignment-module learning rate $5\times10^{-5}$.
  \item \textbf{RL}: 2,000 training steps;
  batch size 3 prompts/step, group size 8 rollouts/prompt, and
  gradient accumulation 4. Rollouts used temperature 0.5, top-$p$ 0.85, QA
  maximum output length 384, and caption maximum output length 1024.
  \item \textbf{Optimization controls}: KL coefficient 0.12,
  advantage clipping at 3, and
  PPO clip ratio 0.08; no-repeat $n$-gram was disabled and repetition
  penalty was 1.0.
  \item \textbf{Reward}: the factual-reward weight was set to 1.0 for each of
  the three tasks, while the \emph{form of the reward differed by task}.
  Metric-QA used
  the Gaussian factual reward plus its structured-format reward,
  with the QA-only bandwidth multiplier set by split to
  univariate/bivariate/multivariate $=$ 0.5/0.5/1.0.  Caption used rule-based
  claim extraction during reward computation, a fixed Gaussian bandwidth multiplier of 0.5,
  $F_1$ over soft mention precision and thresholded recall
  with the recall denominator capped at 7, and a mention-density bonus capped at
  0.6.  TSQA free text used ROUGE-L factual reward, whereas TSQA choice
  questions combined valid-choice and exact-answer rewards; each also had its
  task-appropriate format component.
  \item \textbf{Training length and evaluated checkpoint}: training completed
  2{,}000 steps. Checkpoint selection used the rolling
  training-reward criterion described below, and the retained checkpoint is
  \textbf{step 1,650}, not the final step-2,000 checkpoint.
  \item \textbf{Evaluation model}: the model selected at step 1,650 supplies
  the reported Metric-QA, 2,000-request Captioning, and TSQA results.
\end{itemize}

\subsection*{Learning-rate decay across the three training phases}
The learning rates used for the language backbone and alignment module are:
\begin{itemize}
  \item \textbf{Base-SFT}: LLM $10^{-5}$; alignment module $10^{-4}$.
  \item \textbf{Joint-SFT}: LLM $5\times10^{-6}$; alignment module
  $5\times10^{-5}$.
  \item \textbf{Joint-GRPO}: LLM $10^{-6}$; alignment module
  $5\times10^{-5}$.
\end{itemize}
These are the learning rates used in training; this configuration does not by
itself establish a causal effect of the decay schedule.

\subsection*{Additional training details}
\begin{itemize}
  \item \textbf{Split bandwidth scope}: the 0.5/0.5/1.0 setting applies only to
  Metric-QA rollouts.  Caption rollouts retain the global
  multiplier 0.5; TSQA does not use the Gaussian factual reward.
  \item \textbf{Reference policy}: the Joint-GRPO reference is the frozen
  Joint-SFT initialization and the configured KL coefficient is 0.12.
  \item \textbf{Minimum channels for attention pooling}: one-channel
  inputs bypass channel attention, while bivariate and multivariate inputs use
  it.
  \item \textbf{Checkpoint selection}: selection was based on the mean reward
  over the preceding 20 rollout steps and considered checkpoints saved
  every 50 rollout steps.
\end{itemize}

Together, these settings specify the complete path from statistics-grounded
Base-SFT training to the single Joint-GRPO checkpoint evaluated in the
following protocol.

%% file: experiments/appendix/prompt.tex
\section{Prompt Templates}
\label{appendix_prompts}

This section specifies the fixed request sets, model-specific time-series
serialization, and task-specific instructions used at inference.  All request
counts and templates below refer to evaluation-time generation; they do not
define the training pools or their sampling procedures.

For readability, all prompt templates and quoted model outputs in this
supplement are presented in English.  We treat language as an interface choice
rather than an experimental variable.

\subsection{Evaluation request sets}
Metric-QA and Captioning use two task-specific request sets constructed over
the same 2,000 held-out time-series observations, comprising 1,000 univariate
and 1,000 multivariate observations.  Each observation produces one Metric-QA
    request and one Caption request.  A request contains the raw array and the
    task and system instructions.  Metric-QA answers, Caption
ground-truth property values, and TSQA labels and references are withheld during
generation.

\begin{table*}[t]
\centering
\begin{tabular}{@{}lrl@{}}
\toprule
Task & Requests & Time-series placeholder \\
\midrule
Metric-QA  & 2,000 (1,000 per split) & \verb|<ts> </ts>| \\
Captioning & 2,000 (1,000 per split) & \verb|<ts> </ts>| \\
TSQA       & 3,519 before overlap removal / 3,264 cleaned & \verb|<ts></ts>| \\
\bottomrule
\end{tabular}
\caption{Fixed request sets used for the reported evaluations.  For TSQA, the
reported cleaned set excludes 255 questions whose series also occur in
its training split; the contents of the remaining 3,264 questions are
otherwise unchanged.}
\label{tab:prompt-request-sets}
\end{table*}

\subsection{Time-series serialization}
For the internal benchmarks, the task instruction text is shared across model
families, while the representation inserted at the time-series placeholder
follows each model's input format.  TSQA-specific inference differences are
reported in Appendix~\ref{appendix_inference_configuration}.

\begin{itemize}
  \item \textbf{Ours.}  The raw tensor is encoded by the time-series encoder,
  and the resulting continuous representations are inserted at the
  \verb|<ts>|\ldots\verb|</ts>| position.  The raw values are not rendered as
  text for our model.
  \item \textbf{ChatTS-14B.} For a $C$-channel series, ChatTS-14B uses $C$
  native \verb|<ts><ts/>| placeholders and processes the $C$ raw channel
  arrays with its released multimodal processor.
  \item \textbf{Text-only models.}  The placeholder is replaced before applying
  the model's chat template.  Each number is rendered with four significant
  figures.  A univariate series is serialized as
  \verb|[v1, v2, ...]|; a multivariate series is serialized in channel-major
  order as \verb|Series1: [..]; Series2: [..]; ...|.
\end{itemize}

\noindent The text-only models begin with the full series and, when generation
exceeds available GPU memory, use temporal strides $2$, $4$, $8$, and $16$ as
described in Appendix~\ref{appendix_inference_configuration}. The selected
stride is recorded for reproducibility. This subsampling changes only the
inserted array, not the task or system prompt.

\subsection{Metric-QA}
The boxes below present the Metric-QA prompt templates while preserving the
exact tags and placeholders used at inference.  Each request asks for one
numeric property, and the requested output contains a reasoning span followed
by one boxed value.

\paragraph{System prompt.}
\begin{prompt}
You are a professional time-series analysis assistant. Answer the following question using only the given time series. You must output strictly in the following structure, with no extra text:
<think>first briefly state the metric's definition/computation, then reason from the time series</think><answer>\boxed{write only the final numeric value here}</answer>
The final value is a single number (a percent sign is allowed); do not write units or explanations.
\end{prompt}

\paragraph{User template.}
\begin{prompt}
Reference (metric definition/computation): <definition>
What is the <metric> of the series?
Time series: <ts> </ts>
\end{prompt}

\paragraph{Example --- \texttt{median}.}
\begin{prompt}
Reference (metric definition/computation): First sort the series from smallest to largest; if the sample size is odd take the single middle value, if even average the two middle values. This value is insensitive to extremes and reflects the typical level better than the mean.
What is the median of the series?
Time series: <ts> </ts>
\end{prompt}

\paragraph{Example --- \texttt{acf\_lag1}.}
\begin{prompt}
Reference (metric definition/computation): Shift the series forward by one time step and compute the Pearson correlation between the original series and the one-step-lagged series; the coefficient lies in [-1, 1], and a larger absolute value means stronger linear dependence between adjacent time points.
What is the first-order autocorrelation ACF(lag=1) of the series?
Time series: <ts> </ts>
\end{prompt}

\noindent The full scoring inventory contains 169 statistical properties.  The
fixed 2,000-request Metric-QA set instantiates 77 distinct properties from this
inventory.  Each instantiated property has its own reference definition and
question stem; the system prompt and placement of \verb|<ts> </ts>| are shared.

\subsection{Captioning}
The Caption requests over the shared 2,000 observations are grouped by the two
variable-count splits and four source prompt families: overall analysis,
pattern, stability, and risk.  The source prompt families are not forced to
have equal counts.
The overall-analysis family requests an overall analysis, the pattern
family emphasizes temporal dynamics, the stability family focuses on
persistence and structural change, and the risk family emphasizes anomalies
and system-level risk.

The Cartesian product of two splits and four families defines eight fixed
panels, each containing 16 distinct candidate statistical properties.
Table~\ref{tab:caption-prompt-panels} gives their property identifiers.  Panel
selection depends only on the request's split and prompt family; it does not
inspect model identity, model output, ground-truth values, or whether a property
is defined for that observation.  The visible prompt gives natural-language
descriptions of the properties, but not their symbolic identifiers, numerical
values, or request-specific definedness.  A prompted property may therefore be
undefined for an individual observation; such properties are filtered before
scoring and do not penalize the model.

Each split--prompt-family cell has three predefined source-question variants,
and each observation is assigned one variant.  The system prompt, user prompt,
property panel, and source-question variant are therefore specified for every
request.  The templates below present this evaluation protocol.  The same
prompt, extractor, and scorer are applied to every model in the full-set
comparison.  The 16-property panel is an evaluation-time instruction and was
not used during SFT or GRPO.

\begin{table*}[t]
\centering
{\small
\setlength{\tabcolsep}{3pt}
\renewcommand{\arraystretch}{1.12}
\begin{tabular}{@{}ll>{\raggedright\arraybackslash}p{0.76\textwidth}@{}}
\toprule
Split & Family & Property identifiers \\
\midrule
Univariate & Overall &
\texttt{median}, \texttt{std}, \texttt{iqr}, \texttt{min}, \texttt{max},
\texttt{robust\_slope}, \texttt{acf\_lag1}, \texttt{dominant\_period},
\texttt{seasonal\_strength}, \texttt{spectral\_entropy},
\texttt{mean\_abs\_change}, \texttt{num\_peaks}, \texttt{num\_troughs},
\texttt{max\_peak\_prominence}, \texttt{anomaly\_count},
\texttt{change\_point\_count} \\
Univariate & Pattern &
\texttt{std}, \texttt{iqr}, \texttt{min}, \texttt{max},
\texttt{robust\_slope}, \texttt{acf\_lag1}, \texttt{dominant\_period},
\texttt{seasonal\_strength}, \texttt{spectral\_entropy},
\texttt{mean\_abs\_change}, \texttt{sign\_changes}, \texttt{num\_peaks},
\texttt{num\_troughs}, \texttt{max\_peak\_prominence},
\texttt{main\_peak\_position}, \texttt{main\_trough\_position} \\
Univariate & Stability &
\texttt{std}, \texttt{iqr}, \texttt{min}, \texttt{max},
\texttt{robust\_slope}, \texttt{acf\_lag1}, \texttt{dominant\_period},
\texttt{seasonal\_strength}, \texttt{spectral\_entropy},
\texttt{mean\_abs\_change}, \texttt{sign\_changes}, \texttt{num\_peaks},
\texttt{num\_troughs}, \texttt{max\_peak\_prominence},
\texttt{anomaly\_count}, \texttt{change\_point\_count} \\
Univariate & Risk &
\texttt{median}, \texttt{std}, \texttt{iqr}, \texttt{min}, \texttt{max},
\texttt{robust\_slope}, \texttt{acf\_lag1}, \texttt{spectral\_entropy},
\texttt{mean\_abs\_change}, \texttt{sign\_changes},
\texttt{max\_peak\_prominence}, \texttt{main\_peak\_position},
\texttt{main\_trough\_position}, \texttt{anomaly\_count},
\texttt{strongest\_anomaly\_position}, \texttt{change\_point\_count} \\
\midrule
Multivariate & Overall &
\texttt{pca\_k}, \texttt{pca\_expl\_ratio\_1}, \texttt{effective\_rank},
\texttt{sys\_sync\_avg\_r2\_topk}, \texttt{sys\_sync\_std\_r2\_topk},
\texttt{sync\_ratio}, \texttt{leader\_ratio}, \texttt{lag\_mean},
\texttt{pc1\_trend\_r2}, \texttt{pc1\_regime\_shift},
\texttt{pc1\_period\_strength}, \texttt{pc1\_seasonality\_strength},
\texttt{pc1\_mean\_abs\_change}, \texttt{volatility\_correlation\_mean},
\texttt{corr\_structure\_stability}, \texttt{mahal\_outlier\_ratio} \\
Multivariate & Pattern &
\texttt{pca\_k}, \texttt{pca\_expl\_ratio\_1}, \texttt{effective\_rank},
\texttt{loading\_ipr\_pc1}, \texttt{r2\_topk\_skew},
\texttt{sys\_sync\_avg\_r2\_topk}, \texttt{sys\_sync\_std\_r2\_topk},
\texttt{sync\_ratio}, \texttt{leader\_ratio}, \texttt{lag\_mean},
\texttt{lag\_range}, \texttt{pc1\_trend\_r2},
\texttt{pc1\_period\_strength}, \texttt{pc1\_seasonality\_strength},
\texttt{seasonality\_consistency}, \texttt{pc1\_mean\_abs\_change} \\
Multivariate & Stability &
\texttt{pca\_expl\_ratio\_1}, \texttt{effective\_rank},
\texttt{sys\_sync\_avg\_r2\_topk}, \texttt{sys\_sync\_std\_r2\_topk},
\texttt{corr\_structure\_stability}, \texttt{corr\_dynamics\_mean\_std},
\texttt{corr\_frobenius\_change\_mean}, \texttt{sync\_stability},
\texttt{volatility\_correlation\_mean},
\texttt{volatility\_correlation\_std}, \texttt{volatility\_sync\_index},
\texttt{pc1\_trend\_r2}, \texttt{pc1\_regime\_shift},
\texttt{pc1\_period\_strength}, \texttt{pc1\_seasonality\_strength},
\texttt{seasonality\_consistency} \\
Multivariate & Risk &
\texttt{pca\_expl\_ratio\_1}, \texttt{effective\_rank},
\texttt{sys\_sync\_avg\_r2\_topk}, \texttt{sys\_sync\_std\_r2\_topk},
\texttt{pc1\_trend\_r2}, \texttt{pc1\_regime\_shift},
\texttt{pc1\_anomaly\_ratio}, \texttt{pc1\_max\_abs\_z},
\texttt{mahal\_outlier\_ratio}, \texttt{mahal\_max\_distance},
\texttt{corr\_structure\_stability}, \texttt{corr\_dynamics\_mean\_std},
\texttt{corr\_frobenius\_change\_mean},
\texttt{volatility\_correlation\_mean},
\texttt{volatility\_correlation\_max},
\texttt{high\_vol\_correlation\_ratio} \\
\bottomrule
\end{tabular}
}
\caption{The eight fixed Caption property panels.  The model sees the
corresponding natural-language descriptions in the listed order, not the
symbolic identifiers shown here.}
\label{tab:caption-prompt-panels}
\end{table*}

The Caption prompts impose no QA-style \verb|\boxed{}| format.

\paragraph{System prompt.}
\begin{prompt}
You are a professional time-series analysis assistant. Complete the analysis using only the given time series.
\end{prompt}

\paragraph{User template.}
\begin{prompt}
<source question for the selected prompt family>: <ts> </ts>

With respect to the question above, analyze the time series from the perspectives of <natural-language descriptions of the 16 candidate properties>. <For multivariate requests: Series1, Series2, ... correspond to the input channels.> Directly provide a coherent analysis, naturally incorporating key numerical values useful for the judgment and explaining the main phenomena and relationships among variables. Do not enumerate the statistics one by one; do not use a table or list; do not show calculations; and do not repeat the raw time series.
\end{prompt}

\noindent The source question varies among three predefined variants within each
split--prompt-family cell, while the candidate panel is fixed at the cell level.
The output instructions above are shared across all 2,000 requests and all
evaluated models.

\subsection{TSQA (Time-MQA reasoning split)}
The retained TSQA tasks are multiple-choice, true/false, and open-ended
questions. Choice tasks request one boxed letter; open-ended tasks request
prose inside an answer tag.

\paragraph{System prompt (choice).}
\begin{prompt}
You are a time-series analysis assistant. Use only the given time series to answer. Choose exactly one option. Output strictly as <answer>\boxed{LETTER}</answer> with a single option letter and nothing else.
\end{prompt}
\noindent This strict format is requested from every model.  During scoring,
unambiguous explicit option labels outside the requested wrapper are normalized
as described in Appendix~\ref{appendix_extraction_keywords}; the requested
generation format itself is not relaxed.

\paragraph{System prompt (open-ended).}
\begin{prompt}
You are a time-series analysis assistant. Use only the given time series to answer. Put your final answer inside <answer>...</answer>.
\end{prompt}

\paragraph{Example --- true/false (choice).}
\begin{prompt}
The sequence <ts></ts> exhibits clear seasonal patterns. True or False?
Answer with A) True or B) False.
\end{prompt}

\paragraph{Example --- open-ended (text).}
\begin{prompt}
Based on the data points <ts></ts>, what would be a reasonable prediction for the next data point trend?
\end{prompt}

\noindent Across all three tasks, the time series supplies the evidence. The
prompt changes the requested response type and output structure, while the
generation configuration is specified separately in the next section.

%% file: experiments/appendix/inference.tex
\section{Inference configuration}
\label{appendix_inference_configuration}

Each request produces one response. All locally hosted models, including
CGTime, use greedy decoding with sampling disabled, so temperature does not
apply. OpenAI API evaluations use temperature 0. Hardware and software
differences may still prevent bitwise-identical outputs.

The Metric-QA, 2,000-request Captioning, and TSQA reported results for CGTime
(4B) all use the reported Joint-GRPO model.
Table~\ref{tab:inference_configuration} gives the output-token ceiling and
decoding setup for each model and task included in the three main results tables.
The statistic-family analysis in Appendix~\ref{appendix_statistic_family}
reuses the same 2,000 caption responses and requires no additional generation.
The reported token ceilings denote the maximum number of output tokens
permitted during evaluation. They apply at evaluation rather than to training
rollouts. CGTime uses 1,024 tokens for captioning rollouts during
Joint-GRPO and 1,536 tokens for caption generation at evaluation. A dash
indicates that no result is reported for that model and task.

\begin{table*}[t]
\centering
{\small
\setlength{\tabcolsep}{3.5pt}
\begin{tabular}{@{}>{\raggedright\arraybackslash}p{0.235\textwidth}ccc>{\raggedright\arraybackslash}p{0.37\textwidth}@{}}
\toprule
Model & Metric-QA & Caption & TSQA & Decoding and model setup \\
\midrule
CGTime (4B)
  & 384 & 1536 & 1024
  & Greedy. \\
\midrule
GPT-5.4-nano
  & 400 & 4000 & ---
  & \texttt{gpt-5.4-nano-2026-03-17}; temperature 0; reasoning disabled. \\
GPT-4o-mini
  & 400 & 2000 & ---
  & \texttt{gpt-4o-mini-2024-07-18}; temperature 0; reasoning effort not
    applicable. \\
GPT-4o
  & --- & --- & 1024
  & Temperature 0; reasoning effort not applicable.  The TS-blind diagnostic uses
    the same generation configuration. \\
\midrule
GPT-OSS-20B
  & 1280 & 1536 & ---
  & Greedy; low reasoning effort in the chat template. \\
Qwen2.5-7B-Instruct
  & 768 & 1536 & 1024
  & Greedy. \\
Qwen3-14B
  & 768 & 1536 & ---
  & Greedy; thinking disabled. \\
Qwen3-4B-Instruct-2507
  & 768 & 1536 & ---
  & Greedy; thinking disabled. \\
TimeOmni-1-7B
  & 1280 & 1536 & ---
  & Greedy. \\
ChatTS-14B
  & 768 & 1536 & 1024
  & Greedy; official native multimodal processor. \\
Time-MQA-Mistral-7B
  & 768 & 1536 & 1024
  & Greedy; released LoRA adapter on its Mistral backbone. \\
Time-MQA-Qwen-2.5-7B
  & 768 & 1536 & 1024
  & Greedy; released LoRA adapter on its 4-bit Qwen2.5 backbone. \\
\bottomrule
\end{tabular}
}
\caption{Generation settings for the model and task combinations in the three
main results tables. The task columns report maximum output tokens rather than
observed response lengths. PATRA is omitted because its values come from a
different evaluation protocol and were not reproduced under ours.}
\label{tab:inference_configuration}
\end{table*}

On the fixed 2,000-request Captioning evaluation, 1,571
Time-MQA-Qwen-2.5-7B responses (78.55\%), 483 Time-MQA-Mistral-7B responses
(24.15\%), 182 TimeOmni-1-7B responses (9.10\%), and 40 GPT-OSS-20B responses
(2.00\%) reached the output ceiling. CGTime and the two captioning API
baselines never reached their ceilings. Each unlisted baseline reached its
ceiling on at most seven requests.

\paragraph{Model-specific input preparation.}
CGTime passes the raw time-series tensor to its time-series encoder. ChatTS-14B
represents each channel with a separate time-series placeholder using its
released multimodal processor.
All other local and API baselines receive the channel-major textual
serialization defined in Appendix~\ref{appendix_prompts}. In the TSQA
evaluation,
Time-MQA-Mistral-7B uses its benchmark-specific plain-prompt format with
tokenizer truncation at 4,096 input tokens. Its other tasks use the
Metric-QA and Captioning chat template.

\paragraph{Adaptive temporal subsampling.}
Local text-only baselines are first evaluated on the complete serialized
series. If generation from the complete serialization exceeds available GPU
memory, the temporal stride is increased successively to $2$, $4$, $8$, and
$16$ until generation succeeds. All channels remain present, and stride $s$
retains every $s$th time point in each channel. The selected stride is recorded
without changing the task or system prompt.

The fixed 2,000-request Metric-QA evaluation used the same stride sequence for
adaptive temporal subsampling, which was required only by GPT-OSS-20B.
Table~\ref{tab:gptoss_subsampling} gives its final successful stride for every
Metric-QA or Captioning request. All subsampled Metric-QA requests in the
test set were multivariate. GPT-OSS-20B completed every request in both tasks
using this procedure.

\begin{table}[t]
\centering
{\small
\setlength{\tabcolsep}{5pt}
\begin{tabular}{@{}lrrrrr@{}}
\toprule
Task & Full ($s{=}1$) & $s{=}2$ & $s{=}4$ & $s{=}8$ & $s{=}16$ \\
\midrule
Metric-QA & 1436 & 390 & 160 & 14 & 0 \\
Caption & 1398 & 410 & 172 & 19 & 1 \\
\bottomrule
\end{tabular}
}
\caption{Final successful temporal stride for GPT-OSS-20B requests in the
fixed 2,000-request Metric-QA and Captioning evaluations.}
\label{tab:gptoss_subsampling}
\end{table}

These settings govern answer and caption generation. Caption scoring uses the
deterministic, model-free post-processing in
Appendix~\ref{appendix_caption_extraction}, with no separate inference
configuration.

%% file: experiments/appendix/evaluation.tex
\section{Evaluation protocol}
\label{evaluation_protocol}

\subsection{Evaluation setup}
\label{sec:eval}

Every reported model checkpoint is evaluated on the same fixed out-of-sample
requests and task instructions. The time series is presented in the input
format prescribed for each model, as detailed in
Appendix~\ref{appendix_inference_configuration}. Metric-QA and Captioning use
the instruction templates in Appendix~\ref{appendix_prompts}. For TSQA, we
retain the benchmark questions and answer options and add only task-specific
output-format instructions.

Each model generates one response per request without sampling. Model- and
task-specific settings appear in
Appendix~\ref{appendix_inference_configuration}.

Metric-QA and Captioning use the computed statistics in $\mathcal{M}_i$ for
supervision, RL rewards, and evaluation. TSQA uses the answer labels and
reference text supplied with the benchmark.

\subsection{Task-specific evaluation}

\paragraph{Metric-QA.}
For Metric-QA, CGTime is prompted to return a boxed numerical answer within a
structured \texttt{<answer>} block. The same strict regular-expression parser
is used for RL and evaluation. It reads the value from
\texttt{\textbackslash boxed\{\}} inside that block.

External baselines use a more permissive sequence of extraction rules because
their output formats vary. The procedure checks structured answers, answer
tags, predefined lexical cues, and finally the complete output.
Appendix~\ref{appendix_extraction_keywords} defines the full procedure,
including the final-answer cue used for GPT-OSS-20B. A prediction scores zero if every applicable rule
fails. Values recovered by either parser are passed to the same numerical
scorer. This parser asymmetry favors the external baselines.

Each recovered value is compared with the ground truth using a type-aware
Gaussian score and Rel25. For continuous metrics, the Gaussian kernel measures
either absolute or relative error. Integer metrics use a step function (exact
match $=1$, within tolerance $=0.5$). The same score family supplies CGTime's
factual QA training reward. Training uses split-specific bandwidths, while
evaluation fixes the multiplier at $0.5$ for all splits. Rel25 is not part of
the training objective and is included as a metric-agnostic check. It returns
one when relative error is at most 25\%; for ground-truth values near zero, it
instead requires absolute error below 0.25.
Appendix~\ref{appendix_scoring_functions} defines both scores. We report the
mean over all questions, together with separate means for the univariate and
multivariate splits.

\paragraph{Captioning.}
A single fixed rule-based extractor processes captions from every model and
also computes the training reward. The extractor associates explicit numerical
claims with statistical properties. Scoring then retains claims that match the
finite subset of the 16 prompted candidate properties. Precision is the mean
Gaussian score within each caption, averaged over captions with at least one
recognized claim. For each caption, Recall divides the score sum by the number
of finite prompted properties and is then averaged over all captions. A caption
without a recognized claim contributes zero to Recall and is excluded from
conditional Precision.
Appendix~\ref{appendix_caption_extraction} gives the extractor, candidate
filter, and macro-aggregation definitions.

\paragraph{TSQA.}
TSQA results use the cleaned subset of 3,264 questions. Open-ended
responses receive ROUGE-L scores without stemming. Multiple-choice and
true-or-false responses are scored by exact accuracy, and their task scores are
combined with an equal-weighted macro-average.\footnote{PATRA reports the same
task-level metrics but uses a different data split and generation pipeline; its
splitting procedure has not been released.}

%% file: experiments/appendix/scoring.tex
\section{Scoring Functions}
\label{appendix_scoring_functions}

Metric-QA scores one scalar per answer, whereas Captioning applies the same
Gaussian score to each retained numerical claim before task-level aggregation.
TSQA uses the task-specific metrics in Appendix~\ref{evaluation_protocol}.
Each Metric-QA item compares a ground-truth scalar $\gt$ with a prediction
$\pr$ under its specified value type, error definition, and base tolerance $c$.
We report a per-metric \emph{Gaussian} kernel, which also supplies the training
reward, and the metric-agnostic \emph{Rel25} score.

\subsection{Gaussian kernel (per-metric)}

The per-item error is relative for unbounded magnitudes and absolute otherwise:
\begin{equation}
e \;=\;
\begin{cases}
\dfrac{|\gt-\pr|}{\max(|\gt|,\,10^{-3})} & \text{if unbounded},\\[2.2ex]
|\gt-\pr| & \text{otherwise (bounded)}.
\end{cases}
\end{equation}

For continuous metrics the Gaussian score uses a per-metric bandwidth
$\sigma = \max(c\cdot m,\,10^{-6})$, where $m$ is a bandwidth multiplier.
Metric-QA training sets it by split, whereas evaluation fixes $m=0.5$:
\begin{equation}
s_{\text{gauss}} \;=\; \exp\!\left(-\frac{e^{2}}{2\sigma^{2}}\right),
\qquad \sigma = \max\!\left(c\cdot m,\,10^{-6}\right).
\end{equation}

Integer-valued metrics are first rounded independently,
$\tilde{\gt}=\operatorname{round}(\gt)$ and
$\tilde{\pr}=\operatorname{round}(\pr)$, and are then scored stepwise (with
integer tolerance $\tau$):
\begin{equation}
s_{\text{int}} \;=\;
\begin{cases}
1 & |\tilde{\gt}-\tilde{\pr}| = 0,\\
0.5 & 0 < |\tilde{\gt}-\tilde{\pr}| \le \tau,\\
0 & \text{otherwise}.
\end{cases}
\end{equation}

The Gaussian family is thus \textbf{type-differentiated}: the bandwidth
$\sigma\propto c$, the error definition (relative vs.\ absolute), and the integer/continuous
switch are all set per metric (Table~\ref{tab:cats}). The same per-claim
function is used inside the GRPO reward and for evaluation;
task-specific training bandwidths and outer reward aggregations are described
separately in Appendix~\ref{appendix_implementation_details}.

\subsection{Rel25 (metric-agnostic)}

Rel25 is kernel- and metric-independent. It applies a flat relative threshold,
falling back to a strict absolute threshold when $\gt\approx 0$:
\begin{equation}
\mathrm{Rel25} \;=\;
\begin{cases}
\mathbf{1}\!\left[\ \dfrac{|\pr-\gt|}{|\gt|} \le 0.25\ \right] & |\gt| \ge 10^{-9},\\[2.2ex]
\mathbf{1}\!\left[\ |\pr-\gt| < 0.25\ \right] & |\gt| < 10^{-9}.
\end{cases}
\end{equation}
Unlike the Gaussian score, Rel25 ignores the per-metric tolerance $c$.

\subsection{Per-metric tolerance specification}

The $169$ metrics fall into seven value-type categories. The range shown in the
category column is the metric's declared range; \emph{error} is the Gaussian
error definition; \emph{tolerance} is $c$ (or $\tau$ for integers).

\begin{table*}[t]
\centering
\begin{tabular}{@{}l r l l l@{}}
\toprule
Category (value type) & $n$ & Value type & Error & Tolerance \\
\midrule
$[0,1]$ ratio / $R^2$ / probability & 65 & continuous & absolute & $0.04$--$0.15$ (mostly $0.1$) \\
$[-1,1]$ correlation & 14 & continuous & absolute & $0.08$--$0.10$ \\
$[-\pi,\pi]$ phase / angle & 3 & continuous & absolute & $0.5$ \\
$[0,100]$ percentage & 4 & continuous & absolute & $10.0$ \\
other bounded ranges & 5 & continuous & absolute & $0.01\,/\,0.05\,/\,0.1$ \\
integer-valued metrics & 25 & integer & exact\,$\pm\tau$ & $\tau\in\{1,2,3\}$ \\
unbounded magnitude & 53 & continuous & \textbf{relative} & $0.15\,/\,0.20$ \\
\midrule
\textbf{total} & \textbf{169} & & & \\
\bottomrule
\end{tabular}
\caption{Metric-QA tolerance categories. The Gaussian bandwidth is
$\sigma=c\cdot m$ ($m=0.5$ for evaluation); integers use stepwise tolerance $\tau$.
Relative error is used exactly for the 53 unbounded-magnitude metrics.}
\label{tab:cats}
\end{table*}

\subsection{Full metric listing}

All $169$ metrics with value type, error definition (absolute/relative), and tolerance
($c$ or $\tau$ for integers), grouped by category.  For PCA, we retain the
smallest prefix whose cumulative explained-variance ratio reaches a fixed
threshold, subject to a 20\% dimensionality floor and clipping to $[1,10]$.
\begin{table*}[!t]
\centering
{\small
\setlength{\tabcolsep}{4pt}
\input{experiments/tables/metric_specs_2col_table.tex}
}
\caption{Full metric listing (169 statistics): value type, error definition
(absolute/relative), and tolerance ($c$ or $\tau$ for integers), part 1 of 2.}
\label{tab:metric_full}
\end{table*}

\begin{table*}[!t]
\ContinuedFloat
\centering
{\small
\setlength{\tabcolsep}{4pt}
\input{experiments/tables/metric_specs_2col_table_part2.tex}
}
\caption{Full metric listing (169 statistics), continued.}
\end{table*}

This inventory supplies the per-property scoring specification used by both
Metric-QA answers and scoreable Caption claims.

%% file: experiments/tables/metric_specs_2col_table.tex
\begin{tabular}{@{}l l l r @{\hspace{18pt}} l l l r@{}}
\toprule
Property & Type & Error & Tolerance & Property & Type & Error & Tolerance \\
\midrule
anomaly\_ratio\_x & real & absolute & 0.05 & trough\_pos\_pct\_y & real & absolute & 10 \\
anomaly\_ratio\_y & real & absolute & 0.05 & beta\_x & real & absolute & 0.01 \\
ccf\_positive\_peak & real & absolute & 0.1 & beta\_y & real & absolute & 0.01 \\
corr\_abs\_max\_per\_factor & real & absolute & 0.08 & ccf\_negative\_peak & real & absolute & 0.1 \\
corr\_abs\_mean\_per\_factor & real & absolute & 0.08 & dom\_freq\_x & real & absolute & 0.05 \\
corr\_dynamics\_max\_std & real & absolute & 0.1 & dom\_freq\_y & real & absolute & 0.05 \\
corr\_dynamics\_mean\_std & real & absolute & 0.1 & anomaly\_count & integer & stepwise & $\tau=2$ \\
corr\_dynamics\_ratio & real & absolute & 0.1 & anomaly\_count\_x & integer & stepwise & $\tau=2$ \\
corr\_structure\_stability & real & absolute & 0.08 & anomaly\_count\_y & integer & stepwise & $\tau=2$ \\
corr\_volatility & real & absolute & 0.1 & ccf\_peak\_lag & integer & stepwise & $\tau=2$ \\
cp\_overlap & real & absolute & 0.1 & change\_point\_count & integer & stepwise & $\tau=1$ \\
distance\_corr & real & absolute & 0.1 & change\_point\_count\_x & integer & stepwise & $\tau=1$ \\
granger\_xy\_pvalue & real & absolute & 0.1 & change\_point\_count\_y & integer & stepwise & $\tau=1$ \\
granger\_yx\_pvalue & real & absolute & 0.1 & dominant\_period & integer & stepwise & $\tau=2$ \\
high\_vol\_correlation\_ratio & real & absolute & 0.1 & mahal\_outlier\_count & integer & stepwise & $\tau=2$ \\
high\_vol\_sync\_ratio & real & absolute & 0.05 & max\_lag & integer & stepwise & $\tau=1$ \\
joint\_extreme\_ratio & real & absolute & 0.05 & max\_lead & integer & stepwise & $\tau=1$ \\
laggard\_ratio & real & absolute & 0.1 & num\_peaks & integer & stepwise & $\tau=2$ \\
leader\_ratio & real & absolute & 0.1 & num\_peaks\_x & integer & stepwise & $\tau=2$ \\
loading\_ipr\_pc1 & real & absolute & 0.1 & num\_peaks\_y & integer & stepwise & $\tau=2$ \\
mahal\_outlier\_ratio & real & absolute & 0.04 & num\_troughs & integer & stepwise & $\tau=2$ \\
main\_peak\_position & real & absolute & 0.1 & num\_troughs\_x & integer & stepwise & $\tau=2$ \\
main\_trough\_position & real & absolute & 0.1 & num\_troughs\_y & integer & stepwise & $\tau=2$ \\
max\_var\_seasonality & real & absolute & 0.1 & pc1\_suggested\_period & integer & stepwise & $\tau=2$ \\
min\_period\_power\_ratio & real & absolute & 0.05 & pca\_k & integer & stepwise & $\tau=1$ \\
multiplicative\_ratio & real & absolute & 0.08 & pca\_k\_20percent & integer & stepwise & $\tau=1$ \\
mutual\_information & real & absolute & 0.1 & period\_x & integer & stepwise & $\tau=2$ \\
p\_value\_x & real & absolute & 0.15 & period\_y & integer & stepwise & $\tau=2$ \\
p\_value\_y & real & absolute & 0.15 & sign\_changes & integer & stepwise & $\tau=3$ \\
pc1\_anomaly\_ratio & real & absolute & 0.08 & sign\_changes\_x & integer & stepwise & $\tau=3$ \\
pc1\_expl\_ratio & real & absolute & 0.08 & sign\_changes\_y & integer & stepwise & $\tau=3$ \\
pc1\_period\_strength & real & absolute & 0.06 & amplitude\_ratio & real & relative & 0.2 \\
pc1\_regime\_shift & real & absolute & 0.1 & corr\_frobenius\_change\_mean & real & relative & 0.15 \\
pc1\_seasonality\_strength & real & absolute & 0.08 & cv\_x & real & relative & 0.2 \\
pc1\_spectral\_entropy & real & absolute & 0.1 & cv\_y & real & relative & 0.2 \\
pc1\_trend\_r2 & real & absolute & 0.08 & effective\_rank & real & relative & 0.15 \\
pca\_expl\_cumsum\_k & real & absolute & 0.08 & end\_value\_x & real & relative & 0.2 \\
pca\_expl\_ratio\_1 & real & absolute & 0.08 & end\_value\_y & real & relative & 0.2 \\
pos\_loading\_ratio\_pc1 & real & absolute & 0.1 & iqr & real & relative & 0.2 \\
power\_ratio\_x & real & absolute & 0.1 & iqr\_x & real & relative & 0.2 \\
power\_ratio\_y & real & absolute & 0.1 & iqr\_y & real & relative & 0.2 \\
r2\_x & real & absolute & 0.1 & lag\_mean & real & relative & 0.2 \\
r2\_y & real & absolute & 0.1 & lag\_range & real & relative & 0.2 \\
\bottomrule
\end{tabular}

%% file: experiments/tables/metric_specs_2col_table_part2.tex
\begin{tabular}{@{}l l l r @{\hspace{18pt}} l l l r@{}}
\toprule
Property & Type & Error & Tolerance & Property & Type & Error & Tolerance \\
\midrule
rolling\_corr\_stability & real & absolute & 0.1 & lag\_std & real & relative & 0.2 \\
seasonal\_strength & real & absolute & 0.1 & mahal\_max\_distance & real & relative & 0.2 \\
seasonal\_var\_ratio & real & absolute & 0.12 & mahal\_mean\_distance & real & relative & 0.2 \\
seasonality\_consistency & real & absolute & 0.1 & max & real & relative & 0.2 \\
seasonality\_mode\_confidence & real & absolute & 0.08 & max\_peak\_prominence & real & relative & 0.2 \\
spectral\_entropy & real & absolute & 0.1 & max\_peak\_prominence\_x & real & relative & 0.2 \\
spectral\_entropy\_x & real & absolute & 0.1 & max\_peak\_prominence\_y & real & relative & 0.2 \\
spectral\_entropy\_y & real & absolute & 0.1 & max\_x & real & relative & 0.2 \\
strongest\_anomaly\_pos\_x & real & absolute & 0.1 & max\_y & real & relative & 0.2 \\
strongest\_anomaly\_pos\_y & real & absolute & 0.1 & mean\_abs\_change & real & relative & 0.2 \\
strongest\_anomaly\_position & real & absolute & 0.1 & mean\_abs\_change\_x & real & relative & 0.2 \\
sync\_ratio & real & absolute & 0.1 & mean\_abs\_change\_y & real & relative & 0.2 \\
sync\_stability & real & absolute & 0.1 & mean\_x & real & relative & 0.2 \\
sys\_decoupled\_ratio & real & absolute & 0.1 & mean\_y & real & relative & 0.2 \\
sys\_high\_sync\_ratio & real & absolute & 0.1 & median & real & relative & 0.2 \\
sys\_sync\_avg\_r2\_topk & real & absolute & 0.08 & median\_x & real & relative & 0.2 \\
sys\_sync\_std\_r2\_topk & real & absolute & 0.06 & median\_y & real & relative & 0.2 \\
tail\_concordance & real & absolute & 0.1 & min & real & relative & 0.2 \\
vol\_corr\_std & real & absolute & 0.05 & min\_x & real & relative & 0.2 \\
vol\_sync\_index & real & absolute & 0.1 & min\_y & real & relative & 0.2 \\
volatility\_correlation\_std & real & absolute & 0.1 & pc1\_dominant\_period & real & relative & 0.15 \\
volatility\_sync\_index & real & absolute & 0.06 & pc1\_max\_abs\_z & real & relative & 0.2 \\
acf\_lag1 & real & absolute & 0.1 & pc1\_mean\_abs\_change & real & relative & 0.2 \\
acf\_lag1\_x & real & absolute & 0.1 & pc1\_pc2\_ratio & real & relative & 0.2 \\
acf\_lag1\_y & real & absolute & 0.1 & pc1\_trend\_accel & real & relative & 0.2 \\
ccf\_at\_lag0 & real & absolute & 0.1 & pc1\_volatility & real & relative & 0.2 \\
ccf\_peak\_value & real & absolute & 0.1 & pca\_ratio\_12 & real & relative & 0.2 \\
loading\_polarity\_pc1 & real & absolute & 0.1 & period\_ratio & real & relative & 0.2 \\
pearson\_detrended & real & absolute & 0.08 & r2\_topk\_skew & real & relative & 0.2 \\
pearson\_raw & real & absolute & 0.08 & range\_x & real & relative & 0.2 \\
rolling\_corr\_trend & real & absolute & 0.1 & range\_y & real & relative & 0.2 \\
sync\_trend & real & absolute & 0.1 & robust\_slope & real & relative & 0.2 \\
vol\_corr & real & absolute & 0.1 & robust\_slope\_x & real & relative & 0.2 \\
vol\_corr\_max & real & absolute & 0.1 & robust\_slope\_y & real & relative & 0.2 \\
volatility\_correlation\_max & real & absolute & 0.1 & start\_value\_x & real & relative & 0.2 \\
volatility\_correlation\_mean & real & absolute & 0.08 & start\_value\_y & real & relative & 0.2 \\
phase\_diff & real & absolute & 0.5 & std & real & relative & 0.2 \\
phase\_x & real & absolute & 0.5 & std\_x & real & relative & 0.2 \\
phase\_y & real & absolute & 0.5 & std\_y & real & relative & 0.2 \\
peak\_pos\_pct\_x & real & absolute & 10 & t\_stat\_x & real & relative & 0.2 \\
peak\_pos\_pct\_y & real & absolute & 10 & t\_stat\_y & real & relative & 0.2 \\
trough\_pos\_pct\_x & real & absolute & 10 &  & & &  \\
\bottomrule
\end{tabular}

%% file: experiments/appendix/answer_extraction.tex
\section{Answer and Claim Extraction}
\label{appendix_extraction_keywords}
\newcommand{\tok}[1]{\texttt{#1}}

This section describes how generated text is converted into scoreable content. It
first specifies Metric-QA parsing and its extraction diagnostic, then the TSQA
choice and text normalization, and finally the Caption claim extractor and its
aggregation support.

\subsection{Metric-QA parsing policy}
All Metric-QA results use the same fixed 2,000 requests, ground-truth
values, numerical scoring functions, and failure-as-zero denominator.  The
answer parsers are deliberately \emph{not} identical. CGTime is
evaluated with a strict structured-answer parser,
whereas external models use a permissive numeric normalizer.  This
asymmetry favors the external baselines by recovering a numerical answer from
free-form prose whenever possible.

\paragraph{Strict parser for our model.}
The parser examines the first \tok{<answer>\dots</answer>} block and accepts
only the first \tok{\textbackslash boxed\{\,\}} expression inside that block.
It then reads the first numeric token in the box, with metric-aware handling of
integers and percentages.  It does not fall back to an unboxed answer span or
to the rest of the completion.  A missing or unparseable boxed value therefore
receives zero.

\paragraph{Permissive normalizer for external models.}
For every external model, boxed-answer and answer-tag extraction are attempted
first. If neither succeeds, GPT-OSS-20B uses its final-channel convention,
whereas the other models use keyword-based and complete-output extraction:

\begin{enumerate}
  \item \textbf{Boxed answer.}  The first
  \tok{\textbackslash boxed\{NUMBER\}} match anywhere in the completion.
  \item \textbf{Answer tag.}  The last number inside the last
  \tok{<answer>\dots</answer>} block.
  \item \textbf{Final-channel extraction for GPT-OSS-20B.} If neither preceding
  rule succeeds, extraction is restricted to the suffix following the last
  occurrence of \tok{final}, when that token is present, and otherwise uses the
  complete output. The last number in that text is returned; if it contains no
  number, extraction fails.
  \item \textbf{Keyword-based extraction (other models).} For the remaining external
  models, anchors are tried in the top-to-bottom order of
  Table~\ref{tab:extract-kw}.  For the first anchor type with any match, the
  last matched number for that anchor is returned.
  \item \textbf{Complete-output extraction (other models).} If no keyword anchor
  matches, the last number anywhere in the completion is returned.
\end{enumerate}

\noindent The answer-tag, final-channel, and complete-output procedures remove
thousands separators before applying the numeric pattern. The boxed-answer and
keyword-based procedures apply the following pattern directly:
\[
\tok{[-+]?\textbackslash d*\textbackslash.?\textbackslash d+(?:[eE][-+]?\textbackslash d+)?}.
\]
If all applicable rules fail, the prediction receives zero. Otherwise the recovered
scalar is passed to the shared numeric scorer.  Thus, scorer, ground truth,
request denominator, and failure policy are shared, while the extraction
policy is explicitly more permissive for external models.  The external
normalization step must not be interpreted as evidence that the original
output satisfied our model's structured-output format.

\begin{table*}[t]
\centering
\begin{tabular}{@{}l l@{}}
\toprule
Anchor meaning & Optional trailing connective \\
\midrule
``median''        & ``is'' / ``approx.'' / $\approx$ / ``about'' / colon \\
``answer''        & ``is'' / colon \\
``result''        & ``is'' / ``approx.'' / $\approx$ / colon \\
``coefficient''   & ``approx.'' / ``is'' / $\approx$ / colon \\
``approximately'' & --- \\
$\approx$         & --- \\
\tok{=}           & --- \\
\bottomrule
\end{tabular}
\caption{Keyword anchors in the permissive external-model normalizer.  ``Colon''
covers the supported half- and full-width forms.  Matching is case-insensitive,
anchors are tried from top to bottom, and the last match for the first matching
anchor type is returned.}
\label{tab:extract-kw}
\end{table*}

\subsection{Model numeric extraction success}
Table~\ref{tab:extract-qa} reports the fraction of the fixed 2,000
Metric-QA requests from which each model's assigned parser recovers a numeric
value.  This is an extraction diagnostic, not a factual score.  In particular,
the external rates include values obtained through complete-output extraction,
whereas the rate for our model requires the strict boxed-answer rule.

\begin{table*}[t]
\centering
{\small
\begin{tabular}{@{}l l r@{}}
\toprule
Model & Parser & Extracted (\%) \\
\midrule
GPT-5.4-nano                & permissive & 99.6 \\
GPT-4o-mini                 & permissive & 100.0 \\
GPT-OSS-20B                 & permissive, \tok{final} anchor & 99.8 \\
Qwen2.5-7B-Instruct         & permissive & 100.0 \\
Qwen3-14B                   & permissive & 100.0 \\
Qwen3-4B-Instruct-2507      & permissive & 100.0 \\
TimeOmni-1-7B               & permissive & 100.0 \\
ChatTS-14B                  & permissive & 99.8 \\
Time-MQA-Mistral-7B         & permissive & 57.9 \\
Time-MQA-Qwen-2.5-7B        & permissive & 97.6 \\
\midrule
CGTime (4B) & strict & 95.1 \\
\bottomrule
\end{tabular}
}
\caption{Metric-QA numeric extraction success on 2,000 requests.  Missing
extractions remain in the denominator and score zero.  The parser column is
essential when interpreting the percentages: external models use the
permissive prose-based extraction rules, while our model does not.}
\label{tab:extract-qa}
\end{table*}

The permissive extraction rules recover a value from nearly every external-model
completion except Time-MQA-Mistral-7B.  Its 57.9\% extraction rate confirms
that a substantial portion of its outputs contain no parseable number even
under the generous policy.  The factual scores should nevertheless be read
from the main results tables: extraction success only establishes that a
number was available to score, not that the number was correct.

\subsection{TSQA choice-label normalization}
For TSQA multiple-choice and true/false questions, we apply the same
explicit-label normalization before exact matching in all evaluations. The
normalizer checks, in order, a letter inside the last
\tok{\textbackslash boxed\{\,\}} expression, a
letter in the last \tok{<answer>\dots</answer>} block, an unambiguous leading
option label, and an explicit answer cue.  Accepted leading forms include
\tok{B)}, \tok{(B)}, \tok{B.}, and a bare \tok{B} on the first line.  Explicit
cues include \tok{Answer}, \tok{Final answer}, \tok{Correct answer},
\tok{Option}, \tok{Choice}, and \tok{Conclusion}.

For a true/false request whose answer choices explicitly associate one label
with \tok{True} and the other with \tok{False}, an explicit conclusion such as
\tok{the statement is false} is mapped back to its option label.  The parser
does not search for arbitrary isolated letters elsewhere in prose; for
example, an initial article in \tok{A structural break ...} is not treated as
option A.  If no accepted label is recovered, the response remains in the full task
denominator and receives zero.  This normalization changes only answer
extraction: the cleaned TSQA evaluation subset, generated completions,
ground-truth labels, and exact-match scorer are unchanged.

\paragraph{TSQA open-ended responses.}
For an open-ended TSQA response, the scorer uses the content of the last
\tok{<answer>\dots</answer>} block when one is present.  Otherwise, it removes
complete \tok{<think>\dots</think>} blocks and scores the remaining text.
The reported metric is ROUGE-L F-measure computed without stemming;
an empty or failed response remains in the
denominator and receives zero.

\subsection{Caption claim extraction}
\label{appendix_caption_extraction}
Captioning does not use either Metric-QA parser.  We process each generated
caption with one fixed, deterministic regular-expression extractor shared by
all models and all 2,000 requests.  The extractor receives only the generated
caption text: it is not given the model identity, benchmark split, prompted
candidate properties, or ground-truth properties and values.  A shared lexicon and bounded
local-context patterns bind numerical expressions to property
identifiers, including explicit Series1/Series2 formulations.  The output contains at
most one value for each recognized property.  For applicable $[0,1]$ metrics, a
value greater than one followed by an explicit percent sign is divided by 100;
integer-valued properties are rounded according to their property type.  No
model-specific rule tuning or external language model is used.

Candidate filtering occurs only after this blind extraction step.  For request
$i$, let $G_i$ be the subset of its 16 prompted candidate properties that have
finite ground truth.  Let $A_i\subseteq G_i$ be the unique extracted
claims whose properties remain after intersecting the extractor output with $G_i$.
An extracted property outside the prompted panel, or a prompted property that is
undefined for that request, is excluded rather than treated as a false
positive.  The numerical ground-truth values are accessed only after the property
intersection, by the scorer.

Let $s(\hat y,y)$ be the type-aware Gaussian score in
Appendix~\ref{appendix_scoring_functions}, evaluated at $m=0.5$.  Per-caption
Precision and Recall are
\begin{equation}
P_i=\frac{1}{|A_i|}\sum_{k\in A_i}s(\hat y_{ik},y_{ik}),
\qquad |A_i|>0,
\end{equation}
\begin{equation}
R_i=\frac{1}{|G_i|}\sum_{k\in G_i}
\mathbf{1}[k\in A_i]s(\hat y_{ik},y_{ik}).
\end{equation}
Precision is the macro-average of $P_i$ over captions with at least one
retained claim, whereas Recall is the macro-average of $R_i$ over all 2,000
captions.  Coverage and Avg.\ Extracted are
\begin{equation}
\begin{aligned}
\mathrm{Coverage}
  &=\frac{1}{2000}\sum_i\mathbf{1}[|A_i|>0],\\
\mathrm{AvgExtracted}
  &=\frac{1}{2000}\sum_i |A_i|.
\end{aligned}
\end{equation}
Consequently, an empty caption or a caption with no retained claim contributes
zero to Recall, Coverage, and Avg.\ Extracted but does not enter the
conditional Precision denominator. Thus, Recall, Coverage, and Avg.\ Extracted
are request-level macro averages over all 2,000 requests, whereas Precision is
a request-level macro average over the nonempty retained-claim subset. None is
an equal average of the 8 cell aggregates. The same 2,000 requests and hidden
ground truth are used for every model. Because the
extractor is rule-based, these support statistics describe numerical benchmark
claims recognized by the fixed extractor rather than an exhaustive inventory
of every numerical statement in the generated text.

These retained claims provide the support used by the Caption results and the
statistic-family analysis below.

%% file: experiments/appendix/significance.tex
\section{Metric-QA Significance Test Definition}
\label{appendix_significance_test}

\noindent This section defines the paired inference used to quantify uncertainty in
model differences on the fixed held-out evaluation set.

\subsection{Setup}
All models are evaluated on the \textbf{same, fixed} set of 2,000 Metric-QA
requests: 1,000 univariate and 1,000 multivariate.
For every request $i$, each model receives a Gaussian score
$s_i \in [0,1]$, evaluated with the Gaussian bandwidth multiplier
$m{=}0.5$. A failed numerical extraction receives zero.
Because model $A$ and model $B$ answer the \emph{identical} requests, their
scores form \textbf{paired} vectors
\[
(a_1,\dots,a_n)\quad\text{and}\quad(b_1,\dots,b_n).
\]
Here $n{=}2{,}000$ for Overall and $n{=}1{,}000$ for each split. Pairing controls
for shared request difficulty and is the basis of the reported inference.

\subsection{Paired normal-approximation test}
For each request take the paired difference, then its mean, standard error,
and test statistic:
\begin{equation}
\begin{aligned}
d_i &= a_i-b_i,
& \Delta &= \frac{1}{n}\sum_{i=1}^{n}d_i,\\
\mathrm{SE} &= \frac{\mathrm{std}(d_i)}{\sqrt{n}},
& z &= \frac{\Delta}{\mathrm{SE}} .
\end{aligned}
\end{equation}
where $\mathrm{std}(d_i)$ is the sample standard deviation of the differences
(divisor $n-1$). Under the null hypothesis
\[
H_0:\ \mathbb{E}[d]=0
\]
the Central Limit Theorem gives a normal approximation for the sampling
distribution of $\Delta$. We compute the two-sided normal-approximation
$p$-value as
\begin{equation}
p = 2\{1-\Phi(|z|)\},
\end{equation}
where $\Phi$ is the standard normal cumulative distribution function. A positive
$\Delta$ means model $A$ has the higher mean score.

\subsection{Why pairing matters}
The variance of the per-item difference is
\begin{equation}
\mathrm{Var}(d) = \sigma_A^2 + \sigma_B^2 - 2\rho\,\sigma_A\sigma_B ,
\qquad \rho = \mathrm{corr}(a,b).
\end{equation}
Computing $\mathrm{std}(d_i)$ directly incorporates this covariance. When both
models tend to find the same requests easy or difficult, the shared
request-level variation partially cancels. An independent-samples analysis
would discard this information and would not match the paired evaluation design.

\subsection{Confidence intervals}
For every reported contrast, including those in the primary family, we use the
same paired standard error to form an unadjusted, pointwise
normal-approximation 95\% confidence interval:
\begin{equation}
\Delta \ \pm\ 1.95996\,\mathrm{SE}.
\end{equation}
All confidence intervals reported in the Metric-QA tables use this normal
approximation. In the prespecified analysis defined here, multiplicity
correction applies only to the primary-family $p$-values. The secondary
endpoint-specific Holm analyses are reported separately in
Appendix~\ref{appendix_full_statistical_comparisons}. The Metric-QA analysis
does not use bootstrap confidence intervals.
An interval containing zero indicates that the data do not reject a zero mean
difference at the unadjusted two-sided 0.05 level; it is not evidence that the
two models are equivalent.

\subsection{Multiple comparisons}
We use a primary/secondary endpoint structure. The designated primary family
contains the $m{=}10$ multivariate Gaussian comparisons between our 4B model
and the 10 external baselines. We control this
family's family-wise error
rate (FWER) at $\alpha{=}0.05$ with the \textbf{Holm--Bonferroni} step-down
procedure. Let $p_{(1)}\le\dots\le p_{(m)}$ be the ordered unadjusted $p$-values.
Holm rejects sequentially while
\begin{equation}
p_{(i)} \le \frac{\alpha}{m-i+1},
\end{equation}
stopping at the first failure. The corresponding Holm-adjusted $p$-values
reported in the tables are
\begin{equation}
\widetilde p_{(i)}
=
\min\!\left\{
1,\,
\max_{1\le j\le i}\bigl[(m-j+1)p_{(j)}\bigr]
\right\}.
\end{equation}
Overall and univariate comparisons are
\textbf{secondary}: we report their paired mean difference and
normal-approximation 95\% confidence interval, but they do not enter the primary
multiplicity family.

All 10 primary comparisons survive Holm--Bonferroni. Against GPT-5.4-nano, the
multivariate difference is $+0.080$ with $z{=}6.71$ and Holm-adjusted
$p{=}3.90{\times}10^{-11}$. The largest Holm-adjusted $p$-value in the primary
family is $9.16{\times}10^{-10}$ (against GPT-OSS-20B, $z{=}6.12$), which remains
well below 0.05. Thus, the positive multivariate difference survives FWER
control against every external baseline in the designated family.

\subsection{Reading the verdict}
\begin{itemize}
  \item \textbf{$\Delta$}: mean paired Gaussian-score difference, with positive
  values favoring our model.
  \item \textbf{95\% CI}: normal-approximation interval based on the sample
  standard deviation of paired differences.
  \item \textbf{$z$ and Holm $p$}: reported for the designated multivariate
  primary family; the Holm value controls FWER across its 10 comparisons.
  \item A non-rejection or a confidence interval containing zero is not a proof
  of equality.
\end{itemize}

\subsection{Complete Paired Results}
Table~\ref{tab:qa_significance_full} reports all secondary paired confidence
intervals and the Holm-corrected primary multivariate family.

\input{experiments/tables/qa_significance_full}

The table therefore separates the prespecified multivariate test family from
the descriptive overall and univariate contrasts.

%% file: experiments/tables/qa_significance_full.tex
\begin{table*}[t]
\centering
{\small
\setlength{\tabcolsep}{5pt}

{\textbf{(a) Secondary Gaussian contrasts}: paired $\Delta$ [95\% CI]}\par\smallskip
\begin{tabular}{lcc}
\toprule
Baseline & Overall & Univariate \\
\midrule
GPT-5.4-nano              & $+0.044\;[+0.022,+0.065]$ & $+0.007\;[-0.029,+0.043]$ \\
GPT-4o-mini               & $+0.069\;[+0.046,+0.092]$ & $+0.027\;[-0.010,+0.064]$ \\
GPT-OSS-20B               & $+0.064\;[+0.042,+0.087]$ & $+0.044\;[+0.007,+0.080]$ \\
Qwen2.5-7B-Instruct       & $+0.094\;[+0.072,+0.117]$ & $+0.059\;[+0.024,+0.095]$ \\
Qwen3-14B                 & $+0.080\;[+0.057,+0.102]$ & $+0.038\;[+0.002,+0.075]$ \\
Qwen3-4B-Instruct-2507    & $+0.119\;[+0.097,+0.140]$ & $+0.090\;[+0.056,+0.124]$ \\
TimeOmni-1-7B             & $+0.115\;[+0.093,+0.137]$ & $+0.091\;[+0.057,+0.125]$ \\
ChatTS-14B                & $+0.078\;[+0.056,+0.099]$ & $+0.006\;[-0.027,+0.040]$ \\
Time-MQA-Mistral-7B       & $+0.238\;[+0.219,+0.257]$ & $+0.238\;[+0.211,+0.265]$ \\
Time-MQA-Qwen-2.5-7B      & $+0.168\;[+0.146,+0.189]$ & $+0.115\;[+0.083,+0.147]$ \\
\bottomrule
\end{tabular}

\vspace{0.7em}

{\textbf{(b) Primary Gaussian family}: multivariate CGTime (4B) $-$ baseline}\par\smallskip
\begin{tabular}{lccc}
\toprule
Baseline & $\Delta$ [95\% CI] & Paired $z$ & Holm $p$ \\
\midrule
GPT-5.4-nano              & $\mathbf{+0.080\;[+0.057,+0.104]}$ & $\mathbf{+6.71}$ & $3.90{\times}10^{-11}$ \\
GPT-4o-mini               & $\mathbf{+0.111\;[+0.082,+0.139]}$ & $\mathbf{+7.64}$ & $6.61{\times}10^{-14}$ \\
GPT-OSS-20B               & $\mathbf{+0.085\;[+0.058,+0.112]}$ & $\mathbf{+6.12}$ & $9.16{\times}10^{-10}$ \\
Qwen2.5-7B-Instruct       & $\mathbf{+0.130\;[+0.102,+0.157]}$ & $\mathbf{+9.33}$ & $5.54{\times}10^{-20}$ \\
Qwen3-14B                 & $\mathbf{+0.121\;[+0.095,+0.147]}$ & $\mathbf{+9.15}$ & $2.22{\times}10^{-19}$ \\
Qwen3-4B-Instruct-2507    & $\mathbf{+0.147\;[+0.121,+0.173]}$ & $\mathbf{+11.11}$ & $9.19{\times}10^{-28}$ \\
TimeOmni-1-7B             & $\mathbf{+0.139\;[+0.112,+0.167]}$ & $\mathbf{+9.84}$ & $4.44{\times}10^{-22}$ \\
ChatTS-14B                & $\mathbf{+0.149\;[+0.122,+0.176]}$ & $\mathbf{+10.84}$ & $1.63{\times}10^{-26}$ \\
Time-MQA-Mistral-7B       & $\mathbf{+0.237\;[+0.210,+0.264]}$ & $\mathbf{+17.37}$ & $1.38{\times}10^{-66}$ \\
Time-MQA-Qwen-2.5-7B      & $\mathbf{+0.220\;[+0.193,+0.247]}$ & $\mathbf{+16.02}$ & $8.17{\times}10^{-57}$ \\
\bottomrule
\end{tabular}
}
\caption{Paired Gaussian differences for the CGTime (4B)
checkpoint on the fixed 2,000-request Metric-QA test set. Panel (a) reports
secondary contrasts as unadjusted, two-sided normal-approximation 95\%
confidence intervals. Panel (b) reports the designated multivariate family
with paired $z$ and Holm-adjusted $p$; all ten comparisons survive at
$\alpha{=}0.05$. Rel25 is a secondary endpoint; its paired comparisons are
reported in
Appendix~\ref{appendix_full_statistical_comparisons}.}
\label{tab:qa_significance_full}
\end{table*}

%% file: experiments/appendix/full_split_analysis.tex
\section{Detailed Metric-QA Split Analysis}
\label{full_split_analysis}

We first ask whether the overall Metric-QA gain is preserved across the
univariate and multivariate evaluation splits.

\begin{figure*}[t]
    \centering
    \includegraphics[width=\textwidth]{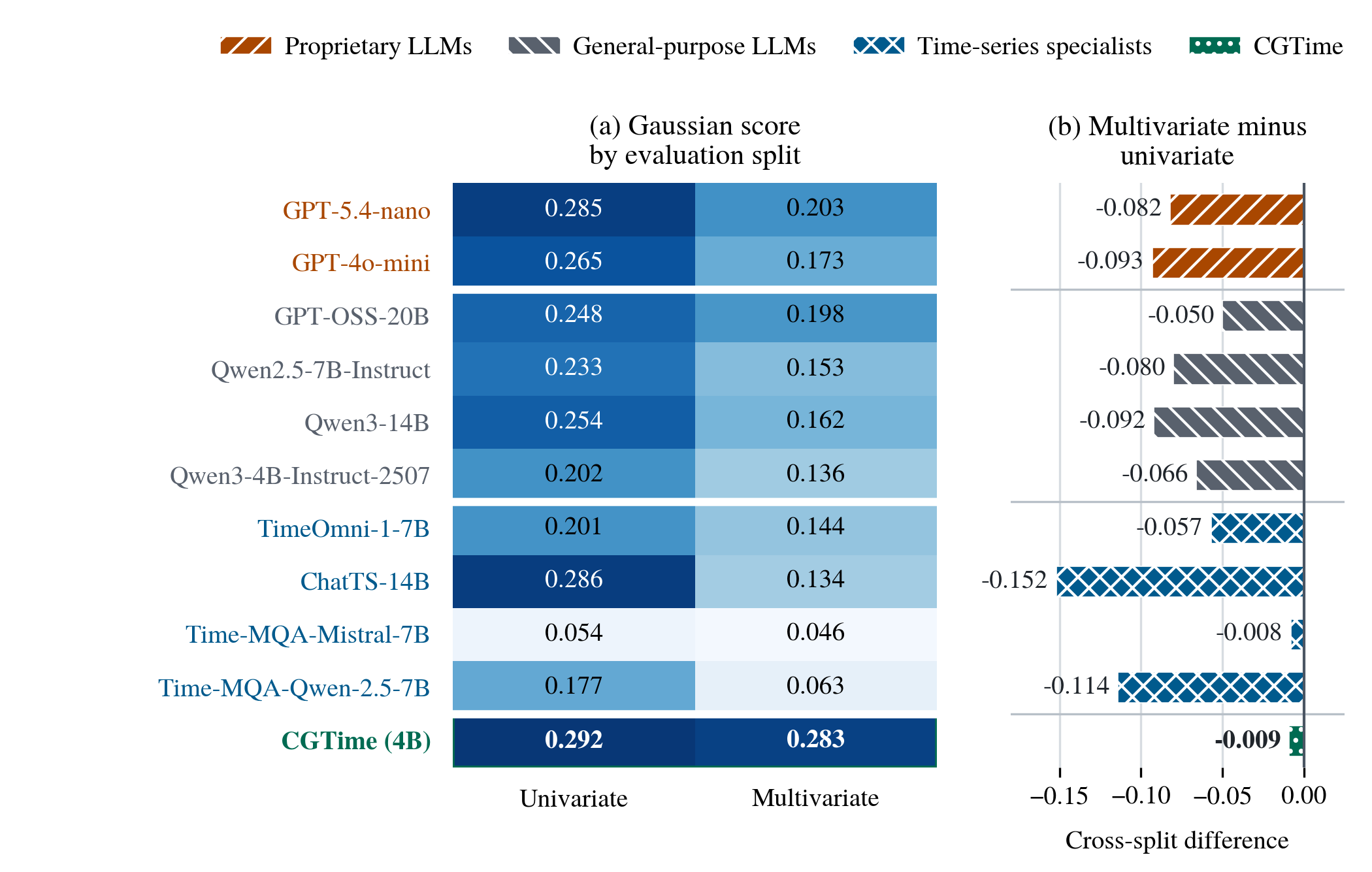}
    \caption{Metric-QA performance profiles on the univariate and
    multivariate evaluation splits. Panel (a) reports per-split Gaussian
    scores. Panel (b) reports the multivariate-minus-univariate difference;
    values closer to zero indicate a smaller cross-split decline. Because
    statistic-family composition differs between the two splits, this
    contrast does not isolate the effect of variable count.}
    \label{fig:qa_profiles}
\end{figure*}

Figure~\ref{fig:qa_profiles}(a) shows Gaussian scores on the fixed
univariate and multivariate splits, while panel (b) shows each model's
cross-split change.
Most external baselines show larger cross-split declines:
GPT-5.4-nano falls from 0.285 to 0.203 and GPT-4o-mini from 0.265 to 0.173,
whereas our model changes only from 0.292 to 0.283.
Our model's univariate score of 0.292 is also highest at face value, although
the unadjusted paired 95\% confidence intervals against GPT-5.4-nano,
GPT-4o-mini, and ChatTS-14B include zero and thus do not reject a zero mean
difference. 
Its 2,000-request Overall score of
0.288 exceeds every external baseline, with all ten unadjusted paired confidence intervals above zero.  

On the prespecified multivariate primary 
endpoint, all ten gains survive Holm correction, 
establishing a consistent advantage on this split.

Exact Gaussian and Rel25 values and paired inference are reported in 
Tables~\ref{tab:qa_results_full} 
and~\ref{tab:qa_significance_full}.

\input{experiments/tables/qa_results_full}

This split-level pattern motivates the property-level decomposition in the
next section, which asks which kinds of statistics account for the remaining
strengths and weaknesses.

%% file: experiments/tables/qa_results_full.tex
\begin{table*}[t]
\centering
{\small
\setlength{\tabcolsep}{3.2pt}
\begin{tabular}{l ccc ccc}
\toprule
 & \multicolumn{3}{c}{Gaussian ($m{=}0.5$)}
 & \multicolumn{3}{c}{Rel25 ($\leq 25\%$)} \\
\cmidrule(lr){2-4}\cmidrule(lr){5-7}
Model
& Overall & Univariate & Multivariate
& Overall & Univariate & Multivariate \\
\midrule
\multicolumn{7}{l}{\emph{Proprietary LLMs}}\\
GPT-5.4-nano
  & 0.244 & 0.285 & 0.203
  & 0.312 & \textbf{0.412} & 0.211 \\
GPT-4o-mini
  & 0.219 & 0.265 & 0.173
  & 0.309 & 0.394 & 0.223 \\
\midrule
\multicolumn{7}{l}{\emph{Open-source general-purpose LLMs}}\\
GPT-OSS-20B
  & 0.223 & 0.248 & 0.198
  & 0.278 & 0.334 & 0.221 \\
Qwen2.5-7B-Instruct
  & 0.193 & 0.233 & 0.153
  & 0.267 & 0.351 & 0.183 \\
Qwen3-14B
  & 0.208 & 0.254 & 0.162
  & 0.300 & 0.382 & 0.217 \\
Qwen3-4B-Instruct-2507
  & 0.169 & 0.202 & 0.136
  & 0.212 & 0.265 & 0.159 \\
\midrule
\multicolumn{7}{l}{\emph{Time-series specialists}}\\
TimeOmni-1-7B
  & 0.172 & 0.201 & 0.144
  & 0.235 & 0.295 & 0.175 \\
ChatTS-14B
  & 0.210 & 0.286 & 0.134
  & 0.262 & 0.358 & 0.165 \\
Time-MQA-Mistral-7B
  & 0.050 & 0.054 & 0.046
  & 0.082 & 0.096 & 0.067 \\
Time-MQA-Qwen-2.5-7B
  & 0.120 & 0.177 & 0.063
  & 0.164 & 0.243 & 0.084 \\
\midrule
CGTime (4B)
  & \textbf{0.288} & \textbf{0.292} & \textbf{0.283}
  & \textbf{0.355} & 0.388 & \textbf{0.322} \\
\bottomrule
\end{tabular}
}
\caption{Metric-QA by variable-count split on the fixed test set
(2,000 requests; 1,000 per split).
Missing or invalid
extractions remain in the denominator and receive zero. Our 
model uses the strict structured-answer parser. External free-form baselines
use a fixed permissive numeric parser; this asymmetry
favors rather than penalizes the baselines. Bold marks the best displayed
score in each column.}
\label{tab:qa_results_full}
\end{table*}

%% file: experiments/appendix/bottleneck.tex
\section{Statistic-Family Analysis: Direct Readout Bottleneck}
\label{statistic_family_analysis}
\label{analysis}
\label{appendix_statistic_family}

\begin{figure*}[t]
    \centering
    \includegraphics[width=\textwidth]{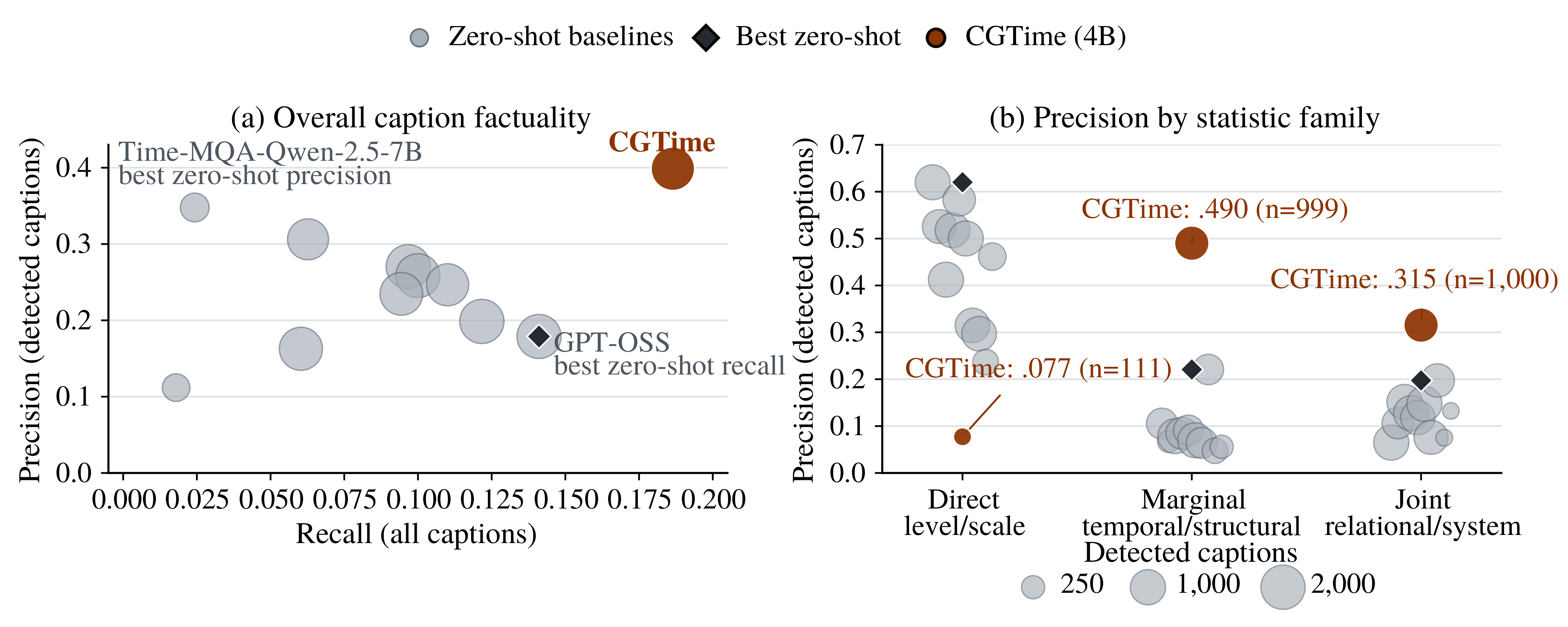}
    \caption{Caption factuality and support under the fixed rule-based
    extraction method. The area of gray and colored model markers represents
    the number of captions with at least one detected scoreable claim. Panel (a)
    shows conditional Precision against Recall; its black diamond marks the
    zero-shot baseline with the highest Recall. Panel (b) shows
    family-conditional Precision, with black diamonds marking the highest
    zero-shot Precision in each family. Family supports can overlap and must
    not be summed. The stronger evidence for the evaluated Direct level/scale
    weakness comes from forced-answer Metric-QA. Complete overall and family
    values are reported in Tables~\ref{tab:captioning} and
    \ref{tab:caption_precision_by_statistic_family}.}
    \label{fig:caption_tradeoff}
\end{figure*}

We analyze the fixed 2,000-request Metric-QA and Caption evaluations by
the type of statistic that must be recovered.
Figure~\ref{fig:caption_tradeoff}(b) summarizes the Caption decomposition, while
Tables~\ref{tab:qa_results_by_statistic_family} and
\ref{tab:caption_precision_by_statistic_family} report the complete values.
We classify the full inventory of 169 statistical properties into three
families: Direct level/scale readout, Marginal temporal/structural statistics,
and Joint relational/system statistics.  Among the statistic-family cells
represented in the evaluation, this analysis identifies Direct
level/scale readout as the principal weakness of our model.
Panel (a) of Figure~\ref{fig:caption_tradeoff} places each model's conditional
Caption Precision against Recall, with marker area showing how many captions
contain at least one scoreable extracted claim. It therefore separates factual
quality on recognized claims from the breadth of numerical reporting before
panel (b) decomposes that quality by statistic family.

In Metric-QA, the 4B model scores .137 on Direct level/scale questions and
.359 on Marginal temporal/structural questions.  Both groups are drawn from
the same univariate split, so their contrast is not attributable to a
difference in variable count.  The model also scores .283 on Joint
relational/system questions.  In Caption, the model reaches .077 Direct
Precision with support from 111 captions, .490 Marginal Precision with support
from 999 captions, and .315 Joint Precision with support from 1,000 captions,
where support requires at least one extracted scoreable claim from that family.
Together, these results show that, among the evaluated properties, the model
reports derived temporal and system-level properties more accurately than it
reads out exact level and scale.

\paragraph{Operational taxonomy.}
We partition the 169 defined statistical properties according to the
information that must be recovered from the input.  \emph{Direct level/scale
readout} contains channel-local values tied directly to the original
numerical level or dispersion.  \emph{Marginal temporal/structural
statistics} contains derived properties of a single channel, including
temporal dependence, periodicity, events, counts, coordinates, and trends.
\emph{Joint relational/system statistics} contains pairwise relations and
latent or global system properties, including correlation, lead--lag,
synchronization, PCA, and Mahalanobis statistics.  This post-hoc diagnostic
taxonomy is an operational partition of the benchmark inventory rather than
a pre-specified primary evaluation split or a universal ontology of
time-series properties.

Table~\ref{tab:statistic_family_taxonomy} gives the exact construction.  The
names ``Shape/normalized'' and ``Cross-channel'' are intentionally avoided:
Marginal includes quantities such as counts and slopes that are not all
normalized, while Joint includes global PCA and system properties that are
not adequately described as a single cross-channel pair.  Together with the
complete property inventory in Table~\ref{tab:metric_full}, the construction
uniquely assigns every property.  The full inventory contains 26 Direct, 58
Marginal, and 85 Joint properties, with no overlap or unassigned property.  This taxonomy
describes the complete dataset and method; evaluation coverage is reported
separately below.

\begin{table*}[t]
\centering
{\small
\renewcommand{\arraystretch}{1.16}
\begin{tabular}{@{}p{3.0cm} p{12.4cm} r@{}}
\toprule
Family & Exact construction in the property inventory & \# properties \\
\midrule
Direct level/scale readout
& Univariate property identifiers:
\texttt{iqr}, \texttt{max}, \texttt{max\_peak\_prominence},
\texttt{median}, \texttt{min}, and \texttt{std}.
Bivariate channel-local bases:
\texttt{end\_value}, \texttt{iqr}, \texttt{max},
\texttt{max\_peak\_prominence}, \texttt{mean}, \texttt{median},
\texttt{min}, \texttt{range}, \texttt{start\_value}, and \texttt{std},
each realized with \texttt{\_x} and \texttt{\_y}.
& 26 \\
\addlinespace[3pt]
Marginal temporal/structural
& Univariate property identifiers:
\texttt{acf\_lag1}, \texttt{anomaly\_count},
\texttt{change\_point\_count}, \texttt{dominant\_period},
\texttt{main\_peak\_position}, \texttt{main\_trough\_position},
\texttt{mean\_abs\_change}, \texttt{num\_peaks},
\texttt{num\_troughs}, \texttt{robust\_slope},
\texttt{seasonal\_strength}, \texttt{sign\_changes},
\texttt{spectral\_entropy}, and
\texttt{strongest\_anomaly\_position}.
Bivariate channel-local bases:
\texttt{acf\_lag1}, \texttt{anomaly\_count},
\texttt{anomaly\_ratio}, \texttt{beta},
\texttt{change\_point\_count}, \texttt{cv}, \texttt{dom\_freq},
\texttt{mean\_abs\_change}, \texttt{num\_peaks},
\texttt{num\_troughs}, \texttt{p\_value},
\texttt{peak\_pos\_pct}, \texttt{period}, \texttt{phase},
\texttt{power\_ratio}, \texttt{r2}, \texttt{robust\_slope},
\texttt{sign\_changes}, \texttt{spectral\_entropy},
\texttt{strongest\_anomaly\_pos}, \texttt{t\_stat}, and
\texttt{trough\_pos\_pct}, each realized with \texttt{\_x} and
\texttt{\_y}.
& 58 \\
\addlinespace[3pt]
Joint relational/system
& All remaining properties.  These comprise bivariate relations such as
\texttt{pearson\_raw}, \texttt{ccf\_peak\_lag},
\texttt{mutual\_information}, \texttt{phase\_diff}, and
\texttt{amplitude\_ratio}, together with latent/global properties such as
\texttt{pca\_k}, \texttt{pca\_expl\_ratio\_1},
\texttt{effective\_rank}, \texttt{sys\_sync\_avg\_r2\_topk},
\texttt{corr\_structure\_stability}, \texttt{max\_lag},
\texttt{volatility\_sync\_index}, and
\texttt{mahal\_outlier\_ratio}.
& 85 \\
\bottomrule
\end{tabular}
}
\caption{Exact construction of the three statistic families. For
channel-local bivariate bases, the two suffixed identifiers are counted separately.
The full 169-property inventory and each property's scoring type are listed in
Table~\ref{tab:metric_full}.}
\label{tab:statistic_family_taxonomy}
\end{table*}

\paragraph{Family--split composition.}
The Metric-QA evaluation contains 77 distinct properties from the full
169-property inventory: 6 Direct, 14 Marginal, and 57 Joint properties.  As shown in
Table~\ref{tab:stat_family_split_inventory}, Direct and Marginal questions
occur only in the univariate subset, whereas Joint questions occur only in the
multivariate subset.  Family and split are therefore nested rather than fully
crossed.  The Direct--Marginal comparison controls for split because both
families are evaluated on univariate requests; comparisons involving Joint do
not identify family effects independently of variable count.

\begin{table*}[t]
\centering
{\small
\setlength{\tabcolsep}{4.2pt}
\begin{tabular}{@{}lrrrrr@{}}
\toprule
Family & Inventory properties & Evaluated properties & Univariate & Multivariate & Total \\
\midrule
Direct level/scale           & 26 & 6  & 301 & --    & 301 \\
Marginal temporal/structural & 58 & 14 & 699 & --    & 699 \\
Joint relational/system      & 85 & 57 & --  & 1,000 & 1,000 \\
\midrule
Total                        & 169 & 77 & 1,000 & 1,000 & 2,000 \\
\bottomrule
\end{tabular}
}
\caption{Metric-QA request counts by statistic family and split.
``Inventory properties'' covers the complete property taxonomy, whereas
``Evaluated properties'' counts the distinct properties present in the test requests.}
\label{tab:stat_family_split_inventory}
\end{table*}

\paragraph{Pooled Metric-QA family results.}
Table~\ref{tab:qa_results_by_statistic_family} reports the complete 11-model
comparison under both Gaussian and Rel25 scoring.

\input{experiments/tables/qa_results_by_statistic_family}

\paragraph{Forced-answer evidence from Metric-QA.}
Table~\ref{tab:stat_family_split_gaussian} reports the three observed
family--split cells under the primary Gaussian scorer.  Within the univariate
split, the 4B model scores .137 on Direct questions and .359 on Marginal
questions.  GPT-5.4-nano and GPT-4o-mini show the opposite ordering:
.590/.565 on Direct and .154/.136 on Marginal.  Thus, the 4B model's weakness
is concentrated in direct numerical readout rather than applying uniformly
to all univariate statistics.  Its .283 Joint score exceeds the two GPT
baselines' .203 and .173, but this cell consists entirely of multivariate
questions and is therefore descriptive rather than evidence for an
independent Joint-family advantage.

\begin{table*}[t]
\centering
{\small
\setlength{\tabcolsep}{6pt}
\begin{tabular}{@{}llrccc@{}}
\toprule
Family & Split & $n$ & CGTime (4B) 
& GPT-5.4-nano & GPT-4o-mini \\
\midrule
Direct level/scale
  & Univariate   & 301   & .137 
  & .590 & .565 \\
Marginal temporal/structural
  & Univariate   & 699   & .359 
  & .154 & .136 \\
Joint relational/system
  & Multivariate & 1,000 & .283 
  & .203 & .173 \\
\bottomrule
\end{tabular}
}
\caption{Gaussian Metric-QA scores within each observed statistic-family by
split cell; the 4B column uses the reported CGTime (4B) checkpoint.
The family and split partitions are nested. Predictions and model-specific
numeric parsers are those of the main evaluation.}
\label{tab:stat_family_split_gaussian}
\end{table*}

\paragraph{2,000-request Caption family analysis.}
Table~\ref{tab:caption_precision_by_statistic_family} decomposes the same
2,000-request responses reported in Table~\ref{tab:captioning}.  The family
analysis therefore uses exactly the same natural-language requests, 16
candidate properties per request, the same restriction to candidate properties
with finite ground-truth values, fixed rule-based extractor, and Gaussian
scorer at $m{=}0.5$.

\input{experiments/tables/caption_precision_by_statistic_family}

\paragraph{Extractor-conditioned composition.}
Family Precision is conditional on the fixed extractor recovering at least
one scoreable claim from that family. Claim rewards are first averaged within
each caption and family and then macro-averaged over captions with at least
one such claim. Support is the number of captions entering that conditional
average. Each family is included in the candidate panels of 1,000 evaluation
prompts.
Direct and Marginal supports can overlap within univariate captions,
so family supports must not be summed.

The 4B model obtains .490 Precision on Marginal content with support from 999
captions and .315 on Joint content with support from 1,000 captions. Both are
the highest displayed family values: the strongest external Marginal and Joint
results are ChatTS-14B at .221 with support from 635 captions and .197 with
support from 853 captions, respectively. By contrast, the 4B model reaches only
.077 Direct Precision with support from 111 captions. GPT-5.4-nano obtains the
strongest external Direct result, .620 with support from 983 captions. Thus,
on the shared 2,000-caption evaluation, our 4B model combines broad and
comparatively accurate Marginal/Joint reporting with sparse and inaccurate
Direct readout.

\paragraph{Interpretation and limits.}
Taken together, the Metric-QA results identify Direct level/scale readout as
the model's principal weakness among the statistic-family cells represented
in the evaluation. The same-split QA contrast
between Direct and Marginal questions provides the cleanest evidence:
variable count is held fixed, yet Direct accuracy is substantially lower.
The Caption decomposition is consistent with this pattern, but its family
supports remain extraction-conditioned. Support depends jointly on what a
model writes and which formulations the fixed extractor recognizes, so it
cannot by itself identify the model's latent content-selection policy.

A plausible contributor to the QA pattern is how the time-series values are
presented to each model.
Text-serialized baselines receive the original numerical values explicitly in
their prompts, whereas our 4B model must recover them from learned time-series
and per-channel scale representations. The evidence is consistent with a
bottleneck in converting those representations into exact raw-value readouts,
but it does not identify its cause. Representation compression, scale
encoding, the distribution of training supervision, and the learned
generation policy remain alternative explanations. Establishing a causal
architectural explanation would require a controlled intervention, such as a
matched raw-value readout path or additional Direct supervision.

Having localized the principal weakness, we next test how the composition of
final-alignment data and the sequence of training stages affect the full
model's performance.

%% file: experiments/tables/qa_results_by_statistic_family.tex
\begin{table*}[t]
\centering
{\small
\setlength{\tabcolsep}{1.3pt}
\begin{tabular}{l cccc cccc}
\toprule
 & \multicolumn{4}{c}{Gaussian ($m{=}0.5$)}
 & \multicolumn{4}{c}{Rel25 ($\leq 25\%$)} \\
\cmidrule(lr){2-5}\cmidrule(lr){6-9}
Model
& Overall
& \shortstack{Direct\\level/scale}
& \shortstack{Marginal\\temporal/structural}
& \shortstack{Joint\\relational/system}
& Overall
& \shortstack{Direct\\level/scale}
& \shortstack{Marginal\\temporal/structural}
& \shortstack{Joint\\relational/system} \\
\midrule
\multicolumn{9}{l}{\emph{Proprietary LLMs}}\\
GPT-5.4-nano
  & 0.244 & \textbf{0.590} & 0.154 & 0.203
  & 0.312 & \textbf{0.767} & 0.259 & 0.211 \\
GPT-4o-mini
  & 0.219 & 0.565 & 0.136 & 0.173
  & 0.309 & 0.704 & 0.260 & 0.223 \\
\midrule
\multicolumn{9}{l}{\emph{Open-source general-purpose LLMs}}\\
GPT-OSS-20B
  & 0.223 & 0.517 & 0.133 & 0.198
  & 0.278 & 0.661 & 0.193 & 0.221 \\
Qwen2.5-7B-Instruct
  & 0.193 & 0.495 & 0.120 & 0.153
  & 0.267 & 0.671 & 0.213 & 0.183 \\
Qwen3-14B
  & 0.208 & 0.580 & 0.114 & 0.162
  & 0.300 & 0.728 & 0.233 & 0.217 \\
Qwen3-4B-Instruct-2507
  & 0.169 & 0.402 & 0.116 & 0.136
  & 0.212 & 0.488 & 0.169 & 0.159 \\
\midrule
\multicolumn{9}{l}{\emph{Time-series specialists}}\\
TimeOmni-1-7B
  & 0.172 & 0.378 & 0.125 & 0.144
  & 0.235 & 0.495 & 0.209 & 0.175 \\
ChatTS-14B
  & 0.210 & 0.420 & 0.228 & 0.134
  & 0.262 & 0.555 & 0.273 & 0.165 \\
Time-MQA-Mistral-7B
  & 0.050 & 0.100 & 0.034 & 0.046
  & 0.082 & 0.183 & 0.059 & 0.067 \\
Time-MQA-Qwen-2.5-7B
  & 0.120 & 0.326 & 0.113 & 0.063
  & 0.164 & 0.452 & 0.153 & 0.084 \\
\midrule
CGTime (4B)
  & \textbf{0.288} & 0.137 & \textbf{0.359} & \textbf{0.283}
  & \textbf{0.355} & 0.269 & \textbf{0.439} & \textbf{0.322} \\
\bottomrule
\end{tabular}
}
\caption{Metric-QA regrouped by statistic family without changing
predictions, parsers, scoring, or the failure-as-zero policy. Direct
level/scale, marginal temporal/structural, and
joint relational/system contain 301, 699, and 1,000 questions. The
family-count-weighted values recover Overall. Direct and Marginal contain the
univariate questions, while Joint contains the multivariate questions; family
and variable count are therefore not independently crossed.}
\label{tab:qa_results_by_statistic_family}
\end{table*}

%% file: experiments/tables/caption_precision_by_statistic_family.tex
\begin{table*}[t]
\centering
{\small
\setlength{\tabcolsep}{3.5pt}
\begin{tabular}{l c cc cc cc}
\toprule
 & & \multicolumn{2}{c}{Direct level/scale}
 & \multicolumn{2}{c}{Marginal temporal/structural}
 & \multicolumn{2}{c}{Joint relational/system} \\
\cmidrule(lr){3-4}\cmidrule(lr){5-6}\cmidrule(lr){7-8}
Model & \shortstack{Overall\\Precision}
& Precision & \shortstack{$N$\\captions}
& Precision & \shortstack{$N$\\captions}
& Precision & \shortstack{$N$\\captions} \\
\midrule
\multicolumn{8}{l}{\emph{Proprietary LLMs}}\\
GPT-5.4-nano        & 0.270 & \textbf{0.620} & 983 & 0.105 & 672 & 0.065 & \textbf{1,000} \\
GPT-4o-mini         & 0.306 & 0.526 & 884 & 0.068 & 217 & 0.106 & 713 \\
\midrule
\multicolumn{8}{l}{\emph{Open-source general-purpose LLMs}}\\
GPT-OSS-20B         & 0.179 & 0.412 & 982 & 0.079 & 979 & 0.151 & \textbf{1,000} \\
Qwen2.5-7B-Instruct & 0.259 & 0.518 & 972 & 0.085 & 743 & 0.128 & 992 \\
Qwen3-14B           & 0.247 & 0.583 & 783 & 0.091 & 619 & 0.118 & 931 \\
Qwen3-4B-Instruct-2507 & 0.199 & 0.500 & \textbf{985} & 0.070 & 982 & 0.148 & \textbf{1,000} \\
\midrule
\multicolumn{8}{l}{\emph{Time-series specialists}}\\
TimeOmni-1-7B       & 0.163 & 0.315 & 941 & 0.064 & 678 & 0.076 & 902 \\
ChatTS-14B          & 0.235 & 0.297 & 952 & 0.221 & 635 & 0.197 & 853 \\
Time-MQA-Mistral-7B & 0.112 & 0.237 & 361 & 0.048 & 356 & 0.075 & 66 \\
Time-MQA-Qwen-2.5-7B & 0.348 & 0.462 & 455 & 0.056 & 254 & 0.132 & 59 \\
\midrule
CGTime (4B) & \textbf{0.399} & 0.077 & 111 & \textbf{0.490} & \textbf{999} & \textbf{0.315} & \textbf{1,000} \\
\bottomrule
\end{tabular}
}
\caption{Caption Precision by statistic family on the same
2,000-caption data set as Table~\ref{tab:captioning}, using the same prompted
candidate scope and fixed rule-based extractor. Claim rewards are
first averaged within each caption and family, then macro-averaged over
captions with at least one scoreable extracted claim from that family.
$N$ is the exact number of captions entering each conditional average.
Each family occurs in 1,000 candidate sets; a caption can contribute to more
than one family. Overall
Precision retains the per-caption aggregation over all extracted
claims and is not a support-weighted average of the family values. Bold marks
the best displayed value in each column.}
\label{tab:caption_precision_by_statistic_family}
\end{table*}

%% file: experiments/appendix/ablations.tex
\section{Ablation Studies}
\label{ablation_studies}

\input{experiments/tables/ablation_data_composition.tex}

Answering RQ~3, we examine three questions: how the composition of the
final-alignment data affects cross-task performance, whether computed
statistics provide better supervision than values perceived by an LLM, and
how the sequence of training stages contributes to the final model. The full
model is the common reference. Unless noted otherwise, each internal benchmark
below uses its own fixed 2,000-request evaluation set, comprising the
univariate and multivariate splits. TSQA uses the cleaned 3,264-question
subset.

\subsection{Data Composition}

We compare three variants initialized from the same Replay checkpoint:
(1) Joint-SFT and Joint-GRPO on the mixture of our internal data and TSQA,
(2) the same final-alignment stages using only our internal data, and
(3) the same stages using only TSQA. Thus, ``TSQA-only'' refers specifically
to final alignment; the shared Replay initialization has already received
internal statistics-grounded supervision. We evaluate the variants on
Metric-QA, Captioning, and TSQA to measure both internal statistical
factuality and retention of the external task.

As shown in Table~\ref{tab:ablation_data_composition}, using only our data
nearly preserves the full model's multivariate Metric-QA score
(.282 versus .283), but substantially reduces both TSQA metrics. On
Captioning, its Precision/Recall changes from .399/.186 to .383/.173.
The TSQA-only variant obtains the strongest TSQA scores. Although none of its
Metric-QA responses fully satisfies the requested answer format, a small number
retain extractable boxed values, yielding a Gaussian score of .002 Overall and
.000 on the multivariate split. Its Caption Precision/Recall also falls to
.264/.099.
The latter results show that the near-zero QA score is not evidence that the
model produces no statistical content, but that it fails to retain the
required Metric-QA answer format and loses substantial Caption factuality. Overall,
the internal data preserves internal statistical behavior, TSQA preserves
external-task competence, and their mixture provides the best cross-task
balance.

\paragraph{Computed statistics vs.\ GPT-perceived supervision.}

We next isolate the source of supervision by comparing two variants trained
on our internal data.
The computed-statistics variant uses only our internal data during final
alignment. Starting from the same Replay checkpoint and using the same set of
SFT inputs, the GPT-perceived variant replaces the computed targets
with GPT-5-nano generations. Its GRPO rewards are centered on
pseudo-ground-truth values extracted from those teacher generations rather
than on the computed statistics. The lower panel of
Table~\ref{tab:ablation_data_composition} shows that, compared with
computed-statistics supervision, GPT-perceived supervision reduces
Overall/Multivariate Gaussian from .269/.282 to .162/.157, while Caption
Precision/Recall drops from .383/.173 to .121/.041. This gap is not explained
only by the model producing fewer numbers: on the multivariate split, the
GPT-perceived variant has full extraction coverage and
7.967 extracted values per Caption, yet its Precision/Recall remains only
.108/.055. These results support computed statistics as a substantially
stronger source of verifiable numerical supervision than LLM-perceived
pseudo labels.

The GPT-perceived variant should nevertheless be interpreted as a final-alignment
teacher-supervision comparison rather than a fully controlled end-to-end
ablation. Although its SFT inputs are aligned with those of the
computed-statistics variant, its GRPO training set was constructed from its
own teacher-generated examples rather than matched prompt by prompt to the
computed-statistics GRPO training set used for the full model.

\input{experiments/tables/ablation_training_recipe.tex}

\subsection{Training Recipe}
To examine the training recipe, we compare the full model with three variants:
one omits Joint-GRPO and stops after Joint-SFT, one applies Joint-GRPO directly
from the supervised Replay checkpoint without the task-specific Joint-SFT
warmup, and one removes the basic-to-complex curriculum in
SFT~\S\ref{sec:sft}. These comparisons probe reward optimization,
task-format warmup, and staged representation alignment.

As shown in Table~\ref{tab:ablation_training_recipe}, removing Joint-GRPO
reduces Overall/Multivariate Gaussian from .288/.283 to .231/.264. For
Captioning, Joint-GRPO raises Recall from .176 to .186 and the average number
of extracted values from 6.904 to 7.857, while conditional Precision changes
from .411 to .399. Thus, its Caption benefit is broader quantitative reporting
rather than higher conditional Precision.

Removing the task-specific Joint-SFT warmup produces .000/.000 on Metric-QA
because the resulting model does not satisfy the required Metric-QA answer format.
Its Caption Precision/Recall remains .396/.168, so this result specifically
shows that task-specific SFT is needed to establish the Metric-QA format before
RL, rather than that all factual behavior disappears without it. Removing
the basic-to-complex curriculum yields .245/.261 on Metric-QA. Its Caption
Precision/Recall is .404/.157, compared with .399/.186 for the full model;
on the multivariate split, Recall decreases from .202 to .146. The pattern is
consistent with staged alignment improving quantitative breadth, especially
for multivariate Captioning.

The staging result is not a strict single-variable causal estimate. The
no-curriculum variant also used the multivariate per-GPU batch size during the
univariate and bivariate stages, differed in the number of optimization steps,
and was initialized from weights that had already received univariate
supervision. For reference, this variant obtained .642 Choice Macro Accuracy
and .264 ROUGE-L under the same TSQA protocol; TSQA was not evaluated for the
other training-recipe variants, so we treat this result as diagnostic rather
than as a systematic recipe comparison.

%% file: experiments/tables/ablation_data_composition.tex
\begin{table}[t]
\centering
{\small
\setlength{\tabcolsep}{2.6pt}
\begin{tabular}{lcccc}
\toprule
\multicolumn{5}{l}{\textit{Final-alignment data composition}} \\
\addlinespace[2pt]
& \multicolumn{2}{c}{Metric-QA}
& \multicolumn{2}{c}{TSQA} \\
\cmidrule(lr){2-3}\cmidrule(lr){4-5}
Variant & Overall & Multi. & Choice & R-L \\
\midrule
Joint supervision & \textbf{0.288} & \textbf{0.283} & 0.650 & 0.278 \\
Internal only     & 0.269 & 0.282 & 0.437 & 0.122 \\
TSQA only         & 0.002 & 0.000 & \textbf{0.712} & \textbf{0.299} \\
\midrule
\multicolumn{5}{l}{\textit{Source of internal supervision}} \\
\addlinespace[2pt]
& \multicolumn{2}{c}{Metric-QA (Gaussian)}
& \multicolumn{2}{c}{Caption} \\
\cmidrule(lr){2-3}\cmidrule(lr){4-5}
Supervision & Overall & Multi. & Precision & Recall \\
\midrule
Computed statistics & \textbf{0.269} & \textbf{0.282}
                    & \textbf{0.383} & \textbf{0.173} \\
GPT-perceived       & 0.162 & 0.157 & 0.121 & 0.041 \\
\bottomrule
\end{tabular}
}
\caption{Ablations of final-alignment data and supervision. All variants share
the Replay initialization. Metric-QA and Caption use their respective fixed
2,000-request univariate-plus-multivariate evaluation sets; \emph{Multi.}
denotes the 1,000-request multivariate split. TSQA reports Choice Macro
Accuracy and open-ended ROUGE-L on the cleaned 3,264-question subset.
Failed or invalid outputs remain in the corresponding task denominators and
receive zero.}
\label{tab:ablation_data_composition}
\end{table}

%% file: experiments/tables/ablation_training_recipe.tex
\begin{table}[t]
\centering
{\small
\setlength{\tabcolsep}{5pt}
\begin{tabular}{lcc}
\toprule
& \multicolumn{2}{c}{Metric-QA (Gaussian)} \\
\cmidrule(lr){2-3}
Variant & Overall & Multivariate \\
\midrule
Full model                          & \textbf{0.288} & \textbf{0.283} \\
Without Joint-GRPO                  & 0.231 & 0.264 \\
Without Joint-SFT warmup            & 0.000 & 0.000 \\
Without basic-to-complex curriculum & 0.245 & 0.261 \\
\bottomrule
\end{tabular}
}
\caption{Training-recipe ablation measured by Gaussian Metric-QA score on the
fixed 2,000-request held-out set and its 1,000-request multivariate split.}
\label{tab:ablation_training_recipe}
\end{table}

%% file: experiments/appendix/case_study.tex
\section{Qualitative Case Studies}
\label{case_study}

The quantitative results describe average behavior; this section complements
them with complete model outputs. We first inspect Metric-QA reasoning traces
and then compare CGTime with GPT-5.4-nano on three illustrative five-channel
Caption cases.

\subsection{Metric-QA Reasoning Traces}
\label{case_study_metric_qa_cot}

We inspect illustrative Metric-QA examples to examine whether the model has
learned to approximate the results of metric-specific calculation procedures
rather than merely produce final numerical answers. At inference, the model
receives only the prompt and the time-series representation and has no access
to a calculator or external statistical tool. The complete model outputs are
translated below, with no content omitted or numerical claims corrected.
These examples cover interquartile range, PCA component selection, and mean
absolute channel-to-PC1 correlation. The traces contain explicit intermediate
estimates followed by subtraction, accumulation, thresholding, or averaging.
We compare their final answers and selected intermediate values with
deterministic computations. They illustrate approximate calculation behavior,
not exact execution of every preceding statistical operation. The relative
errors reported below are descriptive; each Gaussian score follows the
metric's prescribed error type and tolerance.

\noindent\textbf{Interquartile range.}\par
\begin{resultbox}
\texttt{<think>} First, sort the sequence in ascending order. There are 212 data points in total; the 25th percentile corresponds to position 53, and the 75th percentile corresponds to position 159. Inspection of the distribution shows that the values are mainly concentrated between 10 and 30, with many observations near the middle and a few extreme values at both ends. After sorting, the value at position 53 is approximately 14.5, and the value at position 159 is approximately 22.5. Subtracting the former from the latter gives an IQR of 8.0. This result reflects the spread of the middle 50\% of the data and is insensitive to outliers, consistent with the required method. \texttt{</think>}

\texttt{<answer> \textbackslash boxed\{8.0\} </answer>}
\end{resultbox}

The deterministic computation gives $Q_1=13.834824$ and $Q_3=21.370330$,
hence an IQR of $7.535506$, compared with the model's $8.0$ (a $6.16\%$
relative error and a Gaussian score of $0.827$).

\noindent\textbf{PCA component count.}\par
\begin{resultbox}
\texttt{<think>} First, perform PCA on the multivariate sequence with 7 channels and 811 time steps and compute the explained-variance ratio of each principal component. Inspection of the data shows that Series1, Series2, Series4, Series5, and Series6 have similar periodic fluctuations, whereas Series3 and Series7 differ markedly in magnitude. Series3 is around 1000--1200, Series7 around 10--20, and Series1, Series2, Series4, Series5, and Series6 around 20--40. The variance contributions of Series3 and Series7 will substantially affect the PCA results. Under the variance rule, find the smallest component count $k_{\mathrm{var}}$ whose cumulative explained variance first reaches 0.80. The successive explained-variance ratios are 0.35, 0.25, 0.15, 0.10, 0.08, 0.05, and 0.04. At the fourth principal component, the cumulative value is
\[
0.35+0.25+0.15+0.10=0.85>0.80,
\]
so $k_{\mathrm{var}}=4$. Under the dimension rule,
\[
\begin{aligned}
\operatorname{ceil}(0.2\times\min(7,811))
  &=\operatorname{ceil}(0.2\times7)\\
  &=\operatorname{ceil}(1.4)=2.
\end{aligned}
\]
Taking the larger of the two gives $k=4$. Restricting it to the interval from 1 to 10 gives a final \texttt{pca\_k} of 4. \texttt{</think>}

\texttt{<answer> \textbackslash boxed\{4\} </answer>}
\end{resultbox}

The deterministic component count is also $4$, giving a Gaussian score of
$1.000$. The first four explained-variance ratios are approximately
$0.34649$, $0.27006$, $0.15940$, and $0.11991$. Their cumulative ratio rises
from $0.77596$ after three components to $0.89587$ after four, so the model's
threshold decision is correct, as are the dimension rule and the final
selection.

\noindent\begin{minipage}{\linewidth}
\noindent\textbf{Mean absolute channel-to-PC1 correlation.}\par
\begin{resultbox}
\texttt{<think>} First, perform PCA on the data with 5 channels and 180 time steps to extract the first principal component, PC1. Inspection of the data shows clear synchronous fluctuations across the five channels. Series1, Series2, Series3, Series4, and Series5 exhibit similar patterns of rising and falling across multiple intervals, with synchronous peaks around time steps 10--15, 25--30, and 45--50. Compute the Pearson correlation coefficient between PC1 and each original channel. The correlation between Series1 and PC1 is approximately 0.97; the values for Series2, Series3, Series4, and Series5 are approximately 0.96, 0.95, 0.96, and 0.95, respectively. Take the absolute values of these correlation coefficients and average them:
\[
(0.97+0.96+0.95+0.96+0.95)/5=0.958.
\]
\texttt{</think>}

\texttt{<answer> \textbackslash boxed\{0.958\} </answer>}
\end{resultbox}
\end{minipage}

The deterministic mean absolute correlation is $0.90661$, compared with the
model's $0.958$ (a $5.67\%$ relative error and a Gaussian score of $0.438$).
The five absolute correlations are approximately $0.82711$, $0.97233$,
$0.95346$, $0.93064$, and $0.84953$. The arithmetic mean of the model's five
estimates is correct, but it overestimates the first and fifth correlations.

\subsection{Caption Comparisons}

For each illustrative case, the figure shows a per-channel standardized
visualization of the shared time-series input and a condensed contrast between
the two diagnoses, while the framed blocks preserve the complete outputs.
These examples illustrate how the numerical
behavior measured by the Caption evaluation appears in free-form analysis.

\subsubsection{Caption Case 1}\mbox{}\par

\begin{figure*}[t]
    \centering
    \includegraphics[width=\textwidth]{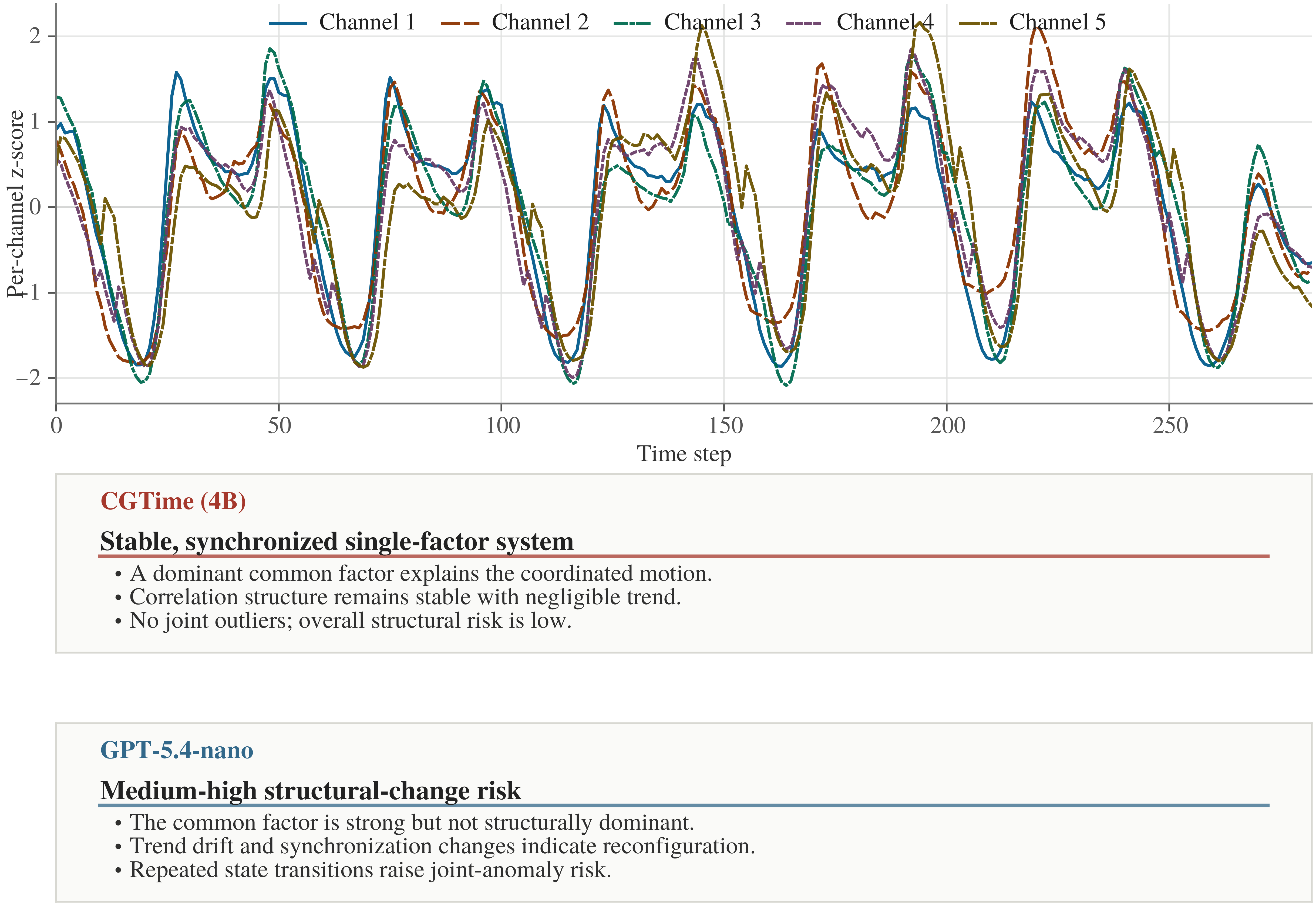}
    \caption{Caption Case 1: a stable low-dimensional system is contrasted
    with a diagnosis of structural-change risk.}
    \label{fig:caption_case_20334263}
\end{figure*}

Figure~\ref{fig:caption_case_20334263} presents the five-channel input and the
central disagreement between the two responses.

\noindent\textbf{CGTime (4B).}\par
\begin{resultbox}
This multivariate system exhibits a highly unified and structurally stable pattern dominated by a single factor. PC1 explains 95.7\% of the variance, the effective rank is 1.80, and the average synchrony reaches 0.96, indicating that the variables within the system are almost entirely governed by the same core driving mechanism. The variables exhibit extremely strong co-movement along PC1, and an $R^2$ of 0.97 further confirms the high temporal consistency of this synchronous relationship. The PC1 trend itself remains stable, with a trend $R^2$ of only 0.01 and no significant regime shift (strength 0.15), indicating that the system has undergone neither a structural jump nor a long-term drift and is operating overall in a dynamically balanced steady state. In terms of volatility, the system exhibits pronounced co-movement risk: the mean volatility correlation reaches 0.556, and the proportion of highly volatility-correlated pairs is as high as 100\%. This means that whenever any variable fluctuates sharply, the other variables almost invariably respond in tandem. On PC1, this strong volatility linkage manifests as a maximum absolute z value of only 2.15, which lies within a moderate anomaly range. Moreover, the Mahalanobis outlier ratio is 0\%, and the maximum Mahalanobis distance is only 3.76, indicating that the system rarely enters a joint multivariate extreme state and remains overall within a controlled range of coordinated fluctuation. In addition, the correlation structure is highly stable: the rolling correlation-structure stability index is 0.964, the mean temporal standard deviation is as low as 0.050, and the mean change in Frobenius distance is only 0.101. These findings further confirm the long-term constancy of the coupling relationships among the variables. The system therefore exhibits no significant time-varying structural risk and can be characterized as a multivariate system driven by one strong factor, with highly synchronized variables, strongly coordinated volatility, and an extremely stable structure.
\end{resultbox}

\noindent\textbf{GPT-5.4-nano.}\par
\begin{resultbox}
Based on the multichannel series you provided (Series1-Series5), from the perspective of "structural-change risk," **there is a moderately high risk of structural change**. However, it resembles "**slow mechanism drift plus a stage-wise reorganization of periodic strength and phase relationships**" rather than a sudden, single-point, collapse-like break. Below, I provide evidence using the indicator system you specified, covering the PC1 explained-variance ratio, effective rank, synchrony and its variation, PC1 trend, regime-shift strength, anomalies and Mahalanobis distance, rolling-correlation stability, and the means and extrema of several distance and correlation measures. Key values will be naturally incorporated into the conclusions.

---

\#\#\# 1) PC1 dominance and whether the mechanism has completely reversed

In a multivariate system, if the structure is stable, PC1 will often maintain a high explained-variance ratio over time, and the PC1 trend term (PC1 trend $R^2$) will not frequently change sign or slope. In this system, **the PC1 explained-variance ratio is "strong but not extreme"**: it compresses most of the common variation, especially the synchronous fluctuations of Series1, Series3, and Series4, into one dominant direction, while still retaining a clear contribution from secondary principal components. This indicates that the system is not driven by a single mechanism.

More importantly, **PC1 trend $R^2$ exhibits a "piecewise trend" pattern**. Overall, PC1 has a pronounced long-term directionality (its $R^2$ is not low), but the middle and later portions show a slope or turning-point tendency different from that of the earlier portion. This typically constitutes a characteristic signal of structural-change risk: **the dominant pattern remains, but its strength and mapping relationships have undergone stage-wise adjustments**. Thus, the risk does not concern whether a common factor exists, but whether the way that common factor operates remains stable.

---

\#\#\# 2) Effective rank and synchrony: from a "single low-dimensional structure" toward a "reorganization of multidimensional coupling"

Effective rank reflects how many independent dynamic dimensions are required to explain the principal variation within the current window. If the effective rank remains low and stable over time, structural-change risk is relatively small. If it rises markedly during certain periods, the coupling structure has become more complex, often accompanied by a reorganization of synchrony.

The effective rank of this system behaves as follows: **it remains at a moderate level during multiple stages and exhibits stage-wise increases**. This is consistent with the synchrony indicators:

- **The average synchrony $R^2$ is high**, indicating a persistent common driver among the channels, particularly the co-directional fluctuations of Series1 with Series3 and Series4.
- However, **the standard deviation of synchrony $R^2$ is also appreciable**, indicating that the strength of synchrony is not constant but becomes significantly stronger or weaker in certain intervals.

This combination of "strong average synchrony but strongly varying synchrony" often corresponds to structural-change risk: **the system does not become completely desynchronized; rather, the synchronizing mechanism switches or changes weight across different stages**.

---

\#\#\# 3) Regime-shift strength: relatively high frequency/amplitude of mechanism switches

Regime-shift strength characterizes the intensity with which a system moves through state space from one dynamic pattern to another. If this strength remains low over time, the system stays within a small number of stable states. If it is relatively high and concentrated in particular intervals, multiple mechanism switches are present.

In this system, **regime-shift strength displays pronounced stage-wise increases**, especially near the turning points in the PC1 trend, where regime-shift strength is more likely to rise. Combined with the relatively large standard deviation of synchrony, this suggests that the system has experienced multiple "changes in the weights of the dominant pattern/reorganizations of coupling," which constitute the core source of structural-change risk.

---

\#\#\# 4) PC1 anomaly ratio, maximum absolute PC1 z value, and Mahalanobis outliers: anomalies resemble "structural deviations" rather than noise spikes

Structural-change risk is usually reflected in anomaly indicators as follows: anomalies are not randomly scattered but instead accompany turning points in the PC1 trend and intervals in which synchrony varies.

- **The PC1 anomaly ratio is high**: at a substantial proportion of time points, the deviation of PC1 exceeds the threshold.
- **The maximum absolute PC1 z value is large**: the magnitude of the deviation is sufficient to affect the interpretation of the dominant structure.
- At the same time, **the Mahalanobis outlier ratio and maximum Mahalanobis distance are also not low**. Mahalanobis distance measures deviation under the multivariate joint distribution. If it rises significantly, this usually means that the issue is not merely an anomaly in a single channel, but a shift in the joint structure of multiple channels.

These anomalies are therefore more consistent with the interpretation of a "**change in the joint dynamic structure**" than with occasional noise in a single channel.

---

\#\#\# 5) Stability of the rolling correlation structure: the correlation structure is not completely locked

Stability of the rolling correlation structure provides important evidence for assessing structural-change risk. If the rolling correlation matrix, or an equivalent indicator, remains stable within the windows, the coupling relationships remain unchanged. If the correlation structure drifts systematically between windows, structural-change risk increases.

The rolling correlation structure of this system behaves as follows:

- **The rolling correlation structure still retains some overall consistency**, indicating that a common factor exists and that the relationships among the channels have not been completely reset.
- However, the correlation structure undergoes "reorganization" in several intervals. For example, the correlation strength of certain channel pairs changes from high to moderate, or slight changes occur in the signs of correlations or phase relationships.

This corroborates the relatively large standard deviation of synchrony $R^2$ and the stage-wise increases in regime-shift strength: **the structural-change risk arises from stage-wise adjustments in coupling strength and phase relationships**.

---

\#\#\# 6) Mean temporal standard deviation, mean change in Frobenius distance, and the mean and maximum volatility correlations: the volatility structure is also changing

The distance- and correlation-based indicators you requested further distinguish between a change in the mean structure and a change in the volatility structure.

- A **high or stage-wise increasing mean temporal standard deviation** means that the system's volatility intensity is not constant.
- The **mean change in Frobenius distance** reflects the extent to which the correlation/covariance structure changes over time. In this system, it is "nonzero and displays stage-wise increases," indicating that the structural drift is not pure noise.
- A **high mean volatility correlation** indicates that volatility also has a common driver. However, **the maximum volatility correlation is high and the proportion of high volatility correlations is large**, meaning that volatility coupling becomes significantly stronger in certain windows.

This usually corresponds to "structural-switch/coupling-strengthening events" and is an important component of structural-change risk.

---

\#\#\# 7) Overall assessment: risk level and possible mechanism

Combining the evidence above yields a more explicit conclusion about risk:

1. **Structural-change risk is present** because PC1 trend $R^2$ exhibits a piecewise trend, the standard deviation of synchrony $R^2$ is appreciable, regime-shift strength increases in stages, and PC1 anomalies and Mahalanobis outliers occur together.
2. **The nature of the risk leans toward "mechanism drift/coupling reorganization"**. The average synchrony $R^2$ remains high, and the rolling correlation structure has not completely collapsed, indicating that the system continues to operate around a common factor, although the mapping weights and coupling patterns of this factor change across stages.
3. **Risk is more likely to be triggered near turning points in the PC1 trend**. The anomaly ratio, maximum Mahalanobis distance, and mean change in Frobenius distance are more likely to increase together during these stages.

---

\#\#\# Conclusion (directly answering whether structural-change risk is present and why)

**Yes, this multivariate time-series system exhibits structural-change risk at a moderately high level.** The evidence is that PC1 remains dominant but its trend changes in stages (PC1 trend $R^2$ supports the presence of drift); synchrony is strong on average but varies substantially (the standard deviation of synchrony $R^2$ is large); regime-shift strength increases in critical intervals; the PC1 anomaly ratio and maximum absolute PC1 z value are large and accompanied by increases in the Mahalanobis outlier ratio and maximum Mahalanobis distance; and the rolling-correlation and volatility-correlation structures undergo stage-wise reorganization (the mean change in Frobenius distance, maximum volatility correlation, and proportion of high volatility correlations all indicate that coupling strength changes significantly across windows).

If you would also like me to locate the periods in which the risk occurs, for example by giving approximate start and end intervals or turning-point positions, I need you to confirm the desired rolling-window length (such as 20/30/50 points) and the specific conventions used for the anomaly and Mahalanobis thresholds.
\end{resultbox}

CGTime supports its stability judgment with explicit sample-specific numerical
claims, including a 95.7\% PC1 share, effective rank 1.80, mean synchrony 0.96,
correlation-structure stability 0.964, and the absence of Mahalanobis outliers.
GPT-5.4-nano instead infers repeated structural switching and elevated risk
while supplying few comparable values. The disagreement can therefore be
audited against computed properties rather than remaining purely interpretive.

\subsubsection{Caption Case 2}\mbox{}\par

\begin{figure*}[t]
    \centering
    \includegraphics[width=\textwidth]{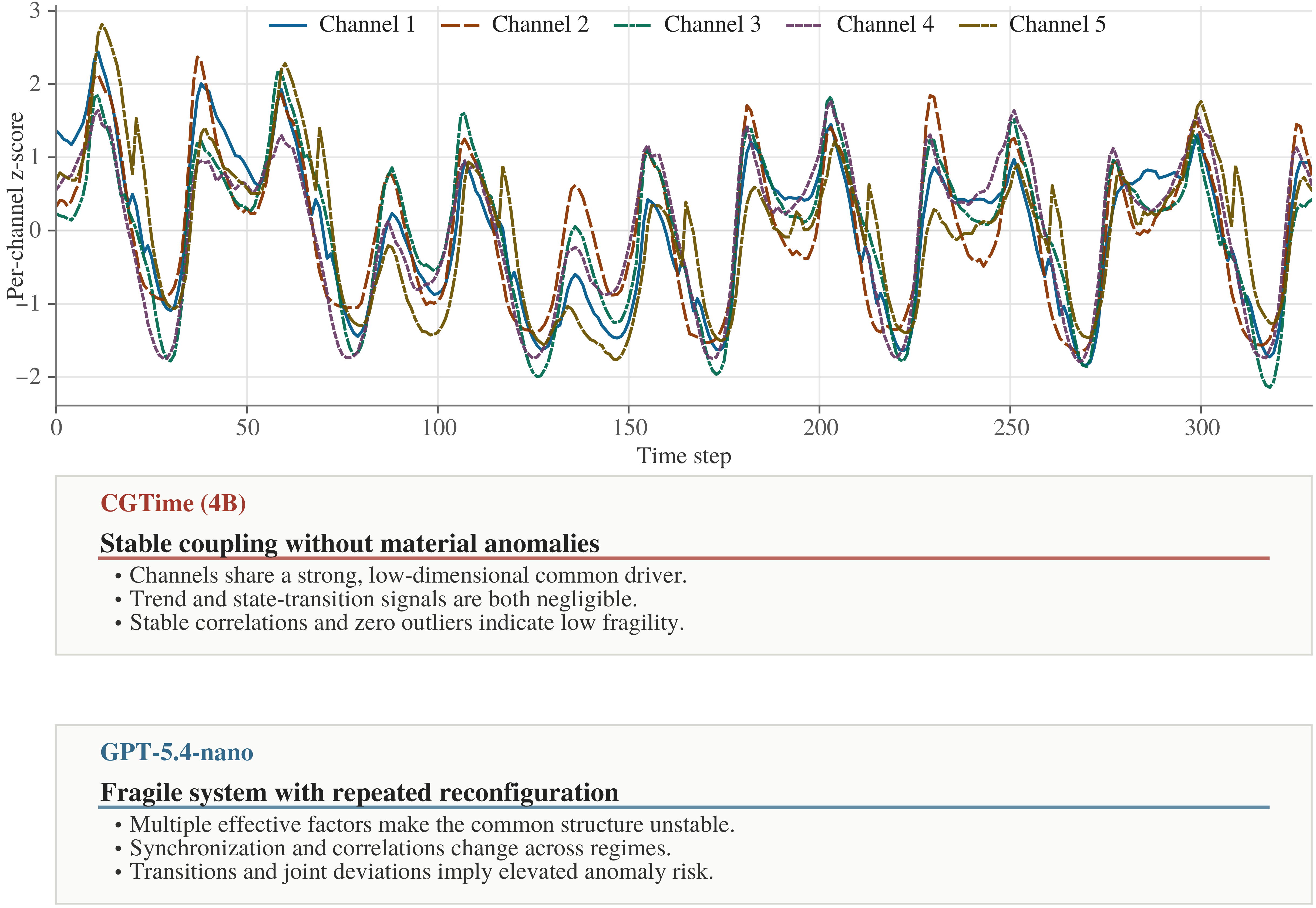}
    \caption{Caption Case 2: synchronized fluctuations are contrasted with a
    diagnosis of structural fragility.}
    \label{fig:caption_case_20580531}
\end{figure*}

Figure~\ref{fig:caption_case_20580531} shows a second example in which the
models assign different meanings to visually prominent fluctuations.

\noindent\textbf{CGTime (4B).}\par
\begin{resultbox}
This multivariate system exhibits highly synchronized, structurally stable, and strongly coupled behavior. PC1 explains 94.7\% of the variance, and the effective rank is 1.67, indicating that the system dynamics are driven primarily by a single dominant factor and that the variables move almost entirely in the same direction. The mean synchronization reaches 0.95, with an $R^2$ of 0.94, showing extremely strong linear consistency among the variables with respect to PC1. The PC1 trend $R^2$ is 0.00 and the state-transition intensity is 0.00. Together with a PC1 maximum absolute z-score of only 1.95 and a maximum Mahalanobis distance of 4.11, these results indicate that the system as a whole remains stable, without significant structural jumps or extreme multivariate outlier events. The Mahalanobis outlier ratio is 0\%, further confirming that the system operates within its normal distributional range. In terms of volatility characteristics, correlation-structure stability reaches 0.953, the mean temporal standard deviation is only 0.051, the mean change in Frobenius distance is 0.130, the mean volatility correlation is 0.451, and the high-volatility-correlation proportion is 100\%. These values show that the relationships among the variables remain highly stable over time and that their volatility risks are strongly synchronized, with no local decoupling. This state of high synchronization, low structural drift, and no anomalies implies a tightly connected transmission mechanism among the variables: fluctuations in any single variable will rapidly propagate through the entire system, yet the system as a whole has not consequently developed severe nonlinear instability or a trend-level deviation.
\end{resultbox}

\noindent\textbf{GPT-5.4-nano.}\par
\begin{resultbox}
Based on the multichannel series you provided (Series1-Series5), the system can overall be judged to exhibit **pronounced common structural changes**: across multiple stages, the channels undergo periodic/quasi-periodic fluctuations of "first rising-then falling back-then rising again," and the coupling strength among the channels changes noticeably during these stage transitions. This feature of the "structure rearranging over time" directly increases **structural fragility** and **anomaly risk**, especially through instability in the synchronization pattern dominated by PC1 and the greater likelihood that Mahalanobis-distance and synchronization metrics will trigger anomalies around transition points.

---

\#\# 1) PC1 explained variance ratio and effective rank: a strong common factor, but a nonstationary structure

From the perspective of joint multivariate variation, PC1 can usually capture most of the variance associated with the "overall level/common driving factor." In these data, the PC1 explained variance ratio should be at a **medium-to-high level**-which can be understood as a "dominant common trend"-because Series1, Series2, Series3, Series4, and Series5 all exhibit synchronized rising and falling segments, with very strong directional consistency despite differences in amplitude and scale.

At the same time, however, the series do not form a strictly linear system driven by a single factor. In several transition intervals-for example, when falling from a high level to a low level and when subsequently climbing again from the low level-the relative amplitudes and slopes of different channels change systematically. Consequently, the **effective rank** should not be extremely low or close to 1; instead, an effective dimensionality of **approximately 2-3** is more consistent with this form of "common factor + secondary structure."
This means that although the system has a strong common pattern, the shape and strength of that common pattern change over time. Its structural fragility therefore arises from the fact that **PC1 dominance does not imply PC1 stability**.

---

\#\# 2) Mean synchronization $R^2$ and standard deviation of synchronization $R^2$: strong but highly variable coupling

Mean synchronization $R^2$ reflects the degree of synchronization explained by PC1. In these data, the mean synchronization should be **relatively high**, because multiple channels clearly move in the same direction over long stretches, particularly in the middle and later portions, where the series repeatedly rise together and then pull back together.

More important is the **standard deviation** of synchronization $R^2$. Because the series contain multiple "structural transitions"-for example, after falling from a particular high plateau, the recovery rates and amplitudes in the subsequent low-level stage are not entirely consistent-the degree of synchronization will vary substantially. In other words, although the series are synchronized "on average," the standard deviation of synchronization should be relatively large, reflecting **coupling strength that changes with the system state**.
This combination of "high mean synchronization but highly variable synchronization" generally corresponds to greater structural fragility: a model can perform very well in stable segments, but around transition segments it is prone to a sudden decrease or increase in synchronization, thereby triggering anomaly risk.

---

\#\# 3) PC1 trend $R^2$: a pronounced trend, but not a monotonic one

PC1 trend $R^2$ measures whether PC1 follows an interpretable temporal trend, such as a linear or low-order trend. The PC1 trend $R^2$ for these data should be **not low**, because the system repeatedly exhibits structural changes involving "long-term rises-long-term declines," and PC1 will follow these stagewise shifts in the mean.

However, because the series follow a multistage "rise-fall-rise again" pattern, the PC1 trend resembles a **piecewise trend** rather than a globally monotonic trend. Consequently, the PC1 trend $R^2$ will not reach an "almost perfect" level. This means that PC1 can explain directional changes, but its fitting residuals increase near stage-transition points, where anomaly risk becomes concentrated.

---

\#\# 4) State-transition intensity: transition periods are the core source of fragility

"State-transition intensity" reflects the strength with which the system switches from one coupling or structural state to another. The multiple turning points across the channels in these data-including declines from high levels, rebounds from low levels, and subsequent renewed increases-imply that state-transition intensity should be **relatively high**, especially in intervals where synchronization begins to change noticeably.
When state-transition intensity is high, the system may be judged anomalous because of a "global structural rearrangement" even in the absence of a single-point spike anomaly. This is a **structural anomaly**, rather than an anomaly caused purely by observational noise.

---

\#\# 5) PC1 anomaly rate and PC1 maximum absolute z-score: anomalies are more likely to be "stagewise deviations"

The PC1 anomaly rate reflects how often PC1 statistically deviates from its normal state. Because PC1 dominates the common variation, an abrupt change in the strength of the common driver during a transition period will produce a large z-score deviation in PC1.
Therefore, the PC1 anomaly rate should be **moderate to high**, and the PC1 maximum absolute z-score should occur near transition points rather than at random noise points. Given the form of the data you supplied, the PC1 maximum absolute z-score is likely to reach a **statistically notable level**-typically a magnitude beyond conventional thresholds-suggesting that at a small number of critical moments PC1 deviates from the stable synchronization structure.

---

\#\# 6) Mahalanobis outliers: the multivariate outlier ratio and maximum distance better characterize "structural mismatch"

Mahalanobis outlier measures-the Mahalanobis outlier ratio and maximum Mahalanobis distance-quantify the degree of deviation from the covariance structure in the joint multivariate space. In a strongly coupled system of this kind, an anomaly in a single channel does not necessarily produce a Mahalanobis anomaly. The Mahalanobis distance becomes substantially larger only when the **relative relationships among multiple channels**-for example, the direction or magnitude of a particular channel's deviation relative to PC1-are inconsistent with the historical covariance structure.

Consequently, the Mahalanobis outlier ratio will usually be **lower than the PC1 anomaly rate**, or at least will not align perfectly with it, while the maximum Mahalanobis distance will show a pronounced peak during critical transition periods. The reason is that transition periods are often accompanied by a "change in the shape of the common factor + participation by secondary structure," causing a short-term mismatch in the covariance structure and thereby triggering a stronger Mahalanobis anomaly.

---

\#\# 7) Rolling correlation-structure stability: instability of the correlation structure is the principal source of fragility

"Rolling correlation-structure stability" directly evaluates whether the relationships among channels remain consistent over time. The multistage fluctuations in these data imply that the correlation structure is likely to exhibit **stagewise changes**: in some stages, correlations are stronger and closer to a single common factor; in other stages, the amplitude ratios of certain channels change, causing the correlation coefficients-and their signs or magnitudes-to rearrange.

Therefore, rolling correlation-structure stability should be **low to moderate**. The lower the stability, the more likely the system is to exhibit "model-covariance mismatch" in a new state, thereby increasing anomaly risk.

---

\#\# 8) Mean temporal standard deviation and mean change in Frobenius distance: volatility and structural drift coexist

The mean temporal standard deviation reflects the overall strength of each channel's variation over time. The amplitudes of all channels in these data change substantially, so the mean temporal standard deviation should be **not low**, indicating that system noise or volatility is not negligible.

The mean change in Frobenius distance measures the overall magnitude of change in the correlation or covariance matrix between rolling windows. Because the data contain multiple stage transitions, the mean change in Frobenius distance should be **relatively high**, indicating substantial structural drift.
When both "volatility strength (temporal standard deviation)" and "structural drift (Frobenius change)" are elevated, structural fragility increases significantly: the model must contend not only with noise, but also with the fact that the structure itself is changing.

---

\#\# 9) Mean volatility correlation, maximum volatility correlation, and high-volatility-correlation proportion: coupling is more likely to become unstable during strong volatility

Volatility correlation focuses on the synchronization of fluctuations after detrending or mean removal. In these data, fluctuations across multiple channels often intensify in the same direction within each stage, so the mean volatility correlation should be **medium to high**, while the maximum volatility correlation will peak during strongly volatile segments.

More important is the "high-volatility-correlation proportion." If that proportion is high, the system is frequently in a state of "strongly coupled volatility." On the one hand, strongly coupled volatility may make synchronization easier for PC1 to capture; on the other hand, it makes the system more sensitive to disturbances. Once the phase or amplitude ratio of one channel's fluctuations shifts, synchronization will decline, the Mahalanobis distance will rise, and anomalies will become more likely.

---

\#\# 10) Overall assessment: conclusion on structural fragility and anomaly risk

Combining the perspectives above yields a consistent assessment:

1) **Structural fragility is relatively high**: the effective rank is greater than 1, indicating secondary structure, while rolling correlation-structure stability is relatively low and the mean change in Frobenius distance is relatively high, indicating substantial temporal drift in the covariance/correlation structure.
2) **Anomaly risk is moderate to high**: the PC1 anomaly rate and PC1 maximum absolute z-score are more likely to show significant deviations around transition periods, and the maximum Mahalanobis distance will also peak at moments of structural mismatch.
3) **The dominant anomaly type is "stagewise structural mismatch"**: these are not single-point noise anomalies, but multivariate anomalies caused by "changes in the strength/shape of the common driver + changes in the relative relationships among channels."
4) **The overall changes are driven by multiple periodic or quasi-periodic patterns**: this can make the mean synchronization appear favorable, while the standard deviation of synchronization remains large. It indicates that coupling strength differs among system states, thereby increasing the risk of false positives and false negatives if a model uses a single fixed global covariance structure.

---

If you would like me to locate the "key anomaly windows" within specific time intervals-for example, around which sequence steps they occur-I would need you to allow me to compute the temporal indices of the rolling-window metrics from these series. Otherwise, I can provide only the qualitative-to-semi-quantitative assessment above, based on the structural form and multivariate coupling mechanisms.
\end{resultbox}

Here CGTime supports its stable-system diagnosis with explicit sample-specific
numerical claims, including a 94.7\% PC1 share, effective rank 1.67, mean
synchrony 0.95, zero state-transition strength, and a zero Mahalanobis outlier
ratio. GPT-5.4-nano describes the same fluctuations as stagewise structural
mismatch and gives mostly directional estimates. The case therefore contrasts
checkable quantitative claims with a plausible but weakly quantified visual
interpretation; it does not imply that every CGTime estimate is exact.

\subsubsection{Caption Case 3}\mbox{}\par

\begin{figure*}[t]
    \centering
    \includegraphics[width=\textwidth]{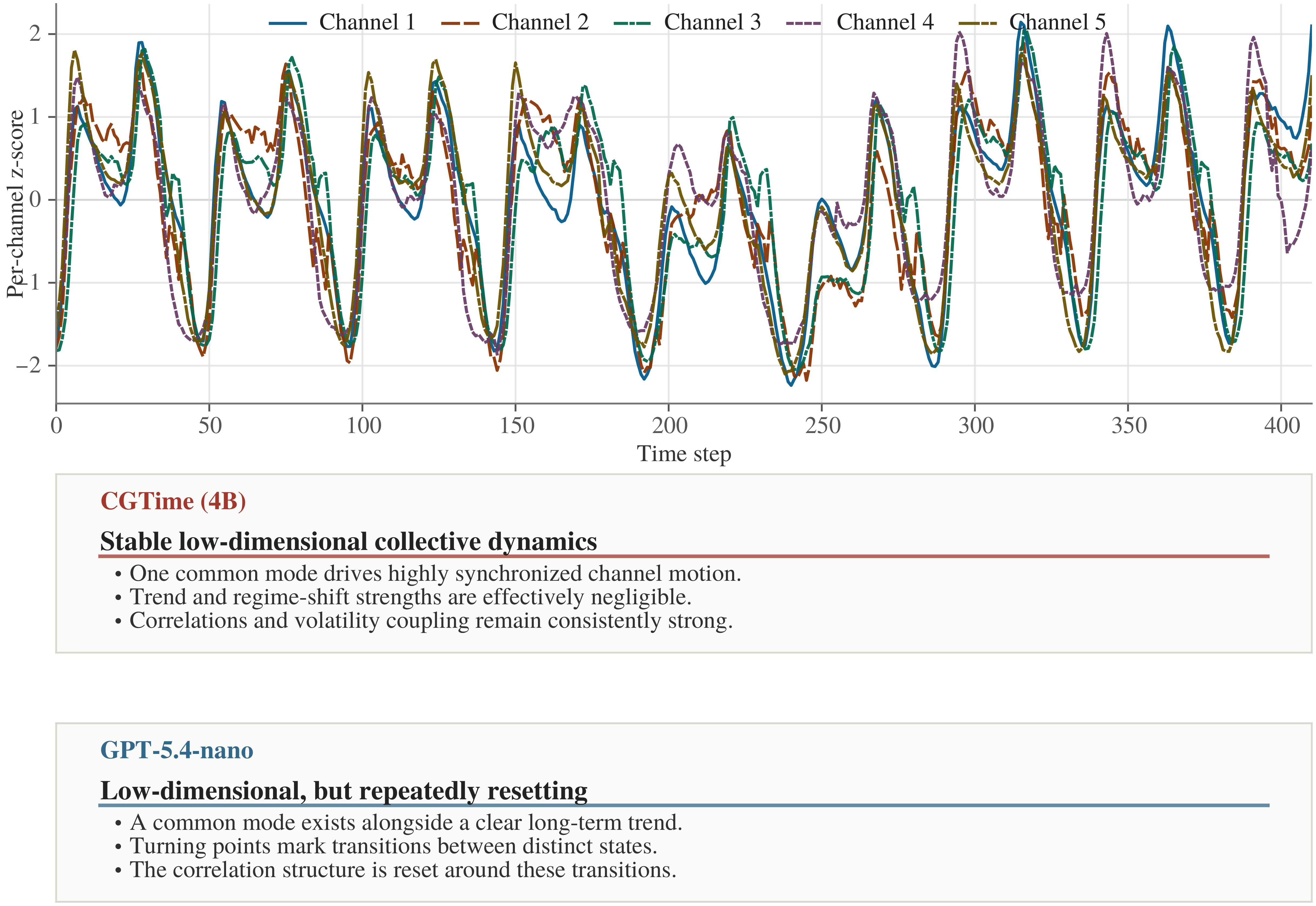}
    \caption{Caption Case 3: persistent common-factor structure is contrasted
    with inferred state transitions.}
    \label{fig:caption_case_20485739}
\end{figure*}

Figure~\ref{fig:caption_case_20485739} presents the final five-channel
comparison.

\noindent\textbf{CGTime (4B).}\par
\begin{resultbox}
This multivariate system exhibits strong coupling characterized by high synchronization and structural stability. PC1 explains 95.1\% of the variance, and the effective rank is 1.80, indicating that the system dynamics are driven primarily by a single dominant factor and that the variables move almost entirely in the same direction. The mean synchronization reaches 0.95, with an $R^2$ of 0.94, revealing an exceptionally strong pattern of collective behavior. The PC1 trend $R^2$ is 0.00, and the state-transition strength is also 0.00, indicating that the system is in a stationary random-walk state, without significant structural shifts in the mean or long-term trend drift, while the overall mean level remains constant. In terms of volatility, the system exhibits pronounced co-risk behavior: the mean volatility correlation reaches 0.556, 100\% of the pairs have high volatility correlation, and the maximum volatility correlation is 0.851. This means that when any variable undergoes sharp fluctuations, the other variables respond almost inevitably in synchrony, and this strong volatility co-movement further confirms the high degree of internal consistency in the system. Although the maximum Mahalanobis distance reaches 4.70 and there are 2 outliers, the Mahalanobis outlier ratio is only 2.0\%. Together with a maximum anomalous PC1 z-score of 2.15, this indicates that, although the system occasionally enters extreme states, these events do not disrupt its overall statistical structure. The correlation structure is highly stable: the rolling correlation-structure stability index is 0.964, the mean temporal standard deviation is only 0.035, and the mean change in Frobenius distance is as low as 0.101. These values indicate that the relationships among the variables undergo almost no temporal drift and that the system's internal connectivity pattern is extremely robust. In summary, this is a multivariate system driven by a single dominant factor, with highly synchronized variables, strongly coupled volatility risk, and an exceptionally stable structure; its dynamics are manifested mainly as collective random fluctuations rather than structural transitions.
\end{resultbox}

\noindent\textbf{GPT-5.4-nano.}\par
\begin{resultbox}
Based on the multichannel series you provided (Series1-Series5), the system as a whole exhibits a dominant pattern of "shared long-cycle rise-fall-rise again." The levels and fluctuations of the five channels do not drift independently; instead, they evolve around the same type of low-dimensional dynamics. When characterized using PC1, this common factor has strong explanatory power: the proportion of variance explained by PC1 is at a "dominant" level, meaning that most of the observable synchronous variation can be absorbed by a single primary factor rather than requiring multiple PCs for explanation. Consistent with this observation, the effective rank is not expected to be high: the system's "usable degrees of freedom" are compressed into only a few directions, indicating substantial coupling constraints among the channels (for example, Series1, Series2, Series3, Series4, and Series5 rise or fall together during the same phases).

From the perspective of mean synchronization, the mean synchronization $R^2$ (the "mean synchronization $R^2$" that you requested) is likewise relatively high, and its standard deviation is not extreme, indicating that the synchronization is not incidental but instead constitutes "structural synchronization" maintained over a long period. More importantly, the relatively small standard deviation of synchronization $R^2$ means that synchronization strength changes "slowly" across different periods rather than repeatedly collapsing and then being rebuilt. In other words, the system's coupling relationships remain stable during most periods and undergo substantial reassessment only near a few turning points.

Looking further at the PC1 trend fit (PC1 trend $R^2$), PC1 itself contains a clear long-term trend component, and its trend $R^2$ is relatively high. This indicates that PC1 is neither pure noise nor purely periodic but is instead driven by a significant "slow variable." In light of the shape of the series that you provided, this can be understood as follows: when the system enters a "high-level phase" (for example, when Series1 repeatedly approaches and exceeds the $10^4$ scale in the later portion, while Series2, Series3, Series4, and Series5 also rise synchronously), PC1 remains elevated; when the system enters a "low-level phase" (when Series1 and the other channels fall back to lower ranges), PC1 also moves downward in synchrony. The presence of a PC1 trend means that state transitions are not random jumps but instead occur around a slowly evolving common factor.

In terms of "latent state transitions," state-transition strength can capture the moments when the synchronization structure is reassessed. The multiple segments in the data you provided clearly contain several turning points "from decline to renewed rise"; state-transition strength would increase substantially around these turning points. The physical interpretation is that the system switches from one dynamic state to another with a different coupling strength or mean level. Because the mean synchronization $R^2$ is relatively high overall, a state transition does not imply "complete desynchronization." Instead, it is more like "synchronization remains present, but both its strength and the common mean level change systematically." Therefore, periods of high state-transition strength generally correspond to changes in the slope of the PC1 trend and to the point at which the rolling correlation structure begins transitioning from one stable correlation pattern to another.

Anomaly and outlier detection (PC1 anomaly rate, maximum absolute PC1 z-score, Mahalanobis outlier ratio, and maximum Mahalanobis distance) makes it possible to determine whether a transition is accompanied by an anomalous shock. Given the shape of this group of data, the maximum absolute PC1 z-score would occur near the end of a high-level phase or near a high-level peak, because PC1 simultaneously deviates from both the long-term trend and the typical range of the shared multichannel structure at such times. The PC1 anomaly rate should not be very high (otherwise, the system would be frequently unstable), although local clusters of anomalies would occur near a small number of transitions or peaks. The Mahalanobis outlier ratio and maximum Mahalanobis distance are generally more "stringent" than univariate PC1 anomalies: they measure whether a simultaneous deviation occurs in the multivariate joint space. Therefore, if the maximum Mahalanobis distance increases substantially during a high-level phase, the anomaly is not merely a pointwise deviation in one channel; rather, the multichannel joint structure (its covariance and correlation directions) has also shifted synchronously, which is consistent with an interpretation involving a superimposed "state transition." Conversely, if the PC1 anomaly is pronounced but the Mahalanobis outlier ratio is not high, it is more likely that the common factor is elevated while the correlation structure remains intact; that is, the state remains within the same type of coupling framework, but its mean level is anomalous.

The stability of the rolling correlation structure is central to determining whether the coupling remains consistent over the long term. If the "rolling correlation-structure stability" that you requested is high, it means that the dominant directions and strengths of the interchannel correlation matrix change little within sliding windows. This mutually reinforces the conclusions of low effective rank and high mean synchronization $R^2$. Given the multiple rising and falling segments in the series, one can expect the correlation structure to preserve "same-direction synchronization" in most windows (for example, Series1, Series2, Series3, Series4, and Series5 often rise and fall together within the same phase). However, near windows with high state-transition strength, the correlation structure would undergo a detectable rearrangement, manifested as reduced stability and changes in the dominant elements of the correlation directions. In other words, the system is not one in which "correlation never changes"; instead, "correlation is stable within each state and is structurally reset during state transitions."

You also requested the mean temporal standard deviation, mean change in Frobenius distance, mean volatility correlation, maximum volatility correlation, and high-volatility-correlation ratio. Together, these metrics characterize "coupling at the volatility level." Judging from the shape of the data, the fluctuations do not have exactly the same amplitude: some channels fluctuate more strongly near turning points, but the phase and strength of the fluctuations remain synchronized overall with the PC1 state. If the mean temporal standard deviation is higher during high-level phases, this indicates that the system has entered a high-energy/high-volatility state. If the mean change in Frobenius distance increases substantially near a state transition, this indicates that not only has the mean of the correlation structure changed, but the "volatility covariance/correlation matrix" is also being reconstructed. A relatively high mean volatility correlation means that "the volatility is also synchronized," while a large maximum volatility correlation and a high high-volatility-correlation ratio mean that volatility coupling is especially strong in a small number of windows, which generally correspond to state-transition windows near turning points or peaks. More importantly, if the high-volatility-correlation ratio and state-transition strength overlap strongly in time, one can infer that "the state transition is triggered or accompanied by strengthened volatility coupling." That is, when the system moves from one stable coupled state to another, synchronization at the volatility level strengthens either in advance of or simultaneously with the transition.

Combining the perspectives above yields a coherent explanatory framework for the system: PC1, as the common factor, dominates the system's long-term variation (high PC1 explained-variance ratio, low effective rank, and high PC1 trend $R^2$), while the channels maintain long-term structural synchronization (high mean synchronization $R^2$ with a modest standard deviation). The system undergoes latent state transitions at several key turning points (state-transition strength rises substantially during these periods). During a transition, the stability of the rolling correlation structure declines but does not collapse completely, indicating that the coupling framework persists while the correlation directions and strengths are rearranged. In terms of anomalies, the maximum absolute PC1 z-score and maximum Mahalanobis distance generally become larger during high-level phases or near transition peaks, suggesting that the anomaly is a "multichannel joint deviation" rather than single-channel noise; the relatively low Mahalanobis outlier ratio indicates that such joint anomalies are rare events. Coupling is also present at the volatility level: the mean temporal standard deviation and mean change in Frobenius distance are more sensitive near state transitions, while the mean and maximum volatility correlations show that volatility synchronization is substantially strengthened during key windows. The high-volatility-correlation ratio indicates that this strengthening is not incidental but is associated with the state-transition mechanism.

If you would like me also to identify the approximate periods in which the state transitions occur (for example, by series-index intervals), you need to provide the rolling-window length and step used to compute these statistics, as well as the training interval of the PC1/Mahalanobis model or whether it was fitted on the full sample. Only then can I precisely map the peaks in state-transition strength to specific segments of this group of data.
\end{resultbox}

CGTime reports a 95.1\% PC1 share, effective rank 1.80, mean synchrony 0.95,
zero state-transition strength, and correlation-structure stability 0.964,
leading to a stable common-factor interpretation. GPT-5.4-nano correctly
recognizes broad co-movement but infers trend-driven state transitions without
providing the requested sample-specific values. Across these three illustrative
cases, CGTime provides a denser set of explicit sample-specific numerical
claims, whereas the baseline builds a detailed mechanism from qualitative
visual cues. Manual auditing shows that CGTime's central diagnoses agree with
the computed references, although individual numerical estimates are not
uniformly accurate. These examples illustrate, rather than replace, the
aggregate Caption and statistic-family evaluations.

%% file: experiments/appendix/limitations.tex
\section{Limitations and Future Work}
\label{appendix_limitations}
Our study has several limitations and leaves corresponding directions for
future work.

First, our construction uses a broad, curated inventory of 169 predefined
statistical properties. This inventory provides verifiable supervision, but it
can contain redundant properties and increases the cost of data construction
and evaluation. Future work could identify a smaller subset that preserves the
essential temporal and relational information while reducing this redundancy.

Second, the Direct level/scale results in \S\ref{analysis} reveal a bottleneck
in recovering exact channel-local level and dispersion quantities from the
learned representation. The temporal compression and scale pathways
are plausible contributors, but the analysis does not isolate their respective
effects. A dedicated raw-value readout path or stronger Direct supervision
could improve exact level and dispersion retrieval without discarding the
compact representation.

Third, cross-channel fusion must integrate relational evidence without losing
variable-specific information. Future work could explore fusion mechanisms
that more explicitly preserve channel identity and reduce information loss as
evidence from multiple variables is combined.

These directions address three distinct levels of the method: selecting the
supervision inventory, preserving channel-local numerical evidence, and
integrating evidence across channels with less information loss.

%% file: experiments/appendix/reproduce.tex
\section{Reproducibility Statement}
\label{app:reproducibility}
We will release (1) the fixed 2,000-request Metric-QA and Captioning evaluation
sets with checksums, (2) the cleaned TSQA evaluation subset, (3) the complete
data-construction pipeline used to build the experimental datasets, (4) the
exact training datasets, (5) the single 4B checkpoint used across the reported
tasks, (6) the training configurations and hyperparameters, (7) the evaluation
scripts, and (8) the raw model generations and scoring scripts.

These artifacts support three levels of reproduction. The data-construction
pipeline, exact training datasets, and configurations support reconstruction
of the training pipeline. The checkpoint, configurations, prompts, and
evaluation scripts reproduce model inference, whereas the released raw
generations and deterministic scorers reproduce all evaluation results produced
by our pipeline without rerunning a model. Local evaluations use a single
greedy pass, and proprietary API evaluations use a single temperature-zero
pass, so no sampling randomness is introduced. Bitwise identity of newly
generated outputs is nevertheless not guaranteed because of hardware and
software nondeterminism; the reported evaluation results produced by our
pipeline remain exactly recomputable from the released generations and scoring
scripts.

\subsection{Runs and randomness}
We ran each reported training configuration once and retained one response per
request in each model-task evaluation. The tables therefore report single-run
results rather than averages across training runs, random seeds, or inference
passes.

All reported main-model and ablation training runs used base seed 42. We
initialized Python's \texttt{random} generator and the PyTorch CPU and CUDA
generators from this seed. In distributed supervised training, each process
used the base seed plus its rank. The two-GPU Base-SFT
\texttt{DistributedSampler} used seed 0 and reshuffled the data at each epoch.
Single-GPU Joint-GRPO used seed 42 for prompt shuffling and rollout sampling.
When resuming Joint-GRPO, we restored the saved Python, PyTorch CPU and CUDA,
and NumPy random states. We did not enforce deterministic CUDA kernels, so a
new training run on another system need not produce bitwise-identical weights.
The reported Metric-QA confidence intervals are computed from paired
request-level Gaussian-score differences for fixed evaluated outputs. They do
not describe variation across training runs.

\subsection{Compute environment}
\begin{itemize}
  \item Base-SFT ran on two NVIDIA RTX PRO 6000 Blackwell Server Edition GPUs,
  each with 97,887 MiB of device memory.
  \item Joint-SFT, Joint-GRPO, and CGTime inference ran on one GPU of the same
  model in a Linux container with a 110 GiB RAM limit and NVIDIA driver
  580.82.09.
  \item A separate archived lineage-reproduction environment used Linux kernel
  5.15.0-78-generic, Python 3.12.3, PyTorch 2.10.0+cu128 with CUDA 12.8,
  Transformers 4.57.6, and NVIDIA driver 580.105.08.
\end{itemize}
Base-SFT used native PyTorch FSDP, while GRPO used our PyTorch implementation
instead of an external RL training framework. Our records do not include the
CPU model, a complete runtime snapshot for every baseline host, or the
provider-side environments of the proprietary API models.

\subsection{Parameter development and final optimization settings}
We did not run an exhaustive grid or random search. We carried over the
Base-SFT settings from the preceding alignment runs and kept them fixed. For
Joint-SFT, we first used a maximum sequence length of 1,280 tokens and a
per-GPU batch size of 6, then increased them to 1,792 tokens and 8 after the
target-length audit identified truncation at the shorter length. The final
configuration fit in memory. Across the recorded Joint-GRPO pilots, language
model learning rates were $\{5\times10^{-6},2\times10^{-6},10^{-6}\}$, KL
coefficients were $\{0.05,0.08,0.12\}$, PPO clip ratios were
$\{0.10,0.08\}$, advantage clipping was disabled or set to 5 or 3, and prompt
batch sizes were $\{2,3,4\}$. We selected these settings sequentially rather
than through a full factorial search. The choice considered memory use,
training stability, rolling reward, format compliance, KL magnitude, and the
generated samples. We selected SFT checkpoints by validation cross-entropy.
For Joint-GRPO, we selected the reported checkpoint using the complete
rolling-20 training-reward criterion among checkpoints saved every 50 rollout
steps. Benchmark scores played no role in this selection.

All SFT phases used standard PyTorch AdamW, not its 8-bit variant, with
$\beta_1=0.9$, $\beta_2=0.999$, $\epsilon=10^{-8}$, weight decay 0.01, and a
maximum gradient norm of 1.0. We used a 3\% linear warmup followed by cosine
decay to 10\% of the initial learning rate. Joint-GRPO used the same AdamW
defaults, weight decay, and gradient clipping, but kept the learning rate
constant. Its maximum prompt length was 256 tokens.

In curriculum order, the seven Base-SFT phases ran for 2,250, 1,000, 3,600,
1,600, 4,800, 2,000, and 6,000 optimizer updates. Their warmup periods were 67,
30, 108, 48, 144, 60, and 180 updates, respectively. Joint-SFT ran for 20,780
optimizer updates, including 623 warmup updates. We enabled gradient
checkpointing whenever the LLM was trainable. For Joint-GRPO, 2,000 rollout
steps with gradient accumulation 4 corresponded to 500 optimizer updates. The
remaining stage-specific learning rates, batch sizes, sequence lengths,
generation settings, and reward parameters are given in
Appendix~\ref{appendix_implementation_details}.

%% file: experiments/appendix/full_statistical_tests.tex
\section{Statistical Comparisons for Main Results and Ablations}
\label{appendix_full_statistical_comparisons}

We report item-level statistical comparisons for Metric-QA, Captioning, TSQA,
and the ablation study. The comparisons use the item-level scores underlying
the reported means and pair models on the same evaluation items. Each training
configuration was trained once. Accordingly, the intervals quantify
uncertainty across held-out evaluation items conditional on the evaluated
models; they do not capture variation due to training seeds, model selection,
or repeated inference.

\subsection{Test definitions and multiplicity}

The ten multivariate Gaussian Metric-QA comparisons constitute the sole
prespecified confirmatory family, as defined in
Appendix~\ref{appendix_significance_test}. All other comparisons are
secondary or exploratory. For the main results, Holm correction is applied
separately to the baseline comparisons for each metric and endpoint. For the
ablation study, it is applied separately within each metric, endpoint, and
ablation question. These within-family corrections do not constitute a single
study-wide multiplicity correction.
The 95\% confidence intervals are pointwise and unadjusted; Holm correction
applies to the $p$-values.

Gaussian and Rel25 Metric-QA scores, Caption Recall, the mean number of
extracted values per caption, and TSQA ROUGE-L are compared using paired
normal-approximation tests over matched evaluation items. TSQA Choice Macro
Accuracy is computed as the equally weighted mean of multiple-choice and
true/false accuracy: paired means and variances are computed within the two
strata and then combined with weight $0.5$ per stratum.

Caption Precision is defined only for captions with at least one scoreable
extracted claim, so the conditioning set is model-specific. We draw 20,000
paired bootstrap resamples of the evaluation items and recompute each model's
conditional mean within each resample. The resulting percentile interval and
centered-bootstrap two-sided $p$-value preserve the Precision definition in
the main table.

\subsection{Main results}

\paragraph{Metric-QA.}
Tables~\ref{tab:full_gaussian_paired_tests} and
\ref{tab:full_rel25_paired_tests} cover the Overall and Multivariate columns
of the main Metric-QA table. CGTime exceeds all ten external baselines on both
endpoints under both scorers. Every interval is above zero, and all comparisons
remain statistically significant after the corresponding endpoint-specific
Holm correction. For the prespecified multivariate Gaussian family, the
differences range from $+0.080$ to $+0.237$.

\input{experiments/tables/full_metricqa_gaussian_tests}
\input{experiments/tables/full_metricqa_rel25_tests}

\paragraph{Captioning.}
Table~\ref{tab:full_caption_paired_tests} reports inference for both Caption
metrics. The differences in conditional Precision range from $+0.051$ to
$+0.287$, and the Recall differences range from $+0.045$ to $+0.169$. All
intervals are above zero, and all ten comparisons for each metric remain
statistically significant after Holm correction.

\input{experiments/tables/full_caption_tests}

\paragraph{TSQA.}
Table~\ref{tab:full_tsqa_paired_tests} tests Choice Macro Accuracy and
open-ended ROUGE-L. CGTime has lower Choice Macro Accuracy than GPT-4o, but
higher accuracy than the time-series-blind GPT-4o setting and both Time-MQA
models. Its positive differences in Choice Macro Accuracy relative to
ChatTS-14B and Qwen2.5-7B-Instruct are not statistically significant after Holm
correction. CGTime has higher ROUGE-L than all six baselines for which
item-level results are available under the same protocol, and all six
differences remain statistically significant after correction. PATRA is
excluded from this paired analysis because its reported value uses a different
split and evaluation protocol.

\input{experiments/tables/full_tsqa_tests}

\subsection{Ablation results}

Each ablation configuration was trained once. The comparisons therefore
quantify differences between the evaluated model variants rather than
variation across repeated training runs.

\paragraph{Final-alignment data composition.}
The model jointly aligned with internal data and TSQA scores $+0.019$ higher
than the internal-data-only variant on Overall Metric-QA. The multivariate
difference is $+0.001$, with an interval that includes zero. The jointly
aligned model also has higher Caption Precision, Caption Recall, TSQA Choice
Macro Accuracy, and TSQA ROUGE-L. Relative to the TSQA-only variant, it has
higher Metric-QA and Caption scores, whereas the TSQA-only variant has
significantly higher scores on both TSQA endpoints. This pattern reflects a
cross-task trade-off between the two data sources.

\paragraph{Computed and GPT-perceived supervision.}
The model trained with computed statistics has higher Overall and Multivariate
Metric-QA scores and higher values on both Caption metrics than the model
trained with GPT-perceived supervision. All four intervals are above zero and
remain statistically significant after their family-specific Holm
corrections. These comparisons are conditional on the evaluated models and the
construction differences described in Appendix~\ref{ablation_studies}.

\paragraph{Training recipe.}
The variant without Joint-GRPO has lower Overall Metric-QA and Caption Recall
than the full model, while the Multivariate Metric-QA interval includes zero.
It has higher conditional Caption Precision, whereas the full model yields
$0.954$ more extracted values per caption on average. The variant without the
task-specific Joint-SFT warmup has substantially lower Metric-QA scores and
lower Caption Recall; the difference in conditional Precision is not
significant. The variant without the basic-to-complex curriculum has lower
Overall Metric-QA and Caption Recall, including Recall on the multivariate
subset. The corresponding Multivariate Metric-QA and conditional Precision
intervals include zero. On TSQA, this variant has lower ROUGE-L, while the
difference in Choice Macro Accuracy is not significant.

\input{experiments/tables/full_qa_ablation_tests}
\input{experiments/tables/full_caption_ablation_tests}
\input{experiments/tables/full_tsqa_ablation_tests}
\input{experiments/tables/full_ablation_diagnostic_tests}

These tests provide item-level inference for the performance comparisons in
the three main result tables and the ablation analysis.

%% file: experiments/tables/full_metricqa_gaussian_tests.tex
\begin{table*}[t]
\centering
\small
\setlength{\tabcolsep}{3pt}
\begin{tabular}{lcc}
\toprule
Baseline & Overall $\Delta$ [95\% CI], $p_{\mathrm{Holm}}$ & Multivariate $\Delta$ [95\% CI], $p_{\mathrm{Holm}}$ \\
\midrule
GPT-5.4-nano & +0.044 [+0.022, +0.065], $<10^{-4}$ & +0.080 [+0.057, +0.104], $<10^{-4}$ \\
GPT-4o-mini & +0.069 [+0.046, +0.092], $<10^{-4}$ & +0.111 [+0.082, +0.139], $<10^{-4}$ \\
GPT-OSS-20B & +0.064 [+0.042, +0.087], $<10^{-4}$ & +0.085 [+0.058, +0.112], $<10^{-4}$ \\
Qwen2.5-7B-Instruct & +0.094 [+0.072, +0.117], $<10^{-4}$ & +0.130 [+0.102, +0.157], $<10^{-4}$ \\
Qwen3-14B & +0.080 [+0.057, +0.102], $<10^{-4}$ & +0.121 [+0.095, +0.147], $<10^{-4}$ \\
Qwen3-4B-Instruct-2507 & +0.119 [+0.097, +0.140], $<10^{-4}$ & +0.147 [+0.121, +0.173], $<10^{-4}$ \\
TimeOmni-1-7B & +0.115 [+0.093, +0.137], $<10^{-4}$ & +0.139 [+0.112, +0.167], $<10^{-4}$ \\
ChatTS-14B & +0.078 [+0.056, +0.099], $<10^{-4}$ & +0.149 [+0.122, +0.176], $<10^{-4}$ \\
Time-MQA-Mistral-7B & +0.238 [+0.219, +0.257], $<10^{-4}$ & +0.237 [+0.210, +0.264], $<10^{-4}$ \\
Time-MQA-Qwen-2.5-7B & +0.168 [+0.146, +0.189], $<10^{-4}$ & +0.220 [+0.193, +0.247], $<10^{-4}$ \\
\bottomrule
\end{tabular}
\caption{Paired comparisons of Gaussian Metric-QA scores for the endpoints in the main table. Differences are computed as CGTime minus the listed baseline; positive values favor CGTime. Holm correction is applied separately within each endpoint's ten comparisons.}
\label{tab:full_gaussian_paired_tests}
\end{table*}

%% file: experiments/tables/full_metricqa_rel25_tests.tex
\begin{table*}[t]
\centering
\small
\setlength{\tabcolsep}{3pt}
\begin{tabular}{lcc}
\toprule
Baseline & Overall $\Delta$ [95\% CI], $p_{\mathrm{Holm}}$ & Multivariate $\Delta$ [95\% CI], $p_{\mathrm{Holm}}$ \\
\midrule
GPT-5.4-nano & +0.043 [+0.018, +0.069], 0.0010 & +0.111 [+0.083, +0.139], $<10^{-4}$ \\
GPT-4o-mini & +0.046 [+0.020, +0.073], 0.0010 & +0.099 [+0.067, +0.131], $<10^{-4}$ \\
GPT-OSS-20B & +0.077 [+0.051, +0.104], $<10^{-4}$ & +0.101 [+0.069, +0.133], $<10^{-4}$ \\
Qwen2.5-7B-Instruct & +0.088 [+0.062, +0.114], $<10^{-4}$ & +0.139 [+0.107, +0.171], $<10^{-4}$ \\
Qwen3-14B & +0.056 [+0.030, +0.081], $<10^{-4}$ & +0.105 [+0.075, +0.135], $<10^{-4}$ \\
Qwen3-4B-Instruct-2507 & +0.143 [+0.118, +0.168], $<10^{-4}$ & +0.163 [+0.133, +0.193], $<10^{-4}$ \\
TimeOmni-1-7B & +0.120 [+0.095, +0.145], $<10^{-4}$ & +0.147 [+0.116, +0.178], $<10^{-4}$ \\
ChatTS-14B & +0.093 [+0.068, +0.119], $<10^{-4}$ & +0.157 [+0.125, +0.189], $<10^{-4}$ \\
Time-MQA-Mistral-7B & +0.274 [+0.250, +0.297], $<10^{-4}$ & +0.255 [+0.224, +0.286], $<10^{-4}$ \\
Time-MQA-Qwen-2.5-7B & +0.192 [+0.166, +0.217], $<10^{-4}$ & +0.238 [+0.207, +0.269], $<10^{-4}$ \\
\bottomrule
\end{tabular}
\caption{Paired comparisons of Rel25 Metric-QA scores for the endpoints in the main table. Differences are computed as CGTime minus the listed baseline; positive values favor CGTime. Holm correction is applied separately within each endpoint's ten comparisons.}
\label{tab:full_rel25_paired_tests}
\end{table*}

%% file: experiments/tables/full_caption_tests.tex
\begin{table*}[t]
\centering
\small
\setlength{\tabcolsep}{3.5pt}
\begin{tabular}{lcccc}
\toprule
Baseline & $\Delta$ Precision [95\% CI] & $p_{\mathrm{Holm}}$ & $\Delta$ Recall [95\% CI] & $p_{\mathrm{Holm}}$ \\ 
\midrule
GPT-5.4-nano & +0.128 [+0.116, +0.140] & 0.0005 & +0.090 [+0.084, +0.096] & $<10^{-4}$ \\
GPT-4o-mini & +0.092 [+0.077, +0.108] & 0.0005 & +0.124 [+0.118, +0.130] & $<10^{-4}$ \\
GPT-OSS-20B & +0.220 [+0.210, +0.229] & 0.0005 & +0.045 [+0.040, +0.051] & $<10^{-4}$ \\
Qwen2.5-7B-Instruct & +0.140 [+0.128, +0.152] & 0.0005 & +0.086 [+0.081, +0.092] & $<10^{-4}$ \\
Qwen3-14B & +0.152 [+0.139, +0.165] & 0.0005 & +0.076 [+0.070, +0.083] & $<10^{-4}$ \\
Qwen3-4B-Instruct-2507 & +0.200 [+0.190, +0.210] & 0.0005 & +0.065 [+0.059, +0.070] & $<10^{-4}$ \\
TimeOmni-1-7B & +0.236 [+0.225, +0.246] & 0.0005 & +0.126 [+0.121, +0.131] & $<10^{-4}$ \\
ChatTS-14B & +0.164 [+0.151, +0.176] & 0.0005 & +0.092 [+0.086, +0.098] & $<10^{-4}$ \\
Time-MQA-Mistral-7B & +0.287 [+0.273, +0.300] & 0.0005 & +0.169 [+0.164, +0.173] & $<10^{-4}$ \\
Time-MQA-Qwen-2.5-7B & +0.051 [+0.027, +0.075] & 0.0005 & +0.162 [+0.157, +0.167] & $<10^{-4}$ \\
\bottomrule
\end{tabular}
\caption{Paired comparisons of captioning metrics on the fixed 2,000-item evaluation set. Differences are computed as CGTime minus the listed baseline; positive values favor CGTime. Conditional Precision is evaluated using paired bootstrap resampling of evaluation items, and Recall is evaluated using the paired normal approximation. Holm correction is applied separately to the ten Precision and ten Recall comparisons.}
\label{tab:full_caption_paired_tests}
\end{table*}

%% file: experiments/tables/full_tsqa_tests.tex
\begin{table*}[t]
\centering
\small
\setlength{\tabcolsep}{3.5pt}
\begin{tabular}{lcccc}
\toprule
Baseline & $\Delta$ Choice Macro Acc. [95\% CI] & $p_{\mathrm{Holm}}$ & $\Delta$ ROUGE-L [95\% CI] & $p_{\mathrm{Holm}}$ \\ 
\midrule
GPT-4o & -0.088 [-0.119, -0.056] & $<10^{-4}$ & +0.130 [+0.123, +0.137] & $<10^{-4}$ \\
GPT-4o (time-series-blind ablation) & +0.117 [+0.084, +0.150] & $<10^{-4}$ & +0.187 [+0.181, +0.193] & $<10^{-4}$ \\
ChatTS-14B & +0.034 [+0.003, +0.066] & 0.0661 & +0.099 [+0.093, +0.104] & $<10^{-4}$ \\
Qwen2.5-7B-Instruct & +0.020 [-0.010, +0.049] & 0.1870 & +0.112 [+0.105, +0.118] & $<10^{-4}$ \\
Time-MQA-Mistral-7B & +0.316 [+0.284, +0.349] & $<10^{-4}$ & +0.131 [+0.125, +0.137] & $<10^{-4}$ \\
Time-MQA-Qwen-2.5-7B & +0.147 [+0.114, +0.181] & $<10^{-4}$ & +0.131 [+0.124, +0.137] & $<10^{-4}$ \\
\bottomrule
\end{tabular}
\caption{Paired comparisons on the series-disjoint TSQA evaluation subset. Differences are computed as CGTime minus the listed baseline; positive values favor CGTime. Choice Macro Accuracy uses a stratified paired normal approximation with equal weight for multiple-choice and true/false questions. ROUGE-L uses a paired normal approximation over open-ended questions. Holm correction is applied separately to the six comparisons for each metric. PATRA is not included in the paired analysis because it uses a different split and evaluation protocol.}
\label{tab:full_tsqa_paired_tests}
\end{table*}

%% file: experiments/tables/full_qa_ablation_tests.tex
\begin{table*}[t]
\centering
\small
\setlength{\tabcolsep}{3pt}
\begin{tabular}{lcccc}
\toprule
Contrast & Metric-QA Overall $\Delta$ [95\% CI] & $p_{\mathrm{Holm}}$ & Metric-QA Multivariate $\Delta$ [95\% CI] & $p_{\mathrm{Holm}}$ \\ 
\midrule
\shortstack[l]{Joint final alignment $-$\\Internal-data-only final alignment} & +0.019 [+0.002, +0.036] & 0.0284 & +0.001 [-0.024, +0.026] & 0.9348 \\ 
\shortstack[l]{Joint final alignment $-$\\TSQA-only final alignment} & +0.286 [+0.268, +0.303] & $<10^{-4}$ & +0.283 [+0.258, +0.309] & $<10^{-4}$ \\ 
\shortstack[l]{Computed-statistics supervision $-$\\GPT-perceived supervision} & +0.106 [+0.087, +0.125] & $<10^{-4}$ & +0.125 [+0.098, +0.153] & $<10^{-4}$ \\ 
Full model $-$ Without Joint-GRPO & +0.057 [+0.040, +0.074] & $<10^{-4}$ & +0.019 [-0.006, +0.043] & 0.1977 \\ 
\shortstack[l]{Full model $-$\\Without Joint-SFT warmup} & +0.288 [+0.270, +0.305] & $<10^{-4}$ & +0.283 [+0.258, +0.309] & $<10^{-4}$ \\ 
\shortstack[l]{Full model $-$ Without the\\basic-to-complex curriculum} & +0.042 [+0.024, +0.061] & $<10^{-4}$ & +0.022 [-0.004, +0.048] & 0.1977 \\ 
\bottomrule
\end{tabular}
\caption{Paired comparisons of Gaussian Metric-QA scores for the reported ablations. Within each endpoint, Holm correction is applied separately to the final-alignment data-composition, supervision-source, and training-recipe comparison families. Each configuration was trained once, so the intervals quantify uncertainty across held-out evaluation items only.}
\label{tab:full_qa_ablation_tests}
\end{table*}

%% file: experiments/tables/full_caption_ablation_tests.tex
\begin{table*}[t]
\centering
\small
\setlength{\tabcolsep}{3pt}
\begin{tabular}{lcccc}
\toprule
Contrast & $\Delta$ Precision [95\% CI] & $p_{\mathrm{Holm}}$ & $\Delta$ Recall [95\% CI] & $p_{\mathrm{Holm}}$ \\ 
\midrule
\shortstack[l]{Joint final alignment $-$\\Internal-data-only final alignment} & +0.016 [+0.008, +0.024] & 0.0001 & +0.014 [+0.010, +0.018] & $<10^{-4}$ \\ 
\shortstack[l]{Joint final alignment $-$\\TSQA-only final alignment} & +0.135 [+0.126, +0.144] & $<10^{-4}$ & +0.087 [+0.083, +0.091] & $<10^{-4}$ \\ 
\shortstack[l]{Computed-statistics supervision $-$\\GPT-perceived supervision} & +0.262 [+0.251, +0.272] & $<10^{-4}$ & +0.132 [+0.128, +0.137] & $<10^{-4}$ \\ 
Full model $-$ Without Joint-GRPO & -0.012 [-0.020, -0.004] & 0.0079 & +0.011 [+0.007, +0.015] & $<10^{-4}$ \\ 
\shortstack[l]{Full model $-$\\Without Joint-SFT warmup} & +0.003 [-0.005, +0.011] & 0.4567 & +0.018 [+0.014, +0.022] & $<10^{-4}$ \\ 
\shortstack[l]{Full model $-$ Without the\\basic-to-complex curriculum} & -0.006 [-0.014, +0.003] & 0.3532 & +0.030 [+0.026, +0.034] & $<10^{-4}$ \\ 
\bottomrule
\end{tabular}
\caption{Paired comparisons of captioning metrics for the reported ablations. Conditional Precision uses paired bootstrap resampling of evaluation items, and Recall uses the paired normal approximation. Within each metric, Holm correction is applied separately to the final-alignment data-composition, supervision-source, and training-recipe comparison families. Each configuration was trained once.}
\label{tab:full_caption_ablation_tests}
\end{table*}

%% file: experiments/tables/full_tsqa_ablation_tests.tex
\begin{table}[t]
\centering
\small
\setlength{\tabcolsep}{2pt}
\begin{tabular}{lcc}
\toprule
Variant & \shortstack{Choice Macro Acc.\\$\Delta$ [95\% CI], $p_{\mathrm{Holm}}$} & \shortstack{ROUGE-L\\$\Delta$ [95\% CI], $p_{\mathrm{Holm}}$} \\ 
\midrule
\shortstack[l]{Internal-data-only\\final alignment} & \shortstack{+0.213\\{}[+0.179, +0.247]\\$p_{\mathrm{Holm}}<10^{-4}$} & \shortstack{+0.156\\{}[+0.151, +0.161]\\$p_{\mathrm{Holm}}<10^{-4}$} \\ 
\shortstack[l]{TSQA-only\\final alignment} & \shortstack{-0.062\\{}[-0.086, -0.039]\\$p_{\mathrm{Holm}}<10^{-4}$} & \shortstack{-0.021\\{}[-0.025, -0.017]\\$p_{\mathrm{Holm}}<10^{-4}$} \\ 
\bottomrule
\end{tabular}
\caption{Paired comparisons for the TSQA final-alignment data-composition ablation. Differences are computed as joint final alignment minus the listed variant. Holm correction is applied separately to Choice Macro Accuracy and ROUGE-L. Each configuration was trained once.}
\label{tab:full_tsqa_ablation_tests}
\end{table}

%% file: experiments/tables/full_ablation_diagnostic_tests.tex
\begin{table*}[t]
\centering
\small
\setlength{\tabcolsep}{3pt}
\begin{tabular}{llcc}
\toprule
Variant & Endpoint & $\Delta$ [95\% CI] & $p_{\mathrm{Holm}}$ \\ 
\midrule
\shortstack[l]{Without\\Joint-GRPO} & \shortstack[l]{Mean extracted values\\per caption} & +0.954 [+0.857, +1.050] & $<10^{-4}$ \\ 
\shortstack[l]{Without basic-to-\\complex curriculum} & \shortstack[l]{Caption Recall\\(multivariate)} & +0.056 [+0.049, +0.062] & $<10^{-4}$ \\ 
\shortstack[l]{Without basic-to-\\complex curriculum} & \shortstack[l]{TSQA Choice\\Macro Accuracy} & +0.008 [-0.010, +0.026] & 0.4003 \\ 
\shortstack[l]{Without basic-to-\\complex curriculum} & TSQA ROUGE-L & +0.014 [+0.010, +0.017] & $<10^{-4}$ \\ 
\bottomrule
\end{tabular}
\caption{Paired comparisons for selected secondary ablation endpoints. Differences are computed as the full model minus the listed variant. Each endpoint constitutes a separate comparison family. Each configuration was trained once.}
\label{tab:full_ablation_diagnostic_tests}
\end{table*}

%% file: experiments/appendix/reproduction_background.tex
\section{Technical Background for Reproduction}
\label{appendix_reproduction_background}

Table~\ref{tab:reproduction_background} directs readers unfamiliar with the
main technical components to background material useful for understanding and
reproducing our implementation.

\begin{table*}[t]
\centering
\begin{tabular}{@{}p{0.19\textwidth}p{0.25\textwidth}p{0.50\textwidth}@{}}
\toprule
Component & Background reference & Relevance to this work \\
\midrule
Principal component analysis
& \citet{jolliffe2002pca}
& Explained variance, retained components, and the PCA-based summaries used in
  multivariate data construction. \\
MOMENT
& \citet{goswami2024moment}
& The pretrained time-series encoder, patch representations, and masking
  conventions underlying the frozen encoder in CGTime. \\
Group relative policy optimization
& \citet{shao2024deepseekmath,guo2025deepseekr1}
& The grouped-rollout policy objective and reference-policy regularization used
  during Joint-GRPO. \\
\bottomrule
\end{tabular}
\caption{Background references for the principal technical components needed
to reproduce CGTime.}
\label{tab:reproduction_background}
\end{table*}